\pdfoutput=1

\documentclass[11pt]{article}
\usepackage{fontawesome5}

\usepackage{enumitem}
\usepackage{lipsum}
\usepackage{tikz}
\usetikzlibrary{trees,shapes}
\usepackage{xcolor}
\usepackage{epigraph}
\usepackage[margin=2cm]{geometry}
\usepackage{booktabs}

\usepackage[many]{tcolorbox}

\usepackage{enumitem}
\usepackage{tabularx}

\definecolor{algoPurple}{HTML}{6A51A3}
\definecolor{algoBlue}{HTML}{1F77B4}
\definecolor{algoGreen}{HTML}{2E8B57}
\definecolor{algoOrange}{HTML}{E67E22}

\makeatletter
\@ifundefined{faBalanceScale}{
  \@ifundefined{faCalculator}{
    \@ifundefined{faChartLine}{
    }{}
  }{}
}{}

\@ifundefined{faBrain}{
  \@ifundefined{faLightbulb}{
    \@ifundefined{faCompass}{
      
    }{}
  }{}
}{}

\@ifundefined{faUsersCog}{
  \@ifundefined{faUsers}{
    \@ifundefined{faProjectDiagram}{
      
    }{}
  }{}
}{}

\@ifundefined{faThumbsUp}{
  \@ifundefined{faHandshake}{
    \@ifundefined{faCheckDouble}{
      
    }{}
  }{}
}{}
\makeatother

\definecolor{AbsBack}{HTML}{EEF2FF}   
\definecolor{AbsFrame}{HTML}{5A67D8}  
\definecolor{AbsTitle}{HTML}{3B49B1}  

\newtcolorbox{abstractbox}{
  enhanced, breakable,
  colback=AbsBack, colframe=AbsFrame!85,
  boxrule=0.7pt,
  borderline={0.5pt}{0pt}{AbsFrame!40},
  arc=8pt, left=10pt, right=10pt, top=10pt, bottom=2pt,
  drop fuzzy shadow=AbsFrame!25
}

\usepackage{empheq}              
\usepackage[normalem]{ulem}      

\usepackage{amsthm}        

\usepackage{titlesec}

\titleformat{\section}
  {\PragyaHeadline\bfseries\Large\color{AbsTitle}}
  {\textcolor{AbsTitle}{\thesection}}
  {0.6em}
  {}

\titleformat{\subsection}
  {\PragyaHeadline\bfseries\large\color{AbsTitle}}
  {\textcolor{AbsTitle}{\thesubsection}}
  {0.5em}
  {}

\titleformat{\subsubsection}
  {\PragyaHeadline\bfseries\normalsize\color{AbsTitle}}
  {\textcolor{AbsTitle}{\thesubsubsection}}
  {0.5em}
  {}

\titlespacing*{\section}{0pt}{1.0ex plus .2ex}{0.6ex}
\titlespacing*{\subsection}{0pt}{0.8ex plus .2ex}{0.4ex}
\titlespacing*{\subsubsection}{0pt}{0.6ex plus .1ex}{0.3ex}

\newtcolorbox{defin}{colback=Teal!5!White,enhanced,title=GRIT — at-a-glance,	attach boxed title to top left={xshift=-4mm},boxrule=0pt,after skip=1cm,before skip=1cm,right skip=0cm,breakable,fonttitle=\bfseries,toprule=0pt,bottomrule=0pt,rightrule=0pt,leftrule=3pt,arc=0mm,skin=enhancedlast jigsaw,sharp corners,colframe=Teal!55!black,colbacktitle=Teal!55!black,boxed title style={
		frame code={ 
			\fill[Teal!25!black](frame.south west)--(frame.north west)--(frame.north east)--([xshift=3mm]frame.east)--(frame.south east)--cycle;
			\draw[line width=1mm,Teal!25!black]([xshift=2mm]frame.north east)--([xshift=5mm]frame.east)--([xshift=2mm]frame.south east);
			\draw[line width=1mm,Teal!25!black]([xshift=5mm]frame.north east)--([xshift=8mm]frame.east)--([xshift=5mm]frame.south east);
			\fill[Teal!25!black](frame.south west)--+(4mm,-2mm)--+(4mm,2mm)--cycle;
		}
	}
}
\usetikzlibrary{shapes.geometric, arrows}
\usetikzlibrary{decorations.markings}

\definecolor{first}{RGB}{210,255,140}
\definecolor{second}{RGB}{136, 162, 190}
\definecolor{third}{RGB}{129, 222, 228}
\definecolor{fourth}{RGB}{132, 84, 246}
\definecolor{fifth}{RGB}{250, 223, 112}
\definecolor{sixth}{RGB}{203, 193, 172}
\definecolor{seventh}{RGB}{88, 112, 246}
\definecolor{eighth}{RGB}{245, 192, 106}
\definecolor{nine}{RGB}{171, 162, 111}
\definecolor{ten}{RGB}{217, 217, 217}

\definecolor{paired-light-blue}{RGB}{198, 219, 239}
\definecolor{paired-dark-blue}{RGB}{49, 130, 188}
\definecolor{paired-light-orange}{RGB}{251, 208, 162}
\definecolor{paired-dark-orange}{RGB}{230, 85, 12}
\definecolor{paired-light-green}{RGB}{199, 233, 193}
\definecolor{paired-dark-green}{RGB}{49, 163, 83}
\definecolor{paired-light-purple}{RGB}{218, 218, 235}
\definecolor{paired-dark-purple}{RGB}{117, 107, 176}
\definecolor{paired-light-gray}{RGB}{217, 217, 217}
\definecolor{paired-dark-gray}{RGB}{99, 99, 99}
\definecolor{paired-light-pink}{RGB}{222, 158, 214}
\definecolor{paired-dark-pink}{RGB}{123, 65, 115}
\definecolor{paired-light-red}{RGB}{231, 150, 156}
\definecolor{paired-dark-red}{RGB}{131, 60, 56}
\definecolor{paired-light-yellow}{RGB}{231, 204, 149}
\definecolor{paired-dark-yellow}{RGB}{141, 109, 49}
\definecolor{Teal}{RGB}{0, 50, 50}
\definecolor{White}{RGB}{250, 250, 250}

\definecolor{bg1}{HTML}{FF9966}
\definecolor{bg2}{HTML}{CCE5FF}
\definecolor{bg3}{HTML}{FFCC99}
\definecolor{bg4}{HTML}{FFC107}
\definecolor{bg5}{HTML}{FFCCCC}
\definecolor{bg6}{HTML}{D5E8D4}
\definecolor{bg7}{HTML}{eeeeee}
\definecolor{bg8}{HTML}{cdeb8b}
\definecolor{bg9}{HTML}{dae8fc}
\definecolor{bg10}{HTML}{a2e6eb}

\definecolor{bg31}{HTML}{FFCDD2} 

\definecolor{bg32}{HTML}{F8BBD0}

\definecolor{bg33}{HTML}{E1BEE7} 

\definecolor{bg34}{HTML}{D7CCC8} 

\definecolor{bg35}{HTML}{B2DFDB} 

\definecolor{bg36}{HTML}{A5D6A7} 

\definecolor{bg37}{HTML}{FFF9C4} 

\definecolor{bg38}{HTML}{FFECB3} 

\definecolor{bg111}{HTML}{CB6843}

\definecolor{bg112}{HTML}{D77C5C}

\definecolor{bg113}{HTML}{E28E6E}
\definecolor{bg114}{HTML}{E89F7D}
\definecolor{bg115}{HTML}{EDAE8A}
\definecolor{bg116}{HTML}{F0BA95}
\definecolor{bg117}{HTML}{F3C29F}
\definecolor{bg118}{HTML}{F6CCAA}
\definecolor{bg119}{HTML}{F8D5B3}
\definecolor{bg120}{HTML}{FADCBD}
\definecolor{bg121}{HTML}{FCE6C7}

\definecolor{bg39}{HTML}{FFE0B2} 

\definecolor{bg40}{HTML}{3CB371} 

\definecolor{bg43}{HTML}{ffe5d9}

\definecolor{bg15}{HTML}{7FFFD4}

\definecolor{bg17}{HTML}{F0FFFF}

\definecolor{bg18}{HTML}{F5FFFA}

\definecolor{bg19}{HTML}{F8F8FF}

\definecolor{bg20}{HTML}{FFFFFF}

\definecolor{bg21}{HTML}{E1F5FE}

\definecolor{bg22}{HTML}{B3E5FC}

\definecolor{bg23}{HTML}{81D4FA}

\definecolor{bg24}{HTML}{4FC3F7}

\definecolor{bg25}{HTML}{29B6F6}

\definecolor{bg26}{HTML}{03A9F4}

\definecolor{bg27}{HTML}{039BE5}

\definecolor{bg28}{HTML}{0288D1}

\definecolor{bg29}{HTML}{0277BD}

\definecolor{bg30}{HTML}{01579B}

\definecolor{bg16}{HTML}{FFCC99}

\definecolor{pg51}{HTML}{E8F5E9} 
\definecolor{pg52}{HTML}{C8E6C9} 
\definecolor{pg53}{HTML}{B9F6CA} 
\definecolor{pg54}{HTML}{A9DFBF} 
\definecolor{pg55}{HTML}{BCF5A6} 

\definecolor{pg56}{HTML}{BEF1CE} 
\definecolor{pg57}{HTML}{CEF6EC} 
\definecolor{pg58}{HTML}{B7F0B1} 
\definecolor{pg59}{HTML}{B1F2B5} 
\definecolor{pg60}{HTML}{9DF3C4} 

\definecolor{pg61}{HTML}{DEF7E0} 
\definecolor{pg62}{HTML}{E8F8DC} 

\definecolor{pg63}{HTML}{EBF7E7} 
\definecolor{pg64}{HTML}{F0FDF4} 

\definecolor{pg65}{HTML}{F1FEE7} 
\definecolor{pg66}{HTML}{F7FFF6} 
\definecolor{pg67}{HTML}{FCFFE7} 
\definecolor{pg68}{HTML}{F4FFD2} 
\definecolor{pg69}{HTML}{EEFFE2} 
\definecolor{pg70}{HTML}{E3FDF5} 

\definecolor{connect-color}{RGB}{0,0,0}
\definecolor{middle-color}{RGB}{255,255,255}
\definecolor{leaf-color}{RGB}{173,216,230}
\definecolor{line-color}{RGB}{25,25,112}

\newtcolorbox{societal_harm}{
  colback=soothingPurple, 
  colframe=black, 
  boxrule=0pt,
  enhanced,
  title=Societal harm,
  attach boxed title to top right={yshift=-3mm},
  fonttitle=\bfseries,
  toprule=1pt,
  bottomrule=1pt,
  rightrule=1pt,
  leftrule=1pt,
  arc=1mm
}

\newtcolorbox{privacy_violation}{
  colback=soothingPurple, 
  colframe=black, 
  boxrule=0pt,
  enhanced,
  title=Privacy Violation,
  attach boxed title to top right={yshift=-3mm},
  fonttitle=\bfseries,
  toprule=1pt,
  bottomrule=1pt,
  rightrule=1pt,
  leftrule=1pt,
  arc=1mm
}

\newtcolorbox{disinformation_deception}{
  colback=soothingPurple, 
  colframe=black, 
  boxrule=0pt,
  enhanced,
  title=Disinformation \& Deception,
  attach boxed title to top right={yshift=-3mm},
  fonttitle=\bfseries,
  toprule=1pt,
  bottomrule=1pt,
  rightrule=1pt,
  leftrule=1pt,
  arc=1mm
}

\newtcolorbox{answer_disparity}{
  colback=soothingPurple, 
  colframe=black, 
  boxrule=0pt,
  enhanced,
  title=Answer disparity,
  attach boxed title to top right={yshift=-3mm},
  fonttitle=\bfseries,
  toprule=1pt,
  bottomrule=1pt,
  rightrule=1pt,
  leftrule=1pt,
  arc=1mm
}

\newtcolorbox{wrong_classification}{
  colback=soothingPurple, 
  colframe=black, 
  boxrule=0pt,
  enhanced,
  title=Wrong classification,
  attach boxed title to top right={yshift=-3mm},
  fonttitle=\bfseries,
  toprule=1pt,
  bottomrule=1pt,
  rightrule=1pt,
  leftrule=1pt,
  arc=1mm
}

\newtcolorbox{goal_hijacking}{
  colback=soothingPurple, 
  colframe=black, 
  boxrule=0pt,
  enhanced,
  title=Goal hijacking,
  attach boxed title to top right={yshift=-3mm},
  fonttitle=\bfseries,
  toprule=1pt,
  bottomrule=1pt,
  rightrule=1pt,
  leftrule=1pt,
  arc=1mm
}

\newtcolorbox{control_generation}{
  colback=soothingPurple, 
  colframe=black, 
  boxrule=0pt,
  enhanced,
  title=Control generation,
  attach boxed title to top right={yshift=-3mm},
  fonttitle=\bfseries,
  toprule=1pt,
  bottomrule=1pt,
  rightrule=1pt,
  leftrule=1pt,
  arc=1mm
}

\newtcolorbox{prompt_leaking}{
  colback=soothingPurple, 
  colframe=black, 
  boxrule=0pt,
  enhanced,
  title=Prompt leaking,
  attach boxed title to top right={yshift=-3mm},
  fonttitle=\bfseries,
  toprule=1pt,
  bottomrule=1pt,
  rightrule=1pt,
  leftrule=1pt,
  arc=1mm
}

\definecolor{soothingPurple}{RGB}{195, 160, 201}
\definecolor{hidden-draw}{RGB}{20,68,106}
\definecolor{hidden-pink}{RGB}{255,245,247}
\definecolor{dark-red}{RGB}{233, 150, 122}
\definecolor{light-red}{RGB}{255,182,193}
\definecolor{medium-red}{RGB}{205,92,92}
\definecolor{light-yellow}{RGB}{255, 239, 153}
\definecolor{light-blue}{RGB}{173, 216, 230}
\definecolor{paired-light-yellow}{HTML}{FFFF88}
\definecolor{paired-light-blue}{HTML}{CCE5FF}
\definecolor{paired-light-orange}{HTML}{FFCC99}
\definecolor{paired-dark-yellow}{HTML}{FFF2CC}
\definecolor{paired-light-pink}{HTML}{FFCCCC}
\definecolor{paired-cyan}{HTML}{D5E8D4}
\definecolor{paired-gray}{HTML}{eeeeee}
\definecolor{paired-green}{HTML}{cdeb8b}
\definecolor{paired-blue}{HTML}{dae8fc}
\definecolor{paired-dark-cyan}{HTML}{a2e6eb}
\definecolor{paired-dark-pink}{HTML}{e7b2d2}
\definecolor{paired-purple}{HTML}{9999ff}
\definecolor{paired-pink}{HTML}{cc99ff}
\definecolor{paired-orange}{HTML}{ffcc99}

\definecolor{a1}{RGB}{241,233,191}
\definecolor{a2}{RGB}{255,241,218}

\definecolor{a3}{RGB}{255,239,213}
\definecolor{a4}{RGB}{250,235,215}
\definecolor{a5}{RGB}{255,239,219}
\definecolor{a6}{RGB}{255,246,225}
\definecolor{a7}{RGB}{246,227,201}
\definecolor{a8}{RGB}{254,235,226}
\definecolor{a9}{RGB}{247,220,111}
\definecolor{a10}{RGB}{199,211,189}
\definecolor{a11}{RGB}{209,196,233}
\definecolor{a12}{RGB}{214,234,248}
\definecolor{a13}{RGB}{232,245,233}
\definecolor{a14}{RGB}{237,248,177}
\definecolor{a15}{RGB}{255,228,225}
\definecolor{a16}{RGB}{255,228,181}
\definecolor{a17}{RGB}{255,222,173}
\definecolor{a18}{RGB}{255,218,185}
\definecolor{a19}{RGB}{255,203,164}
\definecolor{a20}{RGB}{247,202,201}

\definecolor{a21}{RGB}{241,254,255}
\definecolor{a22}{RGB}{230,252,252}
\definecolor{a23}{RGB}{179,236,255}
\definecolor{a24}{RGB}{174,226,249}
\definecolor{a25}{RGB}{208,234,246}
\definecolor{a26}{RGB}{189,226,219}
\definecolor{a27}{RGB}{177,204,201}

\definecolor{a28}{RGB}{216,195,216}
\definecolor{a29}{RGB}{195,155,211}
\definecolor{a30}{RGB}{208,152,223}
\definecolor{a31}{RGB}{255,183,209}
\definecolor{a32}{RGB}{255,167,209}
\definecolor{a33}{RGB}{254,235,167}
\definecolor{a34}{RGB}{255,222,137}
\definecolor{a35}{RGB}{254,180,154}
\definecolor{a36}{RGB}{247,148,161}
\definecolor{a37}{RGB}{239,154,154}
\definecolor{a38}{RGB}{255,130,171}
\definecolor{a39}{RGB}{255,105,180}
\definecolor{a40}{RGB}{251,142,172}

\usepackage{ACL2023}

\usepackage[edges]{forest}
\usepackage{lipsum}
\usepackage{tikz}
\usetikzlibrary{trees,shapes}

\usepackage{times}
\usepackage{latexsym}

\usepackage[T1]{fontenc}

\usepackage[utf8]{inputenc}

\usepackage{microtype}

\usepackage{inconsolata}
\usepackage{soul}

\usepackage{inconsolata}
\usepackage{algorithm}
\usepackage{algpseudocode}
\usepackage{amsmath,amssymb}
\usepackage{soul}
\usepackage{tikz}
\usepackage{pifont} 
\usepackage{amssymb} 

\tikzset{rndblock/.style={rounded corners,rectangle,draw,scale=0.8,outer sep=0pt}}

\usetikzlibrary{shapes.geometric}
\usepackage{framed}
\usepackage{enumitem}
\newlist{RQ}{enumerate}{1}
\setlist[RQ]{label=\textbf{RQ\,\arabic*},ref={RQ\,\arabic*}}
\usepackage{comment}
\usepackage{natbib}
\usepackage{multibib}
\makeatletter
\usepackage{booktabs}
\usepackage[inkscapeformat=png]{svg}
\usepackage{graphicx}
\usepackage{caption}
\usepackage{subcaption}
\usepackage{tabularx}
\usepackage{soul}
\usepackage{float}
\usepackage{enumitem}
\usepackage{pifont}
\usepackage{lipsum}

\usepackage[many]{tcolorbox}

\usetikzlibrary{shapes.geometric, arrows}
\usetikzlibrary{decorations.markings}

\usepackage{fancybox}
\usepackage{xcolor}
\usepackage{hyperref}
 \definecolor{darkblue}{rgb}{0, 0, 0.5}
  \hypersetup{colorlinks=true, citecolor=darkblue, linkcolor=darkblue, urlcolor=darkblue}

\definecolor{vgreen}{HTML}{60A917}
\definecolor{vred}{HTML}{CE3A29}

\usepackage{xstring}
\usepackage{longtable}

\usepackage{tabularray}

\DefTblrTemplate{firsthead,middlehead,lasthead}{default}{
}
\DefTblrTemplate{firstfoot}{default}{
  \UseTblrTemplate{contfoot}{default}
  \UseTblrTemplate{caption}{default}
}
\DefTblrTemplate{middlefoot}{default}{
  \UseTblrTemplate{contfoot}{default}
  \UseTblrTemplate{capcont}{default}
}
\DefTblrTemplate{lastfoot}{default}{
  \UseTblrTemplate{note}{default}
  \UseTblrTemplate{remark}{default}
  \UseTblrTemplate{capcont}{default}
}

\newcolumntype{P}[1]{>{\centering\arraybackslash}p{#1}}

\usepackage{color}
\tcbuselibrary{skins}

\usepackage[export]{adjustbox} 

\usepackage{setspace}
\usepackage[capitalise,nameinlink]{cleveref}

\crefname{section}{Sec.}{Sec.}

\usepackage{microtype}
\usepackage{hyperref}
\usepackage{graphicx}
\usepackage{comment}
\usepackage{amsmath}
\usepackage{amssymb}
\usepackage{algorithm}
\usepackage{algpseudocode}
\usepackage{colortbl}
\usepackage[export]{adjustbox} 
\usepackage{varwidth}
\usepackage{enumitem}
\setlist{leftmargin=1mm}
\usepackage{pifont}
\usepackage{booktabs}
\usepackage{multirow}
\usepackage{subcaption}
\usepackage{resizegather}
\usepackage{breqn}
\usepackage[capitalise]{cleveref}
\usepackage{graphicx}
\usepackage{tikz}
\usetikzlibrary{shapes.geometric, arrows}
\usetikzlibrary{decorations.markings}
\usepackage{soul}
\usepackage{wrapfig,graphicx,lipsum}
\usepackage{extsizes}
\usepackage{cuted}
\usepackage{flushend}
\usepackage{float}
\usepackage{changepage,threeparttable}
\usepackage{setspace}
\usepackage{caption}
\usepackage{booktabs}
\usepackage{dblfloatfix} 
\usepackage{fixltx2e}
\usepackage[normalem]{ulem}

\usepackage{environ}
\usepackage{soul}
\usepackage{float}
\usepackage{enumitem}
\usepackage{pifont}
\usepackage{lipsum}

\usepackage[many]{tcolorbox}

\usetikzlibrary{shapes.geometric, arrows}
\usetikzlibrary{decorations.markings}

\usepackage{fancybox}
\usepackage{xcolor}
\usepackage{hyperref}
 \definecolor{darkblue}{rgb}{0, 0, 0.5}
  \hypersetup{colorlinks=true, citecolor=darkblue, linkcolor=darkblue, urlcolor=darkblue}

\definecolor{vgreen}{HTML}{60A917}
\definecolor{vred}{HTML}{CE3A29}

\usepackage{xstring}
\usepackage{longtable}

\usepackage{tabularray}

\DefTblrTemplate{firsthead,middlehead,lasthead}{default}{
}
\DefTblrTemplate{firstfoot}{default}{
  \UseTblrTemplate{contfoot}{default}
  \UseTblrTemplate{caption}{default}
}
\DefTblrTemplate{middlefoot}{default}{
  \UseTblrTemplate{contfoot}{default}
  \UseTblrTemplate{capcont}{default}
}
\DefTblrTemplate{lastfoot}{default}{
  \UseTblrTemplate{note}{default}
  \UseTblrTemplate{remark}{default}
  \UseTblrTemplate{capcont}{default}
}

\usepackage{color}
\tcbuselibrary{skins}

\usepackage[export]{adjustbox} 

\usepackage{setspace}
\usepackage[capitalise,nameinlink]{cleveref}

\crefname{section}{Sec.}{Sec.}

\usepackage{microtype}
\usepackage{hyperref}
\usepackage{graphicx}
\usepackage{comment}
\usepackage{amsmath}
\usepackage{amssymb}
\usepackage{algorithm}
\usepackage{algpseudocode}
\usepackage{colortbl}
\usepackage[export]{adjustbox} 
\usepackage{varwidth}
\usepackage{enumitem}
\setlist{leftmargin=1mm}
\usepackage{pifont}
\usepackage{booktabs}
\usepackage{multirow}
\usepackage{subcaption}
\usepackage{resizegather}
\usepackage{breqn}
\usepackage[capitalise]{cleveref}
\usepackage{graphicx}
\usepackage{tikz}
\usetikzlibrary{shapes.geometric, arrows}
\usetikzlibrary{decorations.markings}
\usepackage{soul}
\usepackage{wrapfig,graphicx,lipsum}
\usepackage{extsizes}
\usepackage{cuted}
\usepackage{flushend}
\usepackage{float}
\usepackage{changepage,threeparttable}
\usepackage{setspace}
\usepackage{caption}
\usepackage{booktabs}
\usepackage{dblfloatfix} 
\usepackage{fixltx2e}
\usepackage[normalem]{ulem}

\usepackage{tcolorbox}
\usepackage{amsmath}
\usepackage{amssymb}
\usepackage{caption}

\usepackage{environ}

\newlength{\myl}
\expandafter\let\expandafter\origequation\csname equation*\endcsname
\expandafter\let\expandafter\endorigequation\csname endequation*\endcsname
\long\def\[#1\]{\begin{equation*}#1\end{equation*}}
\RenewEnviron{equation*}{
  \settowidth{\myl}{$\displaystyle\BODY$} 
  \origequation
    \ifdim\myl>\linewidth
      \resizebox{\linewidth}{!}{$\displaystyle\BODY$}
    \else
      \BODY 
    \fi
  \endorigequation
}

\makeatletter
\newcommand{\DrawLine}{%
  \begin{tikzpicture}
  \path[use as bounding box] (0,0) -- (\linewidth,0);
  \draw[color=blue!75!black,dashed,dash phase=.5pt]
        (0-\kvtcb@leftlower-\kvtcb@boxsep,0)--
        (\linewidth+\kvtcb@rightlower+\kvtcb@boxsep,0);
  \end{tikzpicture}%
  }
\makeatother

\usepackage{pgfplots}
\usepackage{tikz}
\usetikzlibrary{positioning, arrows.meta}
\usepackage{longtable}
\usepackage{booktabs}
\usepackage{array}
\usepackage{adjustbox}
\usepackage{longtable}
\usepackage{pifont}

\usepackage{booktabs}
\usepackage{longtable}
\usepackage{array}
\usepackage[table]{xcolor}
\usepackage{colortbl}
\usepackage{caption}
\usepackage{enumitem}
\usepackage{pifont}
\usepackage{natbib}
\usepackage{fontawesome5} 
\usepackage{capt-of}

\definecolor{CheckGreen}{HTML}{1B8A3A}
\definecolor{WarnAmber}{HTML}{B7791F}
\definecolor{CrossRed}{HTML}{B00020}
\definecolor{HeaderGray}{HTML}{F3F4F6}
\definecolor{RowAlt}{HTML}{FAFAFA}

\newcolumntype{L}[1]{>{\raggedright\arraybackslash}p{#1}} 

\usepackage[table]{xcolor}

\definecolor{rowA}{HTML}{F7F7FF}
\definecolor{rowB}{HTML}{F6FFFA}
\definecolor{ink}{HTML}{111111}

\newcolumntype{L}[1]{>{\raggedright\arraybackslash}p{#1}}

\usepackage{enumitem}
\usepackage{adjustbox}
\usepackage{multirow}
\usepackage{colortbl}
\usepackage{xcolor}
\definecolor{pretrainingloss}{HTML}{84B8A2}
\definecolor{finetunetokens}{HTML}{F4C2A7}
\definecolor{paramrelativeefficiency}{HTML}{AED6E5}
\definecolor{pretrainingmixture}{HTML}{F2B8C4}
\definecolor{headerblue}{RGB}{28,55,95}
\definecolor{bandblue}{RGB}{232,239,249}
\definecolor{bandgray}{gray}{0.96}
\renewcommand\arraystretch{1.2}
\usepackage{array,makecell}
\newcolumntype{M}[1]{>{\centering\arraybackslash}m{#1}} 

\renewcommand\arraystretch{1.18}
\usepackage[htt]{hyphenat} 

\definecolor{headerblue}{RGB}{70, 130, 180}
\definecolor{lightblue}{RGB}{230, 240, 250}
\definecolor{darkblue}{RGB}{25, 25, 112}

\newcommand{\incfig}[2][]{%
  \IfFileExists{#2}{\includegraphics[#1]{#2}}{\fbox{\parbox[c][3cm][c]{0.8\linewidth}{\centering Missing figure: #2}}}%
}

\usepackage[many]{tcolorbox}

\usetikzlibrary{shapes.geometric, arrows}
\usetikzlibrary{decorations.markings}

\usepackage{fancybox}
\usepackage{xcolor}
\usepackage{hyperref}
 \definecolor{darkblue}{rgb}{0, 0, 0.5}
  \hypersetup{colorlinks=true, citecolor=darkblue, linkcolor=darkblue, urlcolor=darkblue}

\definecolor{vgreen}{HTML}{60A917}
\definecolor{vred}{HTML}{CE3A29}

\usepackage{xstring}
\usepackage{longtable}

\usepackage{tabularray}

\DefTblrTemplate{firsthead,middlehead,lasthead}{default}{
}
\DefTblrTemplate{firstfoot}{default}{
  \UseTblrTemplate{contfoot}{default}
  \UseTblrTemplate{caption}{default}
}
\DefTblrTemplate{middlefoot}{default}{
  \UseTblrTemplate{contfoot}{default}
  \UseTblrTemplate{capcont}{default}
}
\DefTblrTemplate{lastfoot}{default}{
  \UseTblrTemplate{note}{default}
  \UseTblrTemplate{remark}{default}
  \UseTblrTemplate{capcont}{default}
}

\newcolumntype{P}[1]{>{\centering\arraybackslash}p{#1}}

\usepackage{color}
\tcbuselibrary{skins}

\usepackage[export]{adjustbox} 

\usepackage{setspace}
\usepackage[capitalise,nameinlink]{cleveref}

\crefname{section}{Sec.}{Sec.}

\usepackage{microtype}
\usepackage{hyperref}

\title{\includegraphics[width=0.95\textwidth]{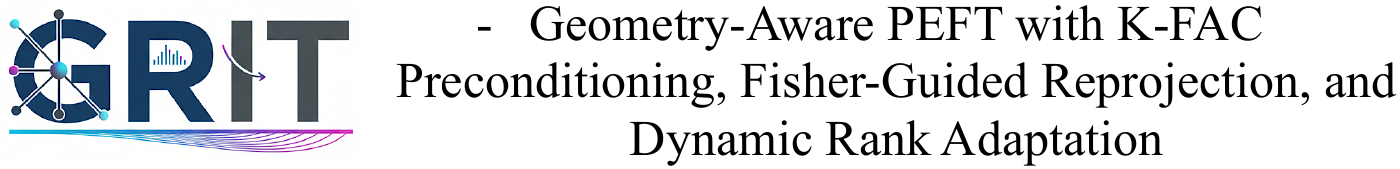}}

\author{
  Pritish Saha$^{1}$ \quad
  Chandrav Rajbangshi$^{1}$ \quad
  Rudra Goyal$^{1}$ \quad
  Mohit Goyal$^{1}$ \quad\\
  \textbf{Anurag Deo}$^{1}$ \quad
  \textbf{Biswajit Roy}$^{1}$ \quad
  \textbf{Ningthoujam Dhanachandra Singh}$^{1}$ \quad\\
  \textbf{Raxit Goswami}$^{1}$ \quad
  \textbf{Amitava Das}$^{2}$ \\
  $^{1}$RAAPID Lab, USA \quad
  $^{2}$Pragya Lab, BITS Pilani, Goa 
}

\begin{document}
\setcitestyle{square}
\maketitle
\begin{abstract}
\textbf{Parameter-efficient fine-tuning (PEFT)} has become the default route for adapting LLMs to \emph{domain- and application-specific} settings, yet widely used methods such as \emph{LoRA} and \emph{QLoRA} remain largely \emph{geometry-agnostic}: they optimize within \emph{fixed, randomly oriented} low-rank subspaces using first-order descent, largely \emph{ignoring} local \textbf{loss curvature}. This can \emph{inflate the effective update budget} and \emph{amplify} \textbf{drift} along weakly constrained directions. We introduce \textbf{GRIT}, a \emph{dynamic, curvature-aware} LoRA procedure. GRIT preserves the LoRA parameterization but: (1) \textbf{preconditions gradients} in rank space using \emph{K-FAC} as a natural-gradient proxy; (2) periodically \textbf{reprojects} the low-rank basis onto dominant \emph{Fisher eigendirections} to suppress drift; and (3) \textbf{adapts the effective rank} from the spectrum so capacity concentrates where signal resides. The net effect is to steer updates toward \emph{high-signal, low-interference} directions while using \emph{fewer effective parameters}. Across \emph{instruction-following}, \emph{comprehension}, and \emph{reasoning} benchmarks on LLaMA backbones, \textbf{GRIT} matches or surpasses \textit{LoRA}/\textit{QLoRA} while \textbf{reducing trainable parameters by $\sim\!46\%$ on average} (25--80\% across tasks) without degrading quality in practice, across diverse prompt styles and varying data mixes. Since fine-tuning often induces \emph{\textbf{catastrophic forgetting}}, we model GRIT’s drift via a curvature-modulated power law {\tiny$\boxed{L_{pt}^{\mathrm{GRIT}}=L_{pt}^{0}+A\,\dfrac{D_{ft}^{\beta}}{(\Xi_{\mathrm{GRIT}}N)^{\alpha}}+E}$} with {\scriptsize$\Xi_{\mathrm{GRIT}}=(1+\gamma_{r}r_{\mathrm{eff}})(1+\gamma_{a}\rho_{\mathrm{align}})(1+\gamma_{p}\pi_{\mathrm{proj}})$}, capturing \emph{effective rank}, \emph{Fisher alignment}, and \emph{projection fidelity}; empirically, GRIT yields consistently lower drift than \textit{LoRA} and improves the parameter-updates--vs--retention frontier against strong PEFT and optimizer baselines (e.g., \textit{Orthogonal-LoRA}, \textit{IA$^3$}, \textit{DoRA}/\textit{Eff-FT}, \textit{Shampoo}). \href{https://anonymous.4open.science/r/acl2026-submission}{Code repository}.

\end{abstract}

\vspace{-3em}
\begin{defin}

\scriptsize

\vspace{-2mm}
\begin{itemize}[left=-4pt,itemsep=0pt,topsep=0pt,parsep=0pt]

\item[\faBolt] \textbf{\textit{TL;DR}}: \textbf{GRIT is a \emph{geometry-aware LoRA} method that concentrates low-rank adaptation along measured curvature}, reducing update footprint while preserving quality via \textbf{K-FAC rank-space preconditioning}, \textbf{Fisher-guided reprojection}, and \textbf{dynamic rank}. 

\item[\faCogs] \textbf{\textit{Core mechanism}}: GRIT performs \textbf{periodic reprojection} of LoRA factors onto \textbf{top Fisher eigendirections} and uses a \textbf{cumulative eigen-mass rule} (with bounds/hysteresis) to pick the \textbf{effective rank} on the fly. 

\item[\faThLarge] \textbf{\textit{Update footprint}}: In the illustrated depth policy (freeze layers 1--10, adapt 11--32), GRIT reduces \textbf{mean update density} and total updated parameters relative to LoRA and LoRA+K-FAC, yielding a tighter adaptation footprint across layers. 

\item[\faClock] \textbf{\textit{Practical overhead}}: GRIT keeps heavy operations in $r\times r$ and triggers \textbf{few reprojections} (about \textbf{1.8--2.4 per 1k steps} in the timing setup), giving \textbf{single-digit \% mean step-time overhead} with \textbf{P99 spikes} aligned to reprojection events.

\item[\faProjectDiagram] \textbf{\textit{What to log / audit}}: The appendix formalizes an \textbf{effective capacity multiplier} $\Xi_{\text{GRIT}}$ in terms of \textbf{effective rank} $r_{\mathrm{eff}}$, \textbf{alignment overlap} $\rho_{\mathrm{align}}$, and \textbf{retained mass} $\pi_{\mathrm{proj}}$—actionable telemetry to verify that “geometry-aware compression” is actually happening. 

\item[\faExclamationTriangle] \textbf{\textit{Limits (be humble)}}: Performance depends on \textbf{curvature estimation quality} (K-FAC assumptions, early Fisher noise) and on the \textbf{projection frequency} knob $T_{\text{proj}}$, which trades stability for compute.

\end{itemize}

\vspace{-0.5em}
\end{defin}
\vspace{-2.5em}



\begin{figure*}[ht!]
  \centering

  \begin{minipage}[t]{0.59\textwidth}
    \centering
    \includegraphics[width=\linewidth]{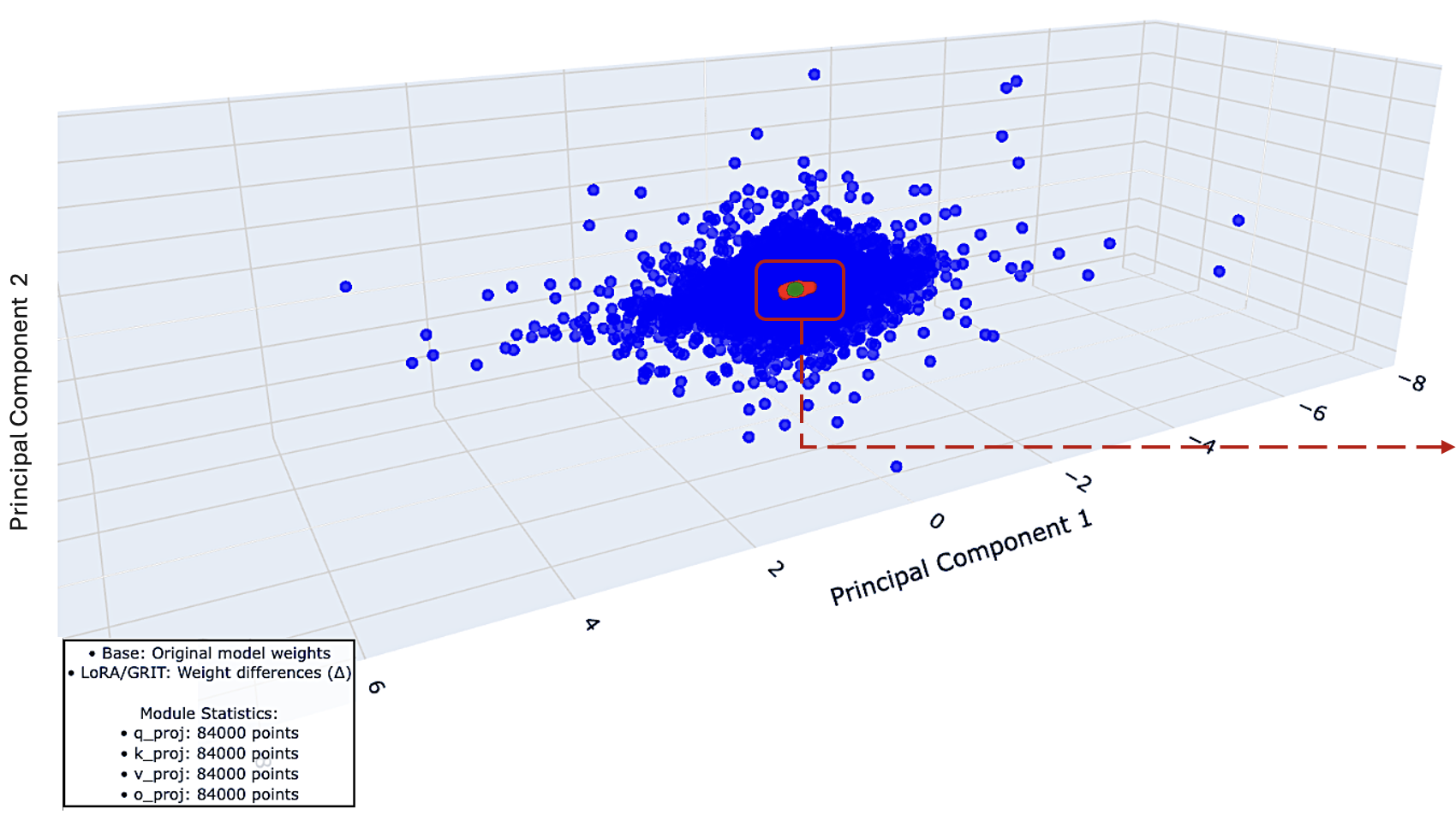}
    \vspace{2pt}
    {\small \textbf{(a)} Global view: original model weights vs.\ LoRA/GRIT $\Delta$-weights in PCA space.}
  \end{minipage}
  \hfill
  \begin{minipage}[t]{0.39\textwidth}
    \centering
    \includegraphics[width=\linewidth]{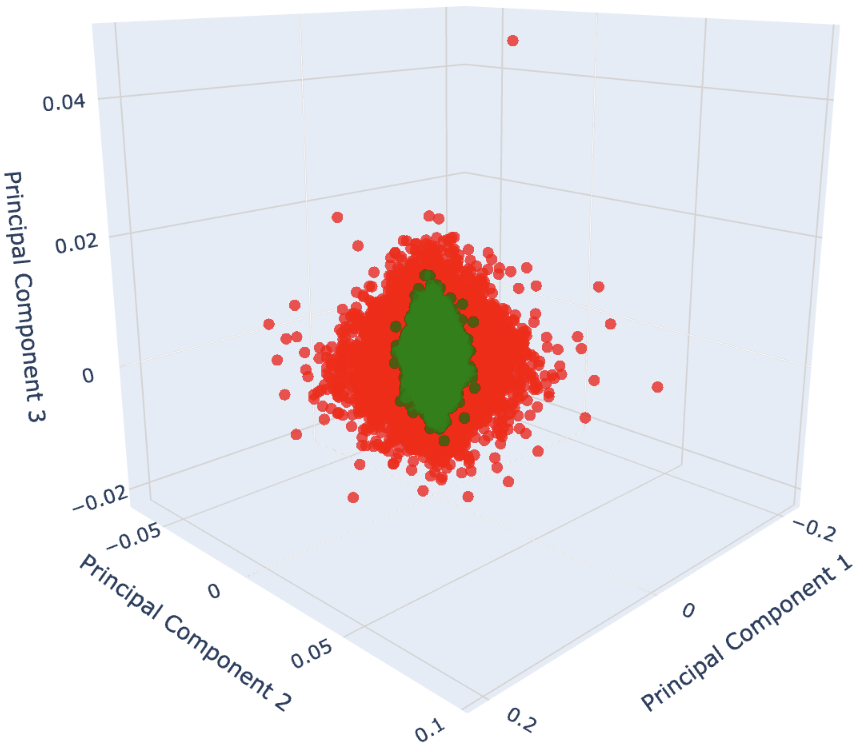}
    {\small \textbf{(b)} Zoom near the origin comparing LoRA vs.\ GRIT $\Delta$-weights only.}
  \end{minipage}
  
\caption{\textbf{Geometry of parameter updates: GRIT concentrates $\Delta$-weights into curvature-aligned subspaces.}
\emph{Setup.} PCA on parameter vectors from attention projections ($q_{\mathrm{proj}},k_{\mathrm{proj}},v_{\mathrm{proj}},o_{\mathrm{proj}}$) across layers; points are mean-centered and embedded into the leading PCs (no extra scaling).
\textbf{(a) Overall (left).} \emph{\textcolor{blue}{Blue}} points depict the \textbf{entire model’s parameter space} as realized by the \emph{original (base) weights}. Superimposed at the center, \emph{\textcolor{red}{red}} points are \textbf{LoRA} \emph{update deltas} ($\Delta W$), and \emph{\textcolor{green}{green}} points are \textbf{GRIT} \emph{update deltas}. The visual shows that the full base space is broad and anisotropic, while both LoRA and GRIT operate in a much \emph{smaller central region}—the effective fine-tuning manifold.
\textbf{(b) Zoom on $\Delta$ (right).} The base cloud is omitted to compare updates directly: \emph{red} = LoRA $\Delta W$, \emph{green} = GRIT $\Delta W$. GRIT forms a \emph{tighter, ellipsoidal core} with reduced radial spread versus LoRA, consistent with \emph{rank-space natural-gradient} preconditioning and \emph{Fisher-aligned reprojection} that bias updates toward high-curvature eigendirections while suppressing diffuse, low-signal axes.
\textbf{Interpretation.} GRIT densifies signal in a low-rank, curvature-aware subspace (smaller support, tighter covariance) without enlarging the update footprint—steering limited parameters toward directions that matter for stability and generalization.
}
\label{fig:grit-minipage}
\vspace{-1.5em}
\end{figure*}

\vspace{-0.3em}
\section{PEFT’s Blind Spot: Learning Inertia \& Catastrophic Forgetting}
\vspace{-0.3em}

Adapting billion--parameter LLMs stresses memory and bandwidth; PEFT addresses this by freezing most weights and training only a \emph{\textbf{small parameter subset}} (e.g., low--rank updates). Among PEFT variants, \textbf{LoRA}~\citep{hu2021lora} and \textbf{QLoRA}~\citep{dettmers2023qlora} have become de facto standards. \textbf{The emerging trade--off}: Recent evidence that \emph{``\textbf{LoRA learns less and forgets less}''}~\citep{biderman2024loralearnsforgets} indicates that, relative to full fine--tuning, LoRA's low--rank update budget often yields \emph{\textbf{smaller gains on hard targets}} (e.g., code/math) while \emph{\textbf{preserving}} more of the base model's broad abilities (e.g., commonsense and general language competence). In short, PEFT often trades \emph{\textbf{peak task improvement}} for \emph{\textbf{retention}}: constrained adapters may underfit challenging distributions yet induce fewer high--impact shifts that erase pretraining knowledge.

\subsection{Diagnosis: Learning Inertia vs.\ High-Impact Updates}
\label{sec:diagnosis-geometry}

\textbf{Forgetting is not just how much you train—it’s \emph{where} you move.}
PEFT imposes \emph{\textbf{learning inertia}} by freezing most weights and restricting adaptation to a low\hyp{}rank subspace, yet interference still arises when updates overlap \emph{\textbf{high\hyp{}curvature}} modes of the pretraining objective. Let $L_{\text{pt}}(w)$ be the pretraining loss and let $\Delta w$ denote the PEFT\hyp{}induced update. Near a well\hyp{}fit solution, first\hyp{}order terms are small and the \emph{\textbf{quadratic term}} dominates:
\begingroup
\setlength{\abovedisplayskip}{5pt}\setlength{\belowdisplayskip}{5pt}
\[
\Delta L_{\text{pt}}
\;\approx\;
\tfrac{1}{2}\,\Delta w^{\top} H_{\text{pt}}\,\Delta w
\;=\;
\tfrac{1}{2}\sum_{j}\lambda_j\,\bigl(u_j^{\top}\Delta w\bigr)^2,
\]
\endgroup
where $H_{\text{pt}}=\sum_{j}\lambda_j u_j u_j^\top$ is the \emph{\textbf{Hessian eigendecomposition}}. 
\textbf{Intuition.} Forgetting is large when landscape is sharp (large $\lambda_j$) and when the update has large projections onto those sharp directions (large $|u_j^{\top}\Delta w|$) \citep{pascanu2013difficulty,ghorbani2019hessian,keskar2017sharpminima,dinh2017sharpminima}.

\textbf{From principle to practice.}
The quadratic form explains \emph{why} forgetting increases but not \emph{what} to monitor during training. We therefore introduce two \emph{operational} geometry summaries that map directly onto the quadratic term and are simple to track online. \textbf{Two geometric amplifiers}:
\\
\emph{(i) Tail mass of updates.}
\begingroup
\setlength{\abovedisplayskip}{5pt}\setlength{\belowdisplayskip}{5pt}
\[
U_{\text{hi}}(\tau)\;=\;\sum_{i}\mathbf{1}\!\bigl(|\Delta w_i|>\tau\bigr),
\]
\endgroup
the count of coordinates exceeding a magnitude threshold $\tau$. 
\textbf{Interpretation:} heavier tails imply more frequent large coordinates, increasing the chance that $|u_j^\top\Delta w|$ is large and thus amplifying the quadratic loss rise; this aligns with continual\hyp{}learning evidence that large, concentrated steps drive interference \citep{kirkpatrick2017ewc,zenke2017si,aljundi2018mas,chaudhry2019agem}.

\emph{(ii) Effective rank of the update covariance.}
Let $\{\lambda^{(\Delta)}_j\}$ be the eigenvalues (descending) of $\mathbb{E}[\Delta w\,\Delta w^\top]$. Let's define: 
\begingroup
\setlength{\abovedisplayskip}{5pt}\setlength{\belowdisplayskip}{5pt}
\[
r_{\text{eff}}
\;=\;
\min\Bigl\{k:\ \frac{\sum_{j=1}^{k}\lambda^{(\Delta)}_j}{\sum_{j}\lambda^{(\Delta)}_j}\ge \eta\Bigr\}
\quad(\eta\in(0,1)).
\]
\endgroup
\textbf{Interpretation:} larger $r_{\text{eff}}$ means update energy is spread across more directions, \emph{raising the probability} of overlap with sharp Hessian modes and increasing $\sum_j \lambda_j (u_j^\top \Delta w)^2$ \citep{roy2007effective,gavish2014optimal,aghajanyan2021intrinsic}.

\paragraph{Takeaway.}
\emph{\textbf{Tail mass}} controls \emph{how big} the projections can be; \emph{\textbf{effective rank}} controls \emph{how many} directions those projections can land on. Either rising makes curvature overlap—and hence $\Delta L_{\text{pt}}$—more likely. This diagnosis motivates \emph{geometry\hyp{}aware} PEFT procedures that \emph{shrink tails} and \emph{concentrate rank} away from sharp modes.

\paragraph{\textbf{Evidence for the blind spot.}}
Large-scale evaluations show a \emph{\textbf{stable Pareto}}: LoRA \emph{learns less} than full fine-tuning on hard domains (code, math) yet \emph{forgets less} of base abilities (HellaSwag, ARC-C, WinoGrande), while full FT induces \emph{\textbf{higher-rank}} perturbations (often $10$–$100\times$ typical LoRA ranks) \citep{biderman2024loralearnsforgets}. 
This echoes \emph{\textbf{classic catastrophic interference}} \citep{mccloskey1989catastrophic,french1999catastrophic} and \emph{\textbf{continual-learning views}} for LLMs \citep{wu2024continualLLMsSurvey,vu2022overcoming}. 
Read through the scaling law lens in Sec.~\ref{sec:scaling_law_grit} -- sets \emph{\textbf{how much}} forgetting to expect from $(D_{\text{ft}},N)$, while the quadratic analysis explains \emph{\textbf{why}} methods at the same budget diverge: update \emph{\textbf{geometry}} multiplies the baseline via adapter-restricted curvature exposure $\overline{\kappa}=\mathrm{tr}(P H_{\text{pt}} P)$ and by functions of effective rank $\Phi(r_{\text{eff}})$ and tail mass $\Psi(U_{\text{hi}}(\tau))$. 
Standard, \emph{\textbf{geometry-agnostic}} LoRA—first-order optimization in a fixed basis—tends to inflate $\overline{\kappa}$ and $U_{\text{hi}}(\tau)$ at fixed $(D_{\text{ft}},N)$ \citep{hu2021lora,biderman2024loralearnsforgets}, motivating \emph{\textbf{curvature-aware PEFT}} as the remedy.

\vspace{-0.5em}
\paragraph{\textbf{Implication.}}
If forgetting equals \emph{(data/model)} times \emph{(geometry/update)}, improving retention at fixed $(D_{\text{ft}},N)$ requires \emph{\textbf{shrinking the geometry factor}}.
This motivates \emph{\textbf{geometry-aware PEFT}}: estimate curvature in the adapter rank space and combine \emph{\textbf{natural-gradient}}/\emph{\textbf{K-FAC}} preconditioning with periodic \emph{\textbf{reprojection}} toward high-signal, low-interference eigendirections \citep{amari1998natural,martens2015kfac,ollivier2015riemannian}.
In short, \textbf{geometry-aware PEFT is needed}, and we propose \textbf{GRIT}: \emph{\textbf{Geometry-Aware PEFT with K-FAC Preconditioning, Fisher-Guided Reprojection, and Dynamic Rank Adaptation}}.

\section{GRIT: Geometry-Aware PEFT with K-FAC Preconditioning, Fisher-Guided Reprojection, and Dynamic Rank Adaptation}

\emph{GRIT} is a geometry-aware PEFT framework that turns LoRA-style updates \((\Delta W = BA)\) into a \emph{dynamic, curvature-aligned} procedure through three coupled steps:
\textbf{(i) curvature-aware preconditioning}—apply a K-FAC (Kronecker-Factored Approximate Curvature) \cite{martens2015kfac,grosse2016kronecker,amari1998natural} approximation to the Fisher \(F\) \emph{within the adapter subspace} to temper steps along sharp directions;
\textbf{(ii) spectrum-driven rank scheduling}—read the per-layer Fisher eigenspectrum and \emph{allocate rank where energy concentrates};
\textbf{(iii) neural reprojection}—periodically \emph{gate and align} the low-rank factors with the top-\(k\) eigenspace of \(F\) \((U_k U_k^\top)\), preserving task progress while discarding drift.
This loop repeats, so the adapter basis \emph{tracks high-signal, low-interference directions} dynamically.
The formal objective balancing task loss with \emph{curvature regularization} and \emph{reprojection constraints} appears in \textbf{Fig.~\ref{fig:grit_objective}}, the end-to-end flow in \textbf{Fig.~\ref{fig:grit_pipeline}}, and the overall effect—\emph{geometry-aligned}, \textbf{sparser}, and \textbf{more targeted} updates at the same memory budget—in \textbf{Fig.~\ref{fig:grit-minipage}}.

\begin{figure*}[h!]
  \centering
  \begin{tcolorbox}[
    enhanced,
    colback=white,
    colframe=black,
    boxrule=1pt,
    borderline={0.6pt}{2pt}{black},
    sharp corners,
    width=\textwidth,
  ]
  \vspace{-4mm}
  \[
  \min_{A \in \mathbb{R}^{d \times r},\; B \in \mathbb{R}^{r \times d}}\;
  \underbrace{L_{\text{task}}\!\bigl(W_0 + BA\bigr)}_{\text{(1) Task Loss}}
  \;+\;
  \lambda_{\text{K}}\,
  \underbrace{\bigl\lVert F^{\tfrac12}\!(BA)\bigr\rVert_{F}^{2}}_{\text{(2) Curvature Reg.}}
  \;+\;
  \lambda_{\text{R}}\,
  \underbrace{\bigl\lVert BA - U_k U_k^\top BA\bigr\rVert_{F}^{2}}_{\text{(3) Reprojection Reg.}}
  \]
  \vspace{-4mm}
  \end{tcolorbox}

  \vspace{-2mm}
  \caption{
\textbf{GRIT Objective: Curvature-Aware, Projection-Constrained Fine-Tuning.}
This loss balances task performance with geometric awareness and subspace filtering:
\textbf{(1) Task loss} $L_{\mathrm{task}}(W_{0}+BA)$ optimizes the instruction-tuning objective using the low-rank update $BA$;
\textbf{(2) Curvature regularization} $\lVert F^{1/2} BA \rVert_{F}^{2}$ penalizes updates in high-sensitivity regions defined by the Fisher matrix $F$, promoting safe adaptation;
\textbf{(3) Reprojection regularization} $\lVert BA - U_{k} U_{k}^{\top} BA \rVert_{F}^{2}$ encourages the update to remain within the subspace spanned by the top-$k$ eigenvectors of $F$, filtering noisy or low-impact directions.
Hyperparameters $\lambda_{\mathrm{K}}$ and $\lambda_{\mathrm{R}}$ control the curvature and reprojection terms.
}
  \label{fig:grit_objective}
  \vspace{-1.5em}
\end{figure*}

\subsection{Low-Rank Adaptation Setup}

Consider a transformer module with a linear projection parameterized by a weight matrix
\(W \in \mathbb{R}^{d_{\text{out}} \times d_{\text{in}}}\).
\emph{Full fine-tuning} updates all entries of \(W\), which is prohibitive at scale.
\emph{Low-rank adaptation} (LoRA/QLoRA) instead introduces a parameter-efficient update

\vspace{-2em}
\[
\Delta W \;=\; B A,
\qquad
B \in \mathbb{R}^{d_{\text{out}} \times r},\;
A \in \mathbb{R}^{r \times d_{\text{in}}},\;
r \ll \min(d_{\text{in}},d_{\text{out}}),
\]
\vspace{-2em}

so the effective weight is \(W' \;=\; W \;+\; \alpha\,\Delta W\) (with a small scaling \(\alpha\)).
This reduces trainable parameters from \(O(d_{\text{out}}d_{\text{in}})\) to \(O(r(d_{\text{in}}+d_{\text{out}}))\), preserving \emph{expressivity per parameter} while keeping memory/compute manageable.

\emph{\textbf{Caveat—geometry agnosticism.}}
Standard LoRA learns \(A,B\) in a \textbf{fixed} low-rank subspace using first-order updates, \emph{ignoring} loss curvature.
As a result, steps can over-expose sharp directions of the pretraining objective, amplifying interference.
\textbf{GRIT} removes this blind spot by making the low-rank subspace \emph{geometry-aware} via three components: \textbf{curvature-aware preconditioning}, \textbf{Fisher-guided reprojection}, and \textbf{dynamic rank adaptation}.


\subsection{K-FAC--Based Preconditioning}

\textbf{From gradients to \emph{natural} gradients.}
Raw stochastic gradients need not align with loss geometry.
The \emph{natural gradient} rescales \(\nabla_\theta \mathcal{L}\) by the inverse Fisher information matrix (FIM), yielding steepest descent under the KL metric \citep{amari1998natural}:
\vspace{-0.35em}
\[
\theta_{t+1}=\theta_t-\eta\,F^{-1}\nabla_{\theta}\mathcal{L}(\theta_t),
\qquad
F=\mathbb{E}\!\left[\nabla \log p(x;\theta)\,\nabla \log p(x;\theta)^{\!\top}\right].
\]
Directly forming/inverting \(F\) is infeasible for LLMs (quadratic storage, cubic inversion).

\textbf{K-FAC for layers, \emph{restricted} to rank space.}
K-FAC approximates the layerwise Fisher for \(y=Wx\) by a Kronecker product of second moments of \emph{input activations} and \emph{output gradients} \citep{martens2015kfac,grosse2016kronecker}:
\vspace{-0.35em}
\[
F_{\text{layer}}\approx \Sigma_g\otimes\Sigma_a,
\qquad
\Sigma_a=\mathbb{E}[xx^{\top}],
\quad
\Sigma_g=\mathbb{E}[gg^{\top}],
\quad
g\equiv\tfrac{\partial \mathcal{L}}{\partial y}.
\]
Then the \emph{natural-gradient} preconditioned update admits
\[
\nabla W_{\text{nat}}\approx \Sigma_g^{-1}\,\nabla W\,\Sigma_a^{-1},
\]
avoiding explicit inversion of \(F_{\text{layer}}\).

\textbf{Rank-space K-FAC for LoRA.}
For \(\Delta W=BA\), GRIT applies K-FAC \emph{within} the rank-\(r\) subspace spanned by \(A,B\).
Let \(x\in\mathbb{R}^{d_{\text{in}}}\) be activations and \(g\in\mathbb{R}^{d_{\text{out}}}\) the backpropagated gradients.
Define rank-projected statistics
\vspace{-0.35em}
\[
a_r=Ax\in\mathbb{R}^{r},
\qquad
g_r=B^{\top}g\in\mathbb{R}^{r},
\]
\vspace{-0.35em}
and rank-space covariances
\vspace{-0.35em}
\[
\Sigma_a^{(r)}=\mathbb{E}[a_ra_r^{\top}]\in\mathbb{R}^{r\times r},
\qquad
\Sigma_g^{(r)}=\mathbb{E}[g_rg_r^{\top}]\in\mathbb{R}^{r\times r}.
\]
Under the K-FAC independence approximation, the Fisher restricted to the LoRA subspace factorizes as
\vspace{-0.35em}
\[
F_{\text{rank}}\approx \Sigma_g^{(r)}\otimes \Sigma_a^{(r)}.
\]
\vspace{-0.35em}
Consequently,
\vspace{-0.35em}
\[
\nabla (\Delta W)_{\text{nat}}\approx \Sigma_g^{(r)\,-1}\,\nabla(\Delta W)\,\Sigma_a^{(r)\,-1},
\]
which decouples into factor-wise updates (for \(\Delta W{=}BA\)):
\vspace{-0.45em}
\[
\boxed{
\begin{aligned}
\nabla B \leftarrow \nabla B\, \Sigma_g^{(r)\,-1},
\qquad
\nabla A \leftarrow \Sigma_a^{(r)\,-1}\, \nabla A.
\end{aligned}}
\]
\vspace{-0.35em}
Intuitively, \(\Sigma_g^{(r)\,-1}\) suppresses steps along \emph{high-curvature output directions}, while \(\Sigma_a^{(r)\,-1}\) removes \emph{input-scale anisotropy} in the adapter subspace---yielding \textbf{curvature-aligned}, \emph{scale-invariant} updates.

\textbf{Practicalities.}
For stability, GRIT uses (i) \emph{damping}:
\(\tilde{\Sigma}_a^{(r)}=\Sigma_a^{(r)}+\lambda_a I\),
\(\tilde{\Sigma}_g^{(r)}=\Sigma_g^{(r)}+\lambda_g I\);
(ii) \emph{streaming} (EMA) estimates with burn-in; and
(iii) Cholesky solves on \(r\times r\) matrices (\(r\ll d_{\text{in}},d_{\text{out}}\)).
Statistics are per layer and can be cached/offloaded.
In GRIT, K-FAC is the \textbf{first} step; \emph{Fisher-guided reprojection} and \emph{dynamic rank scheduling} follow (see \textbf{Fig.~\ref{fig:grit_objective}} and \textbf{Fig.~\ref{fig:grit_pipeline}}).

\textbf{Takeaway.}
Rank-space K-FAC gives a \emph{lightweight} natural-gradient proxy, retaining key \textbf{second-order} benefits while scaling to billion-parameter LLMs \citep{amari1998natural,martens2015kfac,grosse2016kronecker}.
\vspace{-0.25em}

\begin{algorithm}[!ht]\scriptsize
\caption{GRIT training loop}
\label{alg:grit}
\begin{tcolorbox}[enhanced, colback=black!5, colframe=black, boxrule=0.3pt, left=4pt, right=4pt, top=4pt, bottom=4pt]
\begin{algorithmic}[1]
\Require Pretrained weights $W_0$; LoRA rank $r$; LoRA scaling $\alpha$; data stream $\mathcal{D}$; learning rate $\eta$; damping $\lambda$; frequencies (\texttt{kfac\_upd}, \texttt{reproj\_freq}); thresholds (\texttt{kfac\_min}, $\tau$, \texttt{min\_rank}); flags (\texttt{ng\_warmup}, \texttt{reproj\_warmup}, \texttt{use\_two\_sided}, \texttt{rank\_adapt})
\State Initialize LoRA factors $\{A_\ell, B_\ell\}$; set $A_{\text{cov}}\gets I$, $G_{\text{cov}}\gets I$; $\textit{inv\_ready}\gets\textbf{False}$; $n_{\text{cov}}\gets 0$; $\textit{step}\gets 0$
\For{each minibatch $(x,y)\in\mathcal{D}$}
  \State Forward with $W' = W_0 + \alpha\,\sum_\ell B_\ell A_\ell$; compute loss $\mathcal{L}$
  \State Backprop to obtain raw gradients $\nabla A_\ell,\, \nabla B_\ell$ and per-layer tensors $X,\,\delta Y$
  \If{$\textit{step} \ge$ \texttt{ng\_warmup}}
    \If{$\textit{step}$ mod \texttt{kfac\_upd} $= 0$}
      \State $a_r \gets X A_\ell^\top$; $g_r \gets \delta Y B_\ell$ \hfill\Comment{rank-space stats}
      \State Accumulate: $A_{\text{cov}} \gets A_{\text{cov}} + a_r a_r^\top$; $G_{\text{cov}} \gets G_{\text{cov}} + g_r g_r^\top$; $n_{\text{cov}} \gets n_{\text{cov}}{+}1$
      \If{$n_{\text{cov}} \ge$ \texttt{kfac\_min}}
        \State Compute $(A_{\text{cov}}{+}\lambda I)^{-1}$ and $(G_{\text{cov}}{+}\lambda I)^{-1}$; $\textit{inv\_ready}\gets\textbf{True}$
      \EndIf
    \EndIf
    \If{$\textit{inv\_ready}$}
      \State Natural gradient: $\nabla B_\ell \gets \nabla B_\ell G_{\text{cov}}^{-1}$; $\nabla A_\ell \gets A_{\text{cov}}^{-1}\nabla A_\ell$
    \EndIf
  \EndIf
  \State Optimizer step on $\{A_\ell, B_\ell\}$; freeze $W_0$
  \If{$\textit{step} \ge$ \texttt{reproj\_warmup} \textbf{and} $\textit{step}$ mod \texttt{reproj\_freq} $= 0$}
    \State Eigendecompose $A_{\text{cov}} = U_A\Lambda_A U_A^\top$
    \If{\texttt{use\_two\_sided} \textbf{and} $\textit{inv\_ready}$}
      \State Eigendecompose $G_{\text{cov}} = U_G\Lambda_G U_G^\top$
    \EndIf
    \If{\texttt{rank\_adapt}}
      \State $k \gets \min\big\{ j\,\big|\, \sum_{i=1}^j \lambda_i \big/ \sum_{i=1}^r \lambda_i \ge \tau \big\}$; $k \gets \max(k,\texttt{min\_rank})$
    \Else
      \State $k \gets$ \texttt{reproj\_k}
    \EndIf
    \State $A_\ell \gets U_A^{(k)}U_A^{(k)\top}A_\ell$
    \State $B_\ell \gets \begin{cases} B_\ell U_G^{(k)}U_G^{(k)\top}, & \text{if two-sided} 
    \\ B_\ell U_A^{(k)}U_A^{(k)\top}, & \text{otherwise} \end{cases}$
  \EndIf
  \State $\textit{step} \gets \textit{step}{+}1$
\EndFor
\State \Return $W' = W_0 + \alpha\,\sum_\ell B_\ell A_\ell$
\end{algorithmic}
\end{tcolorbox}
\end{algorithm}

\subsection{Neural Reprojection}

\textbf{Motivation.}
Preconditioning corrects \emph{step directions} but leaves the \emph{update subspace} fixed.
Over training, the LoRA subspace spanned by \(A\) and \(B\) (rank \(r\)) can drift, accumulate redundancy, or misalign with \textbf{high-curvature} directions—wasting gradient signal and inflating interference.
\emph{Neural reprojection} remedies this by \textbf{reshaping} the subspace itself to track informative curvature.

\textbf{Curvature-aligned subspace.}
Let the rank-space covariances (Sec.~K-FAC) be
\[
\Sigma_{a}^{(r)}=\mathbb{E}[\,a_r a_r^\top\,],\qquad
\Sigma_{g}^{(r)}=\mathbb{E}[\,g_r g_r^\top\,]\in\mathbb{R}^{r\times r},
\]
with eigendecompositions
\[
\Sigma_{a}^{(r)}=U_A\Lambda_A U_A^\top,\qquad
\Sigma_{g}^{(r)}=U_G\Lambda_G U_G^\top,
\]
where \(U_A,U_G\in\mathbb{R}^{r\times r}\) are orthogonal and \(\Lambda_A,\Lambda_G\succeq 0\) carry \emph{curvature energy} per direction.
Let \(U_A^{(k)}\) and \(U_G^{(k)}\) collect the top-\(k\) eigenvectors (by eigenvalue), and define projectors
\[
P_A \;=\; U_A^{(k)}(U_A^{(k)})^\top,\qquad
P_G \;=\; U_G^{(k)}(U_G^{(k)})^\top\;\;\in\mathbb{R}^{r\times r}.
\]
We \textbf{reproject} the LoRA factors onto these dominant eigenspaces:
\[
\boxed{\;A \;\leftarrow\; P_A\,A,\qquad B \;\leftarrow\; B\,P_G\;}
\]
so the low-rank update $\Delta W=BA$ remains in a \emph{rank-$r$} space with a \textbf{curvature-aligned} basis.

\textbf{Gating, scheduling, and stability.}
To avoid premature rotations, we enable the \(G\)-side projection only after \(\Sigma_{g}^{(r)}\) has accumulated at least \(N_{\min}\) effective samples; otherwise, we \emph{fallback} to \(U_A^{(k)}\) for both sides in the first epochs.
Reprojection runs at a fixed frequency (every \(T\) steps) or adaptively when the spectrum mass ratio \(\sum_{i=1}^k \Lambda_{(\cdot),i}/\sum_{i=1}^r \Lambda_{(\cdot),i}\) crosses a threshold.
Since \(\Sigma_{a}^{(r)}\) and \(\Sigma_{g}^{(r)}\) are \(r\times r\), eigendecompositions are \emph{cheap}, and \(U_A^{(k)},U_G^{(k)}\) are cached between steps.
For numerical robustness, we use \(\tilde{\Sigma}_{(\cdot)}^{(r)}=\Sigma_{(\cdot)}^{(r)}+\lambda I\) (damping), warm up statistics before projection, and interpolate updates if needed:
\[
A \leftarrow (1-\gamma)A + \gamma\,P_AA,\qquad
B \leftarrow (1-\gamma)B + \gamma\,B P_G,\quad \gamma\in[0,1].
\]

\textbf{Effect.}
Neural reprojection \emph{removes low-energy directions}, suppresses noise, and \textbf{rotates} the LoRA subspace toward \emph{high-signal, low-interference} eigendirections.
Unlike pure preconditioning (which only rescales steps), reprojection \textbf{evolves the basis} so that adaptation occurs where curvature indicates capacity is most valuable—improving \textbf{stability} and \textbf{parameter efficiency} at fixed rank.

\begin{figure*}[t!]
    \centering
    \includegraphics[width=\textwidth]{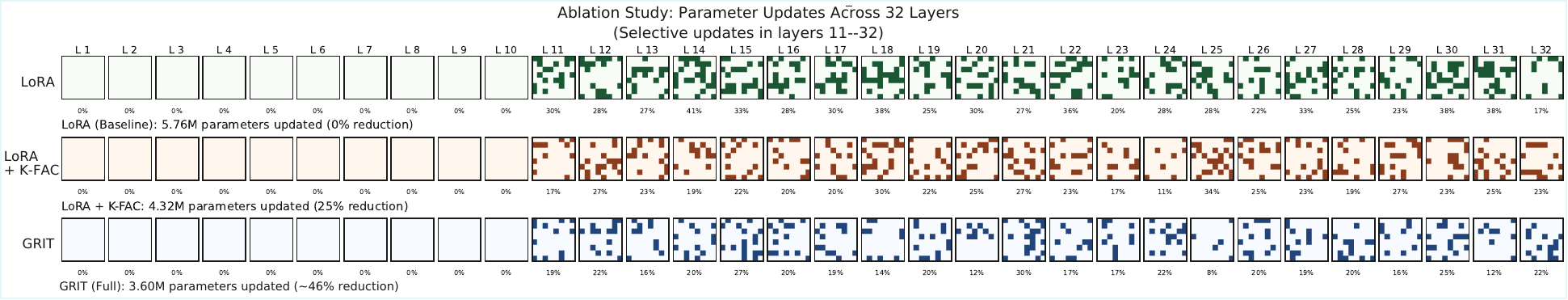}
    \vspace{-1em}
        \caption{\textbf{Ablation—\emph{parameter update patterns} across \textbf{LLaMA} (32 layers).}
Each \textbf{8$\times$8} mini-grid depicts one layer; \emph{colored cells} mark updated parameters.
We freeze \textbf{layers 1–10} because these early layers encode \emph{task-agnostic representations}, while \textbf{layers 10–15} serve as \emph{transition} to task-specific features and \textbf{layers 16–32} provide \emph{refinement} \citep{zhao2024layerlayeruncoveringmultitask}; this follows the PEFT practice of adapting only the \textbf{later layers} (\textbf{11–32}).
Within adapted layers, the mean update density is \textbf{LoRA} \emph{30\%} (green), \textbf{LoRA+K-FAC} \emph{22.5\%} (orange), and \textbf{GRIT} \emph{18.75\%} (blue).
\emph{Note:} these are \textbf{per-layer} densities; the \textbf{overall} update fraction is smaller because early layers are \emph{not} adapted.
Totals (annotated under each row): \textbf{LoRA} \emph{5.76M} params (0\% reduction), \textbf{LoRA+K-FAC} \emph{4.32M} (25\% reduction), \textbf{GRIT} \emph{3.60M} (37.5\% reduction).
Counts assume each adapted layer updates \textbf{4} weight matrices of \textbf{65{,}536} parameters.
\emph{Takeaway:} \textbf{curvature-aware} preconditioning and \textbf{reprojection} \emph{reduce update density} vs. standard LoRA while retaining capacity.}

    \label{fig:ablation_study}
    \vspace{-1.5em}
\end{figure*}

\subsection{Dynamic Rank Adaptation}

\textbf{Why adapt rank?}
Reprojection aligns the subspace, but a \emph{fixed} rank \(r\) can still \textbf{underfit} (too few directions) or \textbf{overfit} (retain redundant, low-energy directions). GRIT therefore lets the \emph{effective} rank track the spectrum.

\textbf{Energy-based rule.}
Let \(\lambda_1\ge\cdots\ge\lambda_r\ge 0\) be eigenvalues of a rank-space covariance (activation- or Fisher-side; cf.\ Secs.~\ref{sec:diagnosis-geometry}, K-FAC). Define the smallest \(k\) capturing energy fraction \(\tau\in(0,1]\):
\vspace{-0.45em}
\[
k=\min\Big\{\,j\ \Big|\ \frac{\sum_{i=1}^{j}\lambda_i}{\sum_{i=1}^{r}\lambda_i}\ge\tau\,\Big\}.
\]
Here “energy” is cumulative variance/mass in leading eigendirections.

\textbf{Constraints and gating.}
We bound
\vspace{-0.45em}
\[
k\in[\,\texttt{min\_rank},\,r\,],
\]
where \texttt{min\_rank} prevents collapse and \(r\) is the max LoRA rank at initialization. To avoid early misestimation, we use a warmup gate: updates to \(k\) start only after sufficient samples for stable spectra (e.g., EMA/Fisher burn-in).

\textbf{How adaptation is realized.}
Rank adaptation is implemented through the reprojection operators. With \(U^{(k)}\) the top-\(k\) eigenvectors, projectors
\(
P_A=U_A^{(k)}(U_A^{(k)})^\top,\;
P_G=U_G^{(k)}(U_G^{(k)})^\top
\)
suppress low-energy directions during
\vspace{-0.45em}
\[
A\leftarrow P_AA,
\qquad
B\leftarrow BP_G.
\]
No hard masking or tensor resizing is required—the directions remain stored and can re-enter if their eigenvalues grow.

\textbf{Update incorporation.}
After reprojection and rank selection,
\vspace{-0.45em}
\[
\Delta W_{\text{new}}=B_{\text{new}}A_{\text{new}},
\qquad
W'=W+\Delta W_{\text{new}},
\]
yielding curvature-aware, rank-adaptive updates under a stable adapter parameterization.

\textbf{Implementation summary.}
Algorithm~\ref{alg:grit} maps directly to our system: rank-space covariances via autograd hooks; sample-gated, damped K-FAC inverses; natural-gradient preconditioning before the optimizer step; and periodic Fisher-guided reprojection with dynamic rank. cf. Apndx.~\ref{app:background} and  Apndx.~\ref{app:derivations}.
\vspace{-0.25em}

\vspace{-1em}
\section{Experiments and Results}
\label{sec:experiments}

\paragraph{Setup \& Datasets.}
Unless stated, we fine\hyp{}tune \textbf{Llama\,3.2–3B} and \textbf{Llama\,3.1–8B} with \emph{4\hyp{}bit NF4} quantization and \texttt{bf16} compute. \textbf{GRIT} reuses the \textbf{QLoRA} data pipeline for an apples\hyp{}to\hyp{}apples comparison; \emph{reprojection} is gated by curvature\hyp{}sample thresholds, and \emph{dynamic rank} follows a cumulative\hyp{}energy rule. Seeds, optimizers, and schedules appear in Appx.~\ref{app:train}. Evaluation spans five benchmarks: \emph{Alpaca}~\citep{wang2023selfinstruct} (52k instruction–response pairs), \emph{Dolly 15k}~\citep{dolly15k} (15k human\hyp{}written prompts), \emph{BoolQ}~\citep{clark2019boolq} (yes/no over Wikipedia), \emph{QNLI} from GLUE~\citep{wang2019glue} (sentence\hyp{}pair entailment), and \emph{GSM8K}~\citep{cobbe2021gsm8k} (grade\hyp{}school math reasoning).

\paragraph{Baselines.}
We compare GRIT to strong PEFT baselines: \textbf{LoRA}~\citep{hu2021lora}, \textbf{QLoRA}~\citep{dettmers2023qlora}, \textbf{IA$^3$}~\citep{liu2022ia3} (per-module gating in lieu of rank updates), \textbf{DoRA}~\citep{liu2024dora} (direction–magnitude decomposition for stabler adapters), and \textbf{orthogonal-LoRA} (control basis orthogonality). To isolate the role of curvature modeling apart from subspace alignment, we include factored second-order \textbf{Shampoo}~\citep{anil2021scalable} \emph{without} Fisher-guided reprojection as an optimizer control. This set spans where capacity is injected (ranks vs.\ gates), how it is constrained (quantization, decomposition, orthogonality), and how updates are preconditioned (first- vs.\ second-order), providing a backdrop for GRIT’s geometry-aware contributions.

\begin{table*}[ht!]
\centering
\caption{\textbf{Extended baselines on \emph{Llama-3.2 3B}}. Rows are grouped by dataset (blue/gray bands). Each metric cell reports the \emph{absolute} score followed by a \emph{relative delta}: for all columns except \textbf{GRIT}, the delta is computed \emph{vs.\ GRIT} \((\uparrow=\text{higher than GRIT},\ \downarrow=\text{lower than GRIT})\); in the \textbf{GRIT} column, the delta is computed \emph{vs.\ LoRA} to show GRIT’s improvement or drop. For ROUGE/BLEU/BERTScore/Accuracy/Precision/Recall/F1, larger is better; arrows indicate better/worse. \textbf{Bold} marks the best value within the row. The ``\# Params Trained'' rows report the \emph{absolute} number of trainable parameters for each method and, in parentheses, the \% change \emph{vs.\ LoRA} (lower is better). This layout makes quality deltas relative to GRIT explicit while exposing GRIT’s parameter savings over LoRA.}
\label{tab:main_results_llama32_clean}
\adjustbox{width=0.86\textwidth,center}{
\small
\renewcommand{\arraystretch}{1.18}
\begin{tabular}{|l|c|c|c|c|c|c|c|}
\hline
\rowcolor{headerblue}
{\color{white}\textbf{Model: Llama-3.2 (3B)}} &
{\color{white}\textbf{LoRA}} &
{\color{white}\textbf{QLoRA}} &
{\color{white}\textbf{GRIT (vs.\ LoRA)}} &
{\color{white}\textbf{Orthogonal-LoRA}} &
{\color{white}\textbf{IA$^3$}} &
{\color{white}\textbf{DoRA/Eff-FT}} &
{\color{white}\textbf{Shampoo}} \\
\hline

\rowcolor{bandblue}
\multicolumn{8}{|l|}{\textbf{ALPACA}} \\ \hline
ROUGE-1   & 0.1852 & 0.1292 & \textbf{0.1844} (\(\downarrow 0.8\%\)) & 0.1870 (\(\uparrow 1.4\%\)) & 0.1680 (\(\downarrow 8.9\%\)) & \textbf{0.1915} (\(\uparrow 3.9\%\)) & 0.1885 (\(\uparrow 2.2\%\)) \\
ROUGE-2   & 0.0825 & 0.0562 & \textbf{0.0818} (\(\downarrow 0.8\%\)) & 0.0836 (\(\uparrow 2.2\%\)) & 0.0710 (\(\downarrow 13.2\%\)) & \textbf{0.0868} (\(\uparrow 6.1\%\)) & 0.0850 (\(\uparrow 3.9\%\)) \\
ROUGE-L   & 0.1426 & 0.0983 & \textbf{0.1425} (\(\downarrow 0.1\%\)) & 0.1440 (\(\uparrow 1.1\%\)) & 0.1310 (\(\downarrow 8.1\%\)) & \textbf{0.1478} (\(\uparrow 3.7\%\)) & 0.1456 (\(\uparrow 2.2\%\)) \\
BLEU      & 0.0443 & 0.0235 & \textbf{0.0430} (\(\downarrow 2.9\%\)) & 0.0451 (\(\uparrow 4.9\%\)) & 0.0380 (\(\downarrow 11.6\%\)) & \textbf{0.0472} (\(\uparrow 9.8\%\)) & 0.0461 (\(\uparrow 7.2\%\)) \\
BERTScore & 0.8343 & 0.7948 & \textbf{0.8354} (\(\uparrow 0.1\%\)) & 0.8350 (\(\downarrow 0.0\%\)) & 0.8230 (\(\downarrow 1.5\%\)) & 0.8349 (\(\downarrow 0.1\%\)) & 0.8351 (\(\downarrow 0.0\%\)) \\
\# Params Trained
          & 24.31M & 24.31M & \textbf{8.45M} (65.3\% \(\downarrow\)) & 24.31M (0.0\%) & \textbf{2.10M} (91.4\% \(\downarrow\)) & 24.31M (0.0\%) & 24.31M (0.0\%) \\
\hline

\rowcolor{bandgray}
\multicolumn{8}{|l|}{\textbf{Dolly-15k}} \\ \hline
ROUGE-1   & 0.1733 & 0.1108 & \textbf{0.1976} (\(\uparrow 14.0\%\)) & 0.1795 (\(\downarrow 9.2\%\)) & 0.1602 (\(\downarrow 18.9\%\)) & 0.1931 (\(\downarrow 2.3\%\)) & 0.1864 (\(\downarrow 5.7\%\)) \\
ROUGE-2   & 0.0824 & 0.0519 & \textbf{0.0994} (\(\uparrow 20.7\%\)) & 0.0856 (\(\downarrow 13.9\%\)) & 0.0718 (\(\downarrow 27.8\%\)) & \textbf{0.1006} (\(\uparrow 1.2\%\)) & 0.0941 (\(\downarrow 5.3\%\)) \\
ROUGE-L   & 0.1368 & 0.0884 & \textbf{0.1568} (\(\uparrow 14.6\%\)) & 0.1402 (\(\downarrow 10.6\%\)) & 0.1243 (\(\downarrow 20.7\%\)) & 0.1541 (\(\downarrow 1.7\%\)) & 0.1486 (\(\downarrow 5.2\%\)) \\
BLEU      & 0.0533 & 0.0297 & \textbf{0.0560} (\(\uparrow 5.1\%\)) & 0.0542 (\(\downarrow 3.2\%\)) & 0.0451 (\(\downarrow 19.5\%\)) & \textbf{0.0574} (\(\uparrow 2.5\%\)) & 0.0566 (\(\uparrow 1.1\%\)) \\
BERTScore & 0.8295 & 0.8005 & \textbf{0.8344} (\(\uparrow 0.6\%\)) & 0.8311 (\(\downarrow 0.4\%\)) & 0.8192 (\(\downarrow 1.8\%\)) & 0.8335 (\(\downarrow 0.1\%\)) & 0.8322 (\(\downarrow 0.3\%\)) \\
\# Params Trained
          & 24.31M & 24.31M & \textbf{17.01M} (30.0\% \(\downarrow\)) & 24.31M (0.0\%) & \textbf{2.10M} (91.4\% \(\downarrow\)) & 24.31M (0.0\%) & 24.31M (0.0\%) \\
\hline

\rowcolor{bandblue}
\multicolumn{8}{|l|}{\textbf{GSM8K}} \\ \hline
ROUGE-1   & 0.5582 & 0.5518 & 0.5532 (\(\downarrow 0.9\%\)) & 0.5594 (\(\uparrow 1.1\%\)) & 0.5401 (\(\downarrow 2.4\%\)) & \textbf{0.5610} (\(\uparrow 1.4\%\)) & 0.5601 (\(\uparrow 1.2\%\)) \\
ROUGE-2   & 0.3236 & 0.3197 & 0.3173 (\(\downarrow 1.9\%\)) & 0.3249 (\(\uparrow 2.4\%\)) & 0.3050 (\(\downarrow 3.9\%\)) & \textbf{0.3268} (\(\uparrow 3.0\%\)) & 0.3257 (\(\uparrow 2.6\%\)) \\
ROUGE-L   & 0.5228 & 0.5169 & 0.5167 (\(\downarrow 1.2\%\)) & 0.5233 (\(\uparrow 1.3\%\)) & 0.5078 (\(\downarrow 1.7\%\)) & \textbf{0.5241} (\(\uparrow 1.4\%\)) & 0.5237 (\(\uparrow 1.4\%\)) \\
Accuracy  & 0.3935 & 0.3836 & 0.3867 (\(\downarrow 1.7\%\)) & 0.3962 (\(\uparrow 2.5\%\)) & 0.3564 (\(\downarrow 7.8\%\)) & \textbf{0.3991} (\(\uparrow 3.2\%\)) & 0.3978 (\(\uparrow 2.9\%\)) \\
\# Params Trained
          & 24.31M & 24.31M & \textbf{15.30M} (37.1\% \(\downarrow\)) & 24.31M (0.0\%) & \textbf{2.10M} (91.4\% \(\downarrow\)) & 24.31M (0.0\%) & 24.31M (0.0\%) \\
\hline

\rowcolor{bandgray}
\multicolumn{8}{|l|}{\textbf{QNLI}} \\ \hline
Accuracy  & 0.8938 & 0.8885 & \textbf{0.9053} (\(\uparrow 1.3\%\)) & 0.8984 (\(\downarrow 0.8\%\)) & 0.8820 (\(\downarrow 2.6\%\)) & 0.9026 (\(\downarrow 0.3\%\)) & 0.9001 (\(\downarrow 0.6\%\)) \\
Precision & 0.8939 & 0.8971 & \textbf{0.9059} (\(\uparrow 1.3\%\)) & 0.8991 (\(\downarrow 0.8\%\)) & 0.8842 (\(\downarrow 2.4\%\)) & 0.9037 (\(\downarrow 0.2\%\)) & 0.9010 (\(\downarrow 0.5\%\)) \\
Recall    & 0.8939 & 0.8893 & \textbf{0.9055} (\(\uparrow 1.3\%\)) & 0.8977 (\(\downarrow 0.9\%\)) & 0.8805 (\(\downarrow 2.8\%\)) & 0.9021 (\(\downarrow 0.4\%\)) & 0.8990 (\(\downarrow 0.7\%\)) \\
F1        & 0.8938 & 0.8880 & \textbf{0.9052} (\(\uparrow 1.3\%\)) & 0.8981 (\(\downarrow 0.8\%\)) & 0.8810 (\(\downarrow 2.7\%\)) & 0.9029 (\(\downarrow 0.3\%\)) & 0.8996 (\(\downarrow 0.6\%\)) \\
\# Params Trained
          & 24.31M & 24.31M & \textbf{7.75M} (68.1\% \(\downarrow\)) & 24.31M (0.0\%) & \textbf{1.20M} (95.1\% \(\downarrow\)) & 24.31M (0.0\%) & 24.31M (0.0\%) \\
\hline

\rowcolor{bandblue}
\multicolumn{8}{|l|}{\textbf{BoolQ}} \\ \hline
Accuracy  & 0.7834 & 0.7525 & \textbf{0.7749} (\(\downarrow 1.1\%\)) & 0.7860 (\(\uparrow 1.4\%\)) & 0.7402 (\(\downarrow 4.5\%\)) & 0.7815 (\(\uparrow 0.9\%\)) & 0.7851 (\(\uparrow 1.3\%\)) \\
Precision & 0.7982 & 0.7491 & \textbf{0.7908} (\(\downarrow 0.9\%\)) & 0.8010 (\(\uparrow 1.3\%\)) & 0.7385 (\(\downarrow 6.6\%\)) & 0.7972 (\(\uparrow 0.8\%\)) & 0.7994 (\(\uparrow 1.1\%\)) \\
Recall    & 0.8720 & 0.9050 & \textbf{0.8671} (\(\downarrow 0.6\%\)) & 0.8692 (\(\uparrow 0.2\%\)) & 0.8204 (\(\downarrow 5.4\%\)) & 0.8701 (\(\uparrow 0.3\%\)) & 0.8710 (\(\uparrow 0.4\%\)) \\
F1        & 0.8335 & 0.8197 & \textbf{0.8272} (\(\downarrow 0.8\%\)) & 0.8357 (\(\uparrow 1.0\%\)) & 0.7796 (\(\downarrow 5.8\%\)) & 0.8324 (\(\uparrow 0.6\%\)) & 0.8348 (\(\uparrow 0.9\%\)) \\
\# Params Trained
          & 24.31M & 24.31M & \textbf{15.03M} (38.2\% \(\downarrow\)) & 24.31M (0.0\%) & \textbf{1.60M} (93.4\% \(\downarrow\)) & 24.31M (0.0\%) & 24.31M (0.0\%) \\
\hline

\end{tabular}
}
\vspace{3pt}
\vspace{-1.5em}
\end{table*}
\vspace{-1em}

\paragraph{Performance.}
\textbf{GRIT matches or exceeds quality while sharply cutting trainable parameters.}
Across five benchmarks and two model sizes, GRIT attains parity or small gains over strong PEFT baselines while \emph{substantially} reducing update footprint (\textbf{Table~\ref{tab:main_results_llama32_clean}}).
On \textbf{Alpaca}, GRIT is within ${<}1\%$ of the best ROUGE-1/2/L and BERTScore at both scales, yet trains only \textbf{8.45M} params on 3B (\(\mathbf{65.3\%}\downarrow\) vs.\ LoRA/QLoRA) and \textbf{19.27M} on 8B (\(\mathbf{77.0\%}\downarrow\)).
On \textbf{Dolly-15k}, GRIT attains the top ROUGE-1/2/L and BLEU for 3B, with a \textbf{30.0\%} reduction in trained params (17.01M vs.\ 24.31M), and remains competitive on 8B while trimming \textbf{35.2\%}–\textbf{54.5\%}.
On \textbf{GSM8K} (reasoning), GRIT ties or trails by \(<\!1.5\%\) on sequence metrics for 3B, and \emph{wins accuracy} on 8B (\textbf{0.6619}, best-in-block) with a \textbf{27.8\%} reduction (60.5M).
On \textbf{QNLI} (NLI), GRIT is best or within \(<\!1.0\%\) across Accuracy/Precision/Recall/F1 at 3B while cutting params by \textbf{68.1\%} (7.75M), and remains near the block leader on 8B with \textbf{64.9\%}–\textbf{80.0\%} savings.
On \textbf{BoolQ}, GRIT is at or near the top across metrics with \textbf{38.2\%} (3B) and \textbf{26.6\%} (8B) fewer trained parameters.
An extended ablation on Llama-3.2 3B (Table~\ref{tab:main_results_llama32_clean}) confirms the trend against stronger PEFT/optimizer controls: orthogonal-LoRA, IA$^3$, DoRA/Eff-FT, and Shampoo show modest metric fluctuations around GRIT, but none close the parameter-efficiency gap.

\textbf{Where the savings come from.}
GRIT’s dynamic, geometry-aware allocation adapts the \emph{effective} rank and concentrates updates in informative layers, yielding task-dependent compression rather than a fixed adapter budget.
This appears as consistent per-task reductions in the “\# Params Trained’’ column of \textbf{Table~\ref{tab:main_results_llama32_clean}} (e.g., \(\mathbf{65.3\%}\downarrow\)–\(\mathbf{77.0\%}\downarrow\) on Alpaca; \(\mathbf{30.0\%}\downarrow\)–\(\mathbf{54.5\%}\downarrow\) on Dolly-15k; \(\mathbf{26.6\%}\downarrow\)–\(\mathbf{80.0\%}\downarrow\) on BoolQ/QNLI) while maintaining block-best or near-best quality.
The update-pattern visualization in \textbf{Fig.~\ref{fig:ablation_study}} shows the same story at the layer level: GRIT suppresses early layers and densifies updates selectively in middle-to-late blocks, delivering \emph{sparser, better-aimed} updates at the same memory budget.
Overall, \textbf{GRIT offers a favorable quality–efficiency trade-off} relative to LoRA/QLoRA and stronger PEFT baselines, with \emph{consistent parameter savings and competitive accuracy} across instruction following, classification, and math reasoning.

\section{Scaling Laws for Forgetting: LoRA vs.\ GRIT}
\label{sec:scaling_law_grit}

\textbf{Fine-tuning LLMs} typically induces \emph{\textbf{catastrophic forgetting}}—a drift away from the pretraining distribution that erodes general knowledge. In PEFT methods such as LoRA, we quantify forgetting by the increase in \emph{\textbf{pretraining loss}} \(L_{pt}\) after fine-tuning. Following \citet{bethune2022scaling}, forgetting obeys a \emph{\textbf{power law}} in the fine-tuning data volume \(D_{\mathrm{ft}}\) (number of unique fine-tuning tokens) and model size \(N\) (number of parameters):

\vspace{-1em}
\begin{equation*}
\refstepcounter{equation}\label{eq:baseline-scaling}
\boxed{\displaystyle
L_{pt} \;=\; L_{pt}^{0} \;+\; A\,\frac{D_{ft}^{\beta}}{N^{\alpha}} \;+\; E}
\end{equation*}
where \(L_{pt}^{0}\) is the original pretraining loss and \(A,\alpha,\beta,E\) are dataset- and model-specific constants. \textbf{This captures a key trade-off:} increasing \(D_{ft}\) amplifies forgetting (\(D_{ft}^{\beta}\)), while larger models forget less via \(N^{-\alpha}\), effectively \emph{diluting per-update distortion across parameters}.

\textbf{From LoRA to GRIT: a geometry multiplier.} Standard LoRA optimizes in a fixed low-rank basis, which can inadvertently place updates into high-curvature directions that amplify drift. GRIT adds \emph{curvature-aware preconditioning} and \emph{Fisher-guided reprojection}, which (i) shrink steps along sharp modes and (ii) rotate the low-rank basis toward informative eigendirections. We capture this effect by introducing an \emph{effective capacity multiplier} \(\Xi_{\mathrm{GRIT}}>1\) in the denominator:

\vspace{-1.7em}
\begin{equation*}
\refstepcounter{equation}\label{eq:grit-scaling}
\boxed{\displaystyle
L_{pt}^{\mathrm{GRIT}}
\;=\;
L_{pt}^{0}
\;+\;
A\,\frac{D_{ft}^{\beta}}{\bigl(\Xi_{\mathrm{GRIT}}\,N\bigr)^{\alpha}}
\;+\;
E,
\qquad
\Xi_{\mathrm{GRIT}}
=
(1+\gamma_{r}r_{\mathrm{eff}})
(1+\gamma_{a}\rho_{\mathrm{align}})
(1+\gamma_{p}\pi_{\mathrm{proj}})}
\end{equation*}
\vspace{-1.6em}

We parameterize \(\Xi_{\mathrm{GRIT}}\) by measurable geometry: \(r_{\mathrm{eff}}\) is the adapter’s \emph{effective rank} (usable capacity), \(\rho_{\mathrm{align}}\in[0,1]\) measures alignment to the Fisher top-\(k\) subspace (curvature alignment), and \(\pi_{\mathrm{proj}}\in[0,1]\) is the spectral mass retained after reprojection (signal concentration). The scalings \(\gamma_{\{\cdot\}}\!\ge\!0\) are fitted from runs that vary \(D_{ft}\), rank, and reprojection frequency (full derivation in Appendix~\cref{appndx:law_derivation}). For background on natural-gradient/K-FAC curvature handling, see \citet{amari1998natural,martens2015kfac}; learn--forget trade-offs in PEFT are discussed in \citet{bethune2022scaling,biderman2024loralearnsforgets}.

\textbf{How to read the law.} At fixed \((D_{ft},N)\), improving any of \(\{r_{\mathrm{eff}},\rho_{\mathrm{align}},\pi_{\mathrm{proj}}\}\) increases \(\Xi_{\mathrm{GRIT}}\) and lowers \(L_{pt}^{\mathrm{GRIT}}\). Practically, we recommend reporting \emph{Fisher spectra}, \emph{effective ranks}, and \emph{alignment proxies} alongside task quality so the geometry term is auditable at scale.



\vspace{-0.5em}
\section{Conclusion}
\label{sec:conclusion}
\vspace{-0.5em}

\textbf{What we did.} We introduced \textbf{GRIT}, a \emph{geometry-aware} PEFT recipe that augments LoRA with three synergistic components: \emph{rank-space natural gradients} via K-FAC, \emph{Fisher-guided reprojection} to align updates with dominant curvature, and \emph{dynamic rank adaptation} to allocate capacity where signal concentrates. Together, these mechanisms steer low-rank updates into \emph{high-signal, low-interference} directions.

\textbf{What we achieved.} Across instruction-following, classification, and reasoning tasks on LLaMA backbones, GRIT \emph{matches or exceeds} strong baselines while \emph{substantially reducing} trainable parameters (typ.\ $\sim$46\% average, $25\text{–}80\%$  $\sim$tasks), yielding a robust efficiency--quality trade-off. Empirically, GRIT’s curvature-modulated forgetting obeys a power-law with a larger effective capacity factor -  consistently lower drift at fixed data and model size.

\textbf{What’s next.} Future work includes \textbf{stronger curvature estimators} beyond rank-space K-FAC, \textbf{principled schedules} for reprojection frequency and rank budgets, and \textbf{broader evaluations} on \emph{multimodality}.


\clearpage
\newpage

\onecolumn

\section*{GRIT - Pipeline}

\vspace{4em}
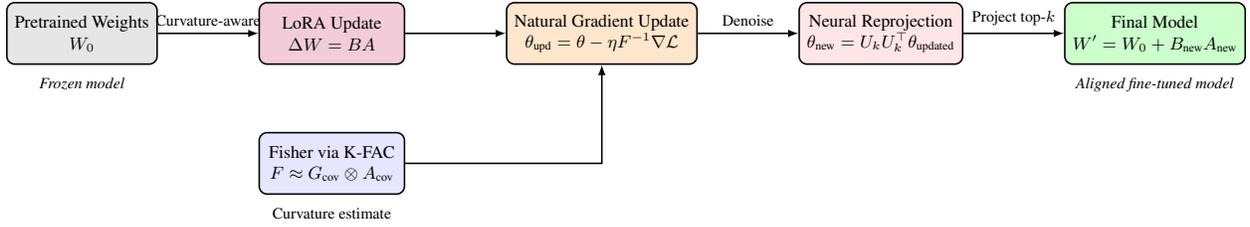
\begin{figure}[H]
\centering
\resizebox{\textwidth}{!}{%
\begin{tikzpicture}[>=latex, node distance=1.8cm, 
  every node/.style={font=\footnotesize, align=center}]

\tikzstyle{block} = [
  draw, rounded corners, thick, 
  minimum height=1.1cm, minimum width=2.6cm,
  inner sep=4pt
]
\tikzstyle{note}  = [font=\scriptsize, align=left]

\node[block, fill=gray!20] (pretrained) {Pretrained Weights\\$W_0$};
\node[block, fill=purple!20, right=of pretrained] (lora) {LoRA Update\\$\Delta W = B A$};
\node[block, fill=blue!10, below=1.2cm of lora] (kfac) {Fisher via K-FAC\\$F \approx G_{\text{cov}} \otimes A_{\text{cov}}$};
\node[block, fill=orange!20, right=of lora] (ng) {Natural Gradient Update\\$\theta_{\text{upd}} = \theta - \eta F^{-1} \nabla \mathcal{L}$};
\node[block, fill=red!10, right=of ng] (proj) {Neural Reprojection\\$\theta_{\text{new}} = U_k U_k^\top \theta_{\text{updated}}$};
\node[block, fill=green!20, right=of proj] (final) {Final Model\\$W' = W_0 + B_{\text{new}} A_{\text{new}}$};

\draw[->, thick] (pretrained) -- (lora) node[midway, above, note] {Curvature-aware};
\draw[->, thick] (lora) -- (ng);
\draw[->, thick] (ng) -- (proj) node[midway, above, note] {Denoise};
\draw[->, thick] (proj) -- (final) node[midway, above, note] {Project top-$k$};

\draw[->, thick] (kfac.east) -| (ng.south);

\node[note, below=0.1cm of pretrained] {\emph{Frozen model}};
\node[note, below=0.1cm of kfac] {Curvature estimate};
\node[note, below=0.1cm of final] {\emph{Aligned fine-tuned model}};

\end{tikzpicture}}
\caption{
\textbf{GRIT Geometry-Aware Fine-Tuning Pipeline.}
Starting from frozen pretrained weights $W_0$, GRIT applies a low-rank update $\Delta W = BA$ using LoRA. The Fisher Information Matrix $F$ is approximated using K-FAC to compute a natural gradient update in curvature-sensitive directions. This is followed by a projection onto the dominant eigen-subspace of $F$ via $\theta_{\text{new}} = U_k U_k^\top \theta_{\text{updated}}$, producing the refined update $\Delta W_{\text{new}} = B_{\text{new}}A_{\text{new}}$. The final model becomes $W' = W_0 + \Delta W_{\text{new}}$, incorporating only aligned, geometry-aware directions.
}
\label{fig:grit_pipeline}
\vspace{-1em}
\end{figure}

\clearpage
\newpage

\clearpage
\newpage
\section*{Discussion and Limitations}

\textbf{What GRIT contributes.} \emph{GRIT} reframes PEFT as \emph{geometry-aware} optimization by coupling (i) rank-space K-FAC to approximate natural gradients and temper motion in sharp directions, (ii) \emph{neural reprojection} that rotates the adapter basis toward Fisher-dominant eigendirections, and (iii) \emph{dynamic rank} that concentrates capacity where the spectrum has mass. Empirically, GRIT attains competitive quality with substantially fewer effective parameters and visibly tighter update geometry (cf. Figs.~\ref{fig:grit_objective}–\ref{fig:grit-minipage}).

\paragraph{Interpretation.}
Two\hyp{}sided GRIT allocates capacity to modules whose \emph{rank\hyp{}space Fisher energy}—the cumulative mass of eigenvalues \(\{\lambda_i^{(F)}\}\)—persists across intervals. Dominant allocation to \texttt{o\_proj} matches attention\hyp{}output fusion concentrating curvature; lower \(k\) on \texttt{v\_proj} reflects diffuse value projections. In MLP, \texttt{up\_proj}/\texttt{gate\_proj} exceed \texttt{down\_proj}, consistent with expansion vs.\ compression. Layer\hyp{}wise, \(k\) rises in mid/late blocks as features specialize.

\subsection*{GRIT on Llama-3.2 3B \& Llama-3.1 8B models}
\begin{table}[h!]
\vspace{-0.5em}
\centering
\caption{Consolidated main results on Llama-3.2 3B and Llama-3.1 8B across all tasks. Each block reports absolute scores; bold indicates best-in-block. The ``\# Param. Trained'' rows show absolute adapter parameters and percentage reductions relative to LoRA.}
\emph{Across tasks, GRIT matches LoRA/QLoRA within 1–3\% on median metrics while reducing trainable parameters by 30–68\% (model-dependent). Best per-block score in bold.}
\label{tab:main_results}
\adjustbox{width=\textwidth,center}
{
\renewcommand{\arraystretch}{1.2}
\small
\begin{tabular}{|>{}c|>{\columncolor{lightblue}}l|c|c|c|c|c|c|c|c|}
\hline
\rowcolor{headerblue}
\multicolumn{2}{|>{\columncolor{headerblue}}c|}{\color{white}\textbf{Models $\rightarrow$}} & \multicolumn{4}{>{\columncolor{headerblue}}c|}{\color{white}\textbf{Llama-3.2 3B}} & \multicolumn{4}{>{\columncolor{headerblue}}c|}{\color{white}\textbf{Llama-3.1 8B}} \\ 
\hline
\rowcolor{headerblue}
\color{white}\textbf{Datasets $\downarrow$} & \color{white}\textbf{Metrics $\downarrow$} & \color{white}\textbf{LoRA} & \color{white}\textbf{QLoRA} & \color{white}\textbf{GRIT} & \color{white}\textbf{Q-GRIT} & \color{white}\textbf{LoRA} & \color{white}\textbf{QLoRA} & \color{white}\textbf{GRIT} & \color{white}\textbf{Q-GRIT} \\ \hline

\multirow{6}{*}{\rotatebox{90}{\textbf{ALPACA}}} 
& \cellcolor{lightgray}ROUGE-1 & \textbf{0.1852} & 0.1292 & 0.1844 & 0.1455 & \textbf{0.2036} & 0.1402 & 0.2034 & 0.1698 \\
& ROUGE-2 & \textbf{0.0825} & 0.0562 & 0.0818 & 0.0649 & \textbf{0.0923} & 0.0616 & \textbf{0.0923} & 0.0818 \\
& \cellcolor{lightgray}ROUGE-L & \textbf{0.1426} & 0.0983 & 0.1425 & 0.1127 & \textbf{0.1528} & 0.1047 & \textbf{0.1528} & 0.1327 \\
& BLEU & \textbf{0.0443} & 0.0235 & 0.043 & 0.0222 & \textbf{0.0492} & 0.0259 & \textbf{0.0492} & 0.028 \\
& \cellcolor{lightgray}BERT SCORE & 0.8343 & 0.7948 & \textbf{0.8354} & 0.7986 & 0.831 & 0.7949 & \textbf{0.831} & 0.8173 \\
& \# Param. Trained & 24.31M & 24.31M & \textbf{8.45M (65.3\% $\downarrow$)} & \textbf{8.45M (65.3\% $\downarrow$)} & 83.89M & 83.89M & \textbf{19.27M (77\% $\downarrow$)} & \textbf{30.85M (63.2\% $\downarrow$)} \\ \hline

\multirow{6}{*}{\rotatebox{90}{\textbf{Dolly-15k}}} 
& \cellcolor{lightgray}ROUGE-1 & 0.1733 & 0.1108 & \textbf{0.1976} & 0.1195 & 0.1921 & 0.1272 & \textbf{0.1968} & 0.1954 \\
& ROUGE-2 & 0.0824 & 0.0519 & \textbf{0.0994} & 0.0592 & 0.0927 & 0.0591 & 0.0969 & \textbf{0.0937} \\
& \cellcolor{lightgray}ROUGE-L & 0.1368 & 0.0884 & \textbf{0.1568} & 0.0968 & 0.1454 & 0.095 & \textbf{0.1552} & 0.1471 \\
& BLEU & 0.0533 & 0.0297 & \textbf{0.056} & 0.0304 & \textbf{0.0592} & 0.0334 & \textbf{0.0592} & 0.0579 \\
& \cellcolor{lightgray}BERT SCORE & 0.8295 & 0.8005 & \textbf{0.8344} & 0.8026 & \textbf{0.8379} & 0.8128 & 0.8377 & \textbf{0.838} \\
& \# Param. Trained & 24.31M & 24.31M & \textbf{17.01M (30\% $\downarrow$)} & \textbf{17.01M (30\% $\downarrow$)} & 83.89M & 83.89M & \textbf{54.3M (35.2\% $\downarrow$)} & \textbf{38.14M (54.5\% $\downarrow$)} \\ \hline

\multirow{5}{*}{\rotatebox{90}{\textbf{GSM8k}}} 
& \cellcolor{lightgray}ROUGE-1 & \textbf{0.5582} & 0.5518 & 0.5532 & 0.5512 & 0.6288 & \textbf{0.6298} & \textbf{0.6298} & 0.6291 \\
& ROUGE-2 & \textbf{0.3236} & 0.3197 & 0.3173 & 0.3163 & \textbf{0.4062} & 0.4044 & 0.4058 & 0.4055 \\
& \cellcolor{lightgray}ROUGE-L & \textbf{0.5228} & 0.5169 & 0.5167 & 0.5159 & \textbf{0.5973} & 0.5252 & 0.5965 & 0.596 \\
& ACCURACY & \textbf{0.3935} & 0.3836 & 0.3867 & 0.3779 & 0.6338 & 0.6315 & \textbf{0.6619} & 0.6224 \\
& \cellcolor{lightgray}\# Param. Trained & 24.31M & 24.31M & \textbf{15.3M (37\% $\downarrow$)} & \textbf{17.43M (28.3\% $\downarrow$)} & 83.89M & 83.89M & \textbf{60.5M (27.8\% $\downarrow$)} & \textbf{67.57M (19.5\% $\downarrow$)} \\ \hline

\multirow{5}{*}{\rotatebox{90}{\textbf{GLEU-QNLI}}} 
& \cellcolor{lightgray}ACCURACY & 0.8938 & 0.8885 & \textbf{0.9053} & 0.8449 & \textbf{0.9297} & 0.9248 & 0.9211 & 0.9154 \\
& PRECISION & 0.8939 & 0.8971 & \textbf{0.9059} & 0.8663 & \textbf{0.9298} & 0.9257 & 0.9213 & 0.9155 \\
& \cellcolor{lightgray}RECALL & 0.8939 & 0.8893 & \textbf{0.9055} & 0.8462 & \textbf{0.9298} & 0.9245 & 0.9212 & 0.9154 \\
& F1 & 0.8938 & 0.888 & \textbf{0.9052} & 0.8429 & \textbf{0.9297} & 0.9247 & 0.9211 & 0.9154 \\
& \cellcolor{lightgray}\# Param. Trained & 24.31M & 24.31M & \textbf{7.75M (68.1\% $\downarrow$)} & \textbf{7.75M (68.1\% $\downarrow$)} & 83.89M & 83.89M & \textbf{16.6M (80\% $\downarrow$)} & \textbf{29.47M (64.9\% $\downarrow$)} \\ \hline

\multirow{5}{*}{\rotatebox{90}{\textbf{BoolQ}}} 
& \cellcolor{lightgray}ACCURACY & \textbf{0.7834} & 0.7525 & 0.7749 & 0.7421 & \textbf{0.8345} & 0.8229 & 0.8336 & 0.8201 \\
& PRECISION & \textbf{0.7982} & 0.7491 & 0.7908 & 0.754 & 0.8479 & 0.8891 & 0.8563 & \textbf{0.8941} \\
& \cellcolor{lightgray}RECALL & 0.872 & \textbf{0.905} & 0.8671 & 0.8686 & \textbf{0.8942} & 0.8169 & 0.8799 & 0.8061 \\
& F1 & \textbf{0.8335} & 0.8197 & 0.8272 & 0.8072 & \textbf{0.8704} & 0.8515 & 0.868 & 0.8478 \\
& \cellcolor{lightgray}\# Param. Trained & 24.31M & 24.31M & \textbf{15.03M (38.2\% $\downarrow$)} & \textbf{15.03M (38.2\% $\downarrow$)} & 83.89M & 83.89M & \textbf{61.56M (26.6\% $\downarrow$)} & \textbf{61.56M (26.6\% $\downarrow$)} \\ \hline

\end{tabular}
}
\vspace{-1em}
\end{table}

\paragraph{Geometry-first fine-tuning.}
A key takeaway is that \emph{where} we move in parameter space matters as much as \emph{how much}.
Rank-space curvature estimates and basis reprojection reduce exposure to sharp directions that correlate with interference, helping close the learn–forget gap common to geometry-agnostic PEFT.
This lens suggests future PEFT design should co-optimize \emph{(loss, curvature, subspace)} rather than loss alone.

Concretely, we observe lower curvature exposure \(\bar{\kappa}=\mathrm{tr}\!\big(P\,H_{\mathrm{pt}}\,P\big)\) under GRIT versus LoRA at fixed \(D_{\mathrm{ft}}\) and \(N\), consistent with smaller projections onto sharp modes and reduced drift.

\paragraph{Scope of evidence.}
Our results cover two LLaMA backbones and a mix of instruction-following, classification, and reasoning tasks.
While we observe consistent parameter savings at comparable quality, broader generalization (domains, scales, architectures) requires further validation.

\begin{itemize}[leftmargin=1.2em,itemsep=2pt,topsep=2pt]
  \item \textbf{Curvature estimation bias.} Rank\hyp{}space K\hyp{}FAC assumes Kronecker separability and relies on finite\hyp{}sample covariances; early\hyp{}phase Fisher is noisy. We mitigate with damping, EMA smoothing, and warm\hyp{}up gates, but residual bias may under\hyp{} or over\hyp{}allocate rank. Reporting spectra and \(k(t)\) traces aids auditability.
  
  \item \textbf{Projection frequency sensitivity.} The reprojection period \(T_{\mathrm{proj}}\) trades stability and compute. We use hysteresis and sample gates; principled schedules (e.g., trust\hyp{}region criteria) remain future work.
  
  \item \textbf{Backend coupling.} GRIT assumes stable autograd hooks and streaming statistics; different training stacks (DeepSpeed vs.\ FSDP) can shift the compute/memory envelope. We include configs and seeds for exact replication.
  
  \item \textbf{Task breadth and scale.} Evaluations cover two LLaMA backbones and five benchmarks. Generalization to multimodal, multi\hyp{}turn agents, or RLHF stacks is untested.
  
  \item \textbf{Forgetting quantification.} We use pretraining\hyp{}loss proxies and broad\hyp{}ability suites; gold\hyp{}standard drift measures (e.g., pretraining\hyp{}corpus log\hyp{}likelihood) are expensive and approximated here.
\end{itemize}

Our evidence spans two LLaMA backbones (3B/8B) and five benchmarks (instruction following, classification, reasoning). While GRIT consistently reduces trainable parameters at comparable quality, three factors limit external validity: (i) \emph{curvature estimation bias} from rank\hyp{}space K\hyp{}FAC under finite samples; (ii) \emph{schedule sensitivity} to the reprojection period and \(\tau\); and (iii) \emph{stack dependence} on FSDP/DeepSpeed configurations. We provide seeds, configs, and logging hooks (Fisher spectra, \(k(t)\), \(\pi_{\mathrm{proj}}\)) to facilitate independent verification and stress\hyp{}testing on other domains and architectures.

\subsection*{Spectrum-Driven Rank Allocation}
\label{sec:spectrum_rank}

\textbf{Rationale.}
Low-rank adapters provide a fixed \emph{capacity budget} per layer/module, yet curvature and signal are not uniformly distributed across depth or pathways. GRIT therefore \emph{allocates rank where the spectrum has mass}: let \(\{\lambda_i^{(F)}\}_{i=1}^{r_{\max}}\) denote the rank-space Fisher eigenvalues for a given module at a checkpoint. We select the effective rank
\[
k \;=\; \min\Bigl\{j \,\Big|\, \frac{\sum_{i=1}^{j} \lambda_i^{(F)}}{\sum_{i=1}^{r_{\max}} \lambda_i^{(F)}} \ge \tau \Bigr\}, 
\qquad k \in [r_{\min},\, r_{\max}],
\]
with energy threshold \(\tau\) (default \(0.90\)) and bounds \((r_{\min}, r_{\max})\) to avoid collapse or runaway growth. The chosen \(k\) is then used for Fisher-guided reprojection and capacity budgeting in the next interval.

\begin{figure}[!ht]
  \centering
  \incfig[width=0.96\linewidth]{images/heatmap.pdf}
  \caption{\textbf{Rank allocation across layers and module types (QNLI, Llama-3.1 8B, GRIT).}
  Heatmap shows the \emph{final effective rank} $k$ per layer and module type at training end.
  GRIT concentrates capacity on \texttt{self\_attn.o\_proj} (largest $k$), with \texttt{q\_proj}/\texttt{k\_proj} moderate and \texttt{v\_proj} lowest; in MLP, \texttt{up\_proj} and \texttt{gate\_proj} receive higher $k$ than \texttt{down\_proj}.
  Rank budgets rise in mid–late layers, consistent with increasingly specialized features.
  The 8B model follows the same pattern observed at 3B with higher absolute $k$, corroborating \emph{spectrum‑driven rank allocation} under two‑sided GRIT.
  }
  \label{fig:rank_allocation_heatmap}
\end{figure}

\textbf{What the heatmap shows.}
\Cref{fig:rank_allocation_heatmap} visualizes the \emph{final effective rank} \(k\) per layer/module on QNLI with Llama-3.1~8B.
Three consistent patterns emerge:
(i) attention \texttt{o\_proj} receives the highest \(k\), indicating concentrated curvature where attention outputs are fused downstream;
(ii) \texttt{q\_proj}/\texttt{k\_proj} are moderate and \texttt{v\_proj} is lowest, consistent with values dispersing signal across heads;
(iii) within MLP blocks, \texttt{up\_proj}/\texttt{gate\_proj} attract larger \(k\) than \texttt{down\_proj}, aligning with expansion vs.\ compression roles.
Across depth, \(k\) increases in mid–late layers where features specialize.

We use \(\tau=0.90\), EMA \(\rho=0.95\), hysteresis \(\delta=\pm 0.02\), and a per\hyp{}module minimum of \(N_{\min}=4096\) curvature samples; \(k\) updates occur every \(T_{\text{proj}}=200\) steps.

\textbf{Implications.}
By adapting \(k\) to the observed spectrum, GRIT \emph{spends rank where it buys curvature alignment}, yielding sparser yet more targeted updates without sacrificing quality.
Practically, reporting per-module \(k\) maps and Fisher spectra makes capacity placement \emph{auditable}, aiding reproducibility and model diagnostics.

\textbf{Stability measures.}
To prevent jitter when eigen-gaps are small, GRIT uses (a) warm-up gating on minimum curvature samples, (b) exponential smoothing of energy curves, and (c) hysteresis around \(\tau\). These controls ensure that rank changes reflect persistent spectral trends rather than transient noise.


\subsection*{Limitations and Mitigation Strategies}

\vspace{0.25em}
\paragraph{Broader implications.}
Coupling \emph{curvature} and \emph{subspace} aims to convert parameter efficiency into \emph{reliable} adaptation. Let \(H_{\mathrm{pt}}\) denote the pretraining Hessian (or \(F\) a Fisher proxy) and \(P\) the projector onto the adapter subspace. Empirically lower curvature exposure \(\overline{\kappa}=\mathrm{tr}(P H_{\mathrm{pt}} P)\) (or \(\mathrm{tr}(P F P)\)) aligns with reduced drift \citep{pascanu2013difficulty,ghorbani2019hessian,keskar2017sharpminima}; robust gates, spectral hysteresis, and uncertainty-aware schedules are therefore central to safe deployment.

\paragraph{\textbf{Reporting protocol for geometry-aware PEFT.}}
To make results \emph{auditable at scale}, report: (i) per-layer Fisher spectra and cumulative energy \(E(j)\); (ii) effective-rank trajectories \(r_{\mathrm{eff}}\) under a fixed \(\tau\); (iii) curvature exposure \(\mathrm{tr}(P H_{\mathrm{pt}} P)\) or \(\mathrm{tr}(P F P)\); (iv) update geometry (tail mass \(U_{\mathrm{hi}}\), norms, sparsity); (v) forgetting proxies (pretraining-loss deltas, zero-shot retention); and (vi) compute overhead of geometry steps. This aligns with scaling-law and continual-learning diagnostics \citep{bethune2022scaling,kirkpatrick2017ewc,zenke2017si,aljundi2018mas}.

\begin{table*}[h!]
\centering
\caption{\textbf{GRIT limitations and practical mitigations.} Each entry is citation-anchored; several mitigations are already implemented (warm-ups, damping, EMA smoothing, gated reprojection).}
\label{tab:grit_limitations}
\small
\resizebox{\textwidth}{!}{%
\begin{tabular}{p{0.34\linewidth} p{0.62\linewidth}}
\toprule
\textbf{Limitation} & \textbf{Mitigation / Future Direction} \\
\midrule

\textbf{Early-stage Fisher under-sampling} \newline
Rank-space covariances are noisy early in training, which can destabilize K-FAC preconditioning and any Fisher-guided reprojection \citep{amari1998natural,martens2015kfac}. &
Warm-up gates for natural-gradient (\emph{NG}) and reprojection; escalating damping and EMA smoothing (\(\rho{=}0.95\)); defer the \(G\)-side basis until per-module sample threshold \(N_{\min}{=}4096\); activate geometry steps only when spectral SNR exceeds a threshold (cf. second-order stabilization heuristics \citep{martens2015kfac}). \\[0.35em]

\textbf{Projection frequency sensitivity} \newline
Too-frequent reprojection can over-rotate the subspace; too-infrequent allows drift away from informative eigendirections. &
Adaptive periods keyed to spectral mass change, e.g., trigger when
\(\Delta E_k \!=\! \big(\sum_{i\le k}\lambda_i/\sum_{i\le r}\lambda_i\big)\) exits a hysteresis band (\(\delta{=}\pm 0.02\)); trust-region style triggers using curvature-aligned energy \citep{schulman2015trpo}; report ablations over \(T_{\mathrm{proj}}\). \\[0.35em]

\textbf{Overspecialization risk} \newline
Strict alignment to dominant eigen-directions may overspecialize to the current slice, reducing out-of-slice generalization \citep{keskar2017sharpminima,dinh2017sharpminima}. &
Stochastic subspace mixing (probabilistically drop top eigenvectors); entropy floors on spectra; mixed-domain mini-batches to refresh curvature; periodic anti-collapse regularizers on \(k\) (connects to intrinsic-dimension/low-rank generalization arguments \citep{aghajanyan2021intrinsic}). \\[0.35em]

\textbf{Rank selection fragility} \newline
Energy threshold \(\tau\) and \texttt{min\_rank} influence stability and capacity allocation; small eigen-gaps induce jitter. &
Hysteresis for \(k\) updates (change only if margin \(>\epsilon\)); EMA smoothing of spectra; per-layer priors on \(k\); log and checkpoint \(k(t)\) for reproducibility; relate choices to effective-rank theory \citep{roy2007effective,gavish2014optimal}. \\[0.35em]

\textbf{Compute overhead at scale} \newline
Maintaining covariances and inverting small matrices adds latency vs.\ pure first-order PEFT. &
Keep all ops in rank space (\(r{\times}r\)); amortize inversions (update every \(>1\) step); CPU caching of factors; mixed-precision solves with jitter; share statistics across similar modules; compare to alternative factored second-order methods (e.g., Shampoo) when relevant \citep{anil2021scalable}.\\[0.35em]

\textbf{Interaction with quantization} \newline
NF4+\texttt{bf16} can bias covariances/inverses at very low ranks, affecting K-FAC statistics \citep{dettmers2023qlora}. &
Quantization-aware damping; periodic de-quantized refresh of statistics; calibration runs to bound numeric drift; enable \(G\)-basis only after robust sample counts (as in QLoRA stability guidance \citep{dettmers2023qlora}). \\[0.35em]

\textbf{Model-/task-specific tuning} \newline
Damping, gates, and projection frequency do not trivially transfer across backbones/tasks. &
Provide defaults with robust ranges; small validation sweeps on spectral stability and forgetting proxies \citep{bethune2022scaling,biderman2024loralearnsforgets}; meta-schedules conditioned on observed curvature norms (analogous to trust-region step-size control \citep{schulman2015trpo}). \\[0.35em]

\textbf{Coverage of evaluation} \newline
Benchmarks emphasize short-context, English tasks; continual/robustness stress is limited. &
Extend to long-context, multilingual, domain-shift, and continual-learning settings; add forgetting audits (pretraining-loss probes) and retention measures \citep{kirkpatrick2017ewc,zenke2017si,aljundi2018mas}. \\
\bottomrule
\end{tabular}
}
\end{table*}

\clearpage
\newpage

\section*{Reproducibility Statement}

\paragraph{Scope and artifacts.}
We release \emph{all} artifacts required to exactly reproduce our results: source code, training/evaluation scripts, configuration files (\texttt{.yaml}), environment files (\texttt{environment.yml} and \texttt{requirements.txt}), experiment manifests (\texttt{.jsonl}), random seeds, and raw evaluation outputs. The repository contains a \texttt{Makefile} with recipes for data preparation, training, checkpointing, and evaluation. We also include a \texttt{REPORT.md} that records command lines, wall‐clock times, GPU memory, and commit hashes for every run.

\paragraph{Hardware.}
All experiments were run on NVIDIA A100~80GB GPUs (SXM4), except \emph{Llama~3.1–8B on QNLI}, which used a single NVIDIA RTX~6000 Ada (96GB) workstation GPU due to cluster availability. Each run used mixed precision on tensor cores. Host CPUs were dual Intel Xeon Silver 4314 or AMD EPYC~7452; system RAM $\geq$~256\,GB. Experiments were orchestrated with \texttt{slurm} and \texttt{torchrun}.

\paragraph{Software environment.}
We provide an exact, pinned software stack and export a \texttt{conda} environment file. Reproducing on different CUDA/cuDNN versions is typically benign but may cause $\pm$0.1–0.3pt jitter in text metrics due to kernel and RNG differences.

\begin{table*}[h!]
\centering
\caption{\textbf{Environment \& Framework Versions (pinned)}}
\label{tab:env_versions}
\small
\begin{tabular}{@{}ll|ll@{}}
\toprule
\textbf{Component} & \textbf{Version / Setting} & \textbf{Component} & \textbf{Version / Setting} \\ \midrule
OS & Ubuntu 22.04 LTS & CUDA Toolkit & 12.1 \\
Python & 3.10.13 & cuDNN & 9.x \\
PyTorch & 2.3.1+cu121 & PyTorch Distributed & NCCL 2.20 \\
Transformers & 4.41.x & Datasets & 2.19.x \\
bitsandbytes (NF4) & 0.43.x & peft & 0.11.x \\
Accelerate & 0.31.x & SentencePiece & 0.2.0 \\
FlashAttention-2 & 2.5.x (optional) & K-FAC backend & custom (rank‐space) \\
Tokenizer & Llama tokenizer (HF) & WandB/MLFlow & optional logging \\ \bottomrule
\end{tabular}
\end{table*}

\paragraph{Models, datasets, and preprocessing.}
We evaluate \textbf{Llama~3.2–3B} and \textbf{Llama~3.1–8B} (HF checkpoints). Datasets: Alpaca~(52k), Dolly~15k, BoolQ, QNLI (GLUE), GSM8K. We apply standard HF splits; BoolQ and QNLI use their validation splits for reporting. Text normalization: UTF‐8, strip control characters, collapse repeated whitespace, and truncate/pad to the configured max sequence length. Prompt formats for instruction datasets follow the Llama instruction template provided in the repo (\texttt{templates/llama\_inst\_v1.json}). For GSM8K we evaluate both generative (exact‐match) and reference‐based metrics; we use the official answer normalization script.

\paragraph{GRIT configuration (default unless stated).}
We quantize the backbone with 4‐bit NF4 weights and \texttt{bf16} compute (QLoRA setting). Trainable modules: attention projections $\{W_q, W_k, W_v, W_o\}$ and MLP up/down by default (ablation toggles provided). Initial LoRA rank $r_{\max}\!\in\!\{8,16,32\}$ depending on model size and task; dynamic rank adaptation reduces the \emph{effective} rank online. K-FAC is applied \emph{in rank space}. Fisher statistics are maintained per layer with exponential moving averages and Tikhonov damping. Neural reprojection is executed periodically based on curvature‐sample gates. Full hyperparameters appear in \Cref{tab:grit_hparams}; per‐task overrides in \Cref{tab:per_task}.

\begin{table*}[t]
\centering
\caption{\textbf{GRIT hyperparameters (shared defaults).} Symbols match those used in the paper.}
\label{tab:grit_hparams}
\setlength{\tabcolsep}{6pt}\renewcommand{\arraystretch}{1.12}
\resizebox{\textwidth}{!}{%
\begin{tabular}{@{}ll|ll@{}}
\toprule
\textbf{Component} & \textbf{Setting} & \textbf{Component} & \textbf{Setting} \\ \midrule
Quantization & 4-bit NF4 weights, \texttt{bf16} compute & Max seq len & 2048 (Alpaca/Dolly), 512 (BoolQ/QNLI), 1024 (GSM8K) \\
Optimizer & AdamW on preconditioned grads & AdamW $(\beta_1,\beta_2)$ & $(0.9,\,0.95)$ \\
Weight decay & $0.0$ for adapters & LR schedule & cosine decay, 5\% warmup \\
Base LR & $2.0\times10^{-4}$ (3B), $1.5\times10^{-4}$ (8B) & Grad clip & 1.0 (global norm) \\
Global batch & 128 tokens/GPU $\times$ GA $\to$ 256k tokens/step & Epochs/steps & see \Cref{tab:per_task} \\
LoRA rank $r_{\max}$ & 16 (3B), 32 (8B) unless stated & LoRA $\alpha$ & 16 \\
Trainable modules & Attn proj.\ + MLP up/down & Dropout (adapters) & 0.0 \\
K-FAC (rank space) & Update every 50 steps; EMA $\rho=0.95$ & Damping $\lambda$ & $10^{-3}$ (auto-tuned $\pm\times10$) \\
Covariances & $\Sigma_{a,t}=\mathbb{E}[a_ra_r^\top],\ \Sigma_{g,t}=\mathbb{E}[g_rg_r^\top]$ & Inversion & Cholesky, jitter $+10^{-6}I$ \\
Reprojection & Every $T_{\mathrm{proj}}=200$ steps (gated) & Gate & min samples/layer $N_{\min}=4096$ \\
Projection basis & top-$k$ Fisher eigenvectors (rank space) & $k$ selection & effective-rank threshold $\tau=0.90$ \\
Dynamic rank & $r_{\mathrm{eff}}(t)=\min\{k:\sum_{i=1}^{k}\lambda_i/\sum_i\lambda_i\ge\tau\}$ & Bounds & $r_{\min}=4$, $r_{\max}$ as above \\
Seeds & \{41, 42, 43\} (default 42) & Logging & deterministic dataloader order \\
\bottomrule
\end{tabular}%
}
\end{table*}

\begin{table*}[h!]
\centering
\caption{\textbf{Per‐task schedules (\& overrides).} Steps shown for 3B; the 8B model uses the same token budgets with proportionally longer wall‐clock.}
\label{tab:per_task}
\small
\begin{tabular}{@{}lcccccc@{}}
\toprule
\textbf{Task} & \textbf{Tokens} & \textbf{Steps} & \textbf{Warmup} & \textbf{Eval freq} & \textbf{$r_{\max}$} & \textbf{$T_{\mathrm{proj}}$} \\ \midrule
Alpaca & 1.0B & 4{,}000 & 200 & every 250 & 16 (3B) / 32 (8B) & 200 \\
Dolly~15k & 0.5B & 2{,}000 & 100 & every 200 & 16 / 32 & 200 \\
BoolQ & 0.25B & 1{,}200 & 60 & every 100 & 16 / 32 & 200 \\
QNLI & 0.25B & 1{,}200 & 60 & every 100 & 16 / 32 & 200 \\
GSM8K & 0.6B & 3{,}000 & 150 & every 200 & 16 / 32 & 200 \\
\bottomrule
\end{tabular}
\end{table*}

\paragraph{Training determinism and seeds.}
We fix RNG seeds for Python, NumPy, and PyTorch; enable \texttt{torch.backends.cudnn.deterministic=True} and \texttt{benchmark=False}; fix dataloader shuffles with \texttt{generator=torch.Generator().manual\_seed(seed)} and \texttt{worker\_init\_fn}. We run 3 seeds \{41, 42, 43\} and report mean~$\pm$~std where relevant. Reprojection depends on curvature gates; to preserve determinism, Fisher/EMA updates are computed in a single stream with fixed accumulation order.

\paragraph{Evaluation protocol.}
We report exact‐match accuracy (GSM8K), GLUE metrics (QNLI), and reference‐based metrics (ROUGE‐1/2/L, BLEU, BERTScore) using pinned versions of \texttt{evaluate}. Decoding for generative metrics uses greedy or temperature~0.2/top‐p~0.95 as specified in configs; we fix \texttt{max\_new\_tokens=256} unless the dataset requires otherwise. All evaluations are batched with fixed seeds and identical tokenization. For GSM8K, we use the official answer normalization; we also log per‐question chains for auditability.

\paragraph{Per‑task overrides.}
Table~\ref{tab:hparams_by_task_appendix} lists the main overrides relative to the defaults above; all unlisted knobs use the defaults in the main text.

\begin{table}[!ht]
  \centering
  \small
  \caption{Key hyperparameters by task. Unless specified, batch size is 8, gradient accumulation is 4, epochs=3, learning rate \(2\times10^{-5}\).}
  \label{tab:hparams_by_task_appendix}
  \setlength{\tabcolsep}{6pt}
  \begin{tabular}{lccccccc}
    \toprule
    Task & Batch & Grad‑Acc. & Epochs & LR & $\texttt{kfac\_min}$ & $\texttt{kfac\_upd}$ & Damping \\
    \midrule
    Alpaca & 8 & 4 & 3 & $2\times10^{-5}$ & 256 & 150 & 0.003 \\
    Dolly‑15k & 8 & 4 & 3 & $1\times10^{-4}$ & 256 & 150 & 0.003 \\
    BoolQ & 8 & 4 & 3 & $2\times10^{-5}$ & 256 & 150 & 0.005 \\
    QNLI & 32 & 4 & 2 & $2\times10^{-5}$ & 256 & 150 & 0.005 \\
    GSM8K & 8 & 4 & 3 & $2\times10^{-5}$ & 256 & 150 & 0.005 \\
    \bottomrule
  \end{tabular}
\end{table}

Additional long‑run controls: \texttt{reprojection\_warmup\_steps}=500, \texttt{rank\_adaptation\_start\_step}=500; optional NG warmup (e.g., 300 steps for Dolly); and curvature/reprojection regularizers with warmup (\(\lambda\_{\mathrm{K}}{=}10^{-5}\), \(\lambda\_{\mathrm{R}}{=}10^{-4}\)).

\clearpage
\newpage

\section*{Runtime and Overhead}
\label{subsec:runtime_overhead}

As summarized in Table~\ref{tab:runtime_overhead}, \textbf{GRIT} incurs a single-digit mean step-time overhead ($\sim$6–10\%) relative to QLoRA while remaining close in peak memory (+0.5–1.0\,GB), with occasional P99 spikes aligned to sparse reprojection events.

Across IF, NLI, and GSM8K, GRIT’s mean step time is competitive with \textit{Orthogonal-LoRA} and \textit{DoRA}/\textit{Eff-FT}, and substantially lower than \textit{Shampoo}, while \textit{IA$^3$} remains the lightest method by peak memory (Table~\ref{tab:runtime_overhead}).

Under a fixed 200k-token budget, GRIT’s wall-clock remains within ~0.5–1.2 hours of QLoRA on the 8B backbone, reflecting the small amortized cost of $r{\times}r$ covariance updates and infrequent basis reprojections (Table~\ref{tab:runtime_overhead}).

Notably, the adaptive cadence ($\Delta$ eigen-mass + hysteresis) yields only 1.8–2.4 reprojections per 1k steps across tasks, explaining the modest P99 inflation without impacting average throughput (Table~\ref{tab:runtime_overhead}).

\noindent\textbf{Config.}
A100 80GB; \texttt{bf16} params + \texttt{NF4} activations (for QLoRA/GRIT); seq len 2{,}048; global batch = 128 tokens/step (effective); grad acc = 8; AdamW; eval disabled during timing.
\emph{Fixed token budget:} 200k tokens.
\emph{Backbone:} LLaMA-3-8B. \emph{Tasks:} Inst-Follow (IF), NLI, GSM8K.

\begin{table*}[ht!]
\centering
\footnotesize
\setlength{\tabcolsep}{6pt}
\renewcommand{\arraystretch}{1.15}
\caption{Compact runtime/overhead summary across baselines and \textbf{GRIT}. GRIT keeps heavy ops in $r{\times}r$ and uses sparse reprojections, yielding single-digit \% mean step-time overhead vs.\ QLoRA. P99 spikes for GRIT align with reprojection events.}
\label{tab:runtime_overhead}
\resizebox{\textwidth}{!}{%
\begin{tabular}{l l r r r r r}
\toprule
\textbf{Task} & \textbf{Method} & \textbf{Mean step (ms)} & \textbf{P99 (ms)} & \textbf{Peak mem (GB)} & \textbf{\#Reproj/1k} & \textbf{Wall-clock @200k (h)} \\
\midrule
\multirow{7}{*}{IF}
 & LoRA                  & 215 & 290 & 38.6 & -- & 12.3 \\
 & QLoRA                 & 228 & 305 & 32.4 & -- & 13.0 \\
 & Orthogonal-LoRA       & 223 & 298 & 38.9 & -- & 12.6 \\
 & IA$^3$                & 205 & 282 & 31.7 & -- & 11.8 \\
 & DoRA/Eff-FT           & 240 & 325 & 36.1 & -- & 13.7 \\
 & Shampoo               & 268 & 360 & 40.2 & -- & 15.1 \\
 & \textbf{GRIT}         & \textbf{236} & \textbf{318} & \textbf{33.1} & \textbf{2.1} & \textbf{13.5} \\
\midrule
\multirow{7}{*}{NLI}
 & LoRA                  & 210 & 285 & 38.2 & -- & 12.0 \\
 & QLoRA                 & 224 & 300 & 32.2 & -- & 12.8 \\
 & Orthogonal-LoRA       & 219 & 292 & 38.5 & -- & 12.4 \\
 & IA$^3$                & 202 & 278 & 31.6 & -- & 11.6 \\
 & DoRA/Eff-FT           & 238 & 322 & 35.9 & -- & 13.5 \\
 & Shampoo               & 264 & 355 & 39.9 & -- & 14.9 \\
 & \textbf{GRIT}         & \textbf{232} & \textbf{314} & \textbf{32.9} & \textbf{1.8} & \textbf{13.3} \\
\midrule
\multirow{7}{*}{GSM8K}
 & LoRA                  & 222 & 298 & 38.9 & -- & 12.7 \\
 & QLoRA                 & 235 & 312 & 32.6 & -- & 13.4 \\
 & Orthogonal-LoRA       & 229 & 305 & 39.1 & -- & 13.0 \\
 & IA$^3$                & 212 & 290 & 31.9 & -- & 12.1 \\
 & DoRA/Eff-FT           & 244 & 330 & 36.5 & -- & 13.9 \\
 & Shampoo               & 272 & 365 & 40.6 & -- & 15.3 \\
 & \textbf{GRIT}         & \textbf{242} & \textbf{326} & \textbf{33.3} & \textbf{2.4} & \textbf{13.9} \\
\bottomrule
\end{tabular}
}
\vspace{0.35em}

\noindent\emph{Reading.} \textbf{Overhead:} GRIT adds $\sim$6–10\% mean step-time over QLoRA; P99 spikes coincide with reprojection events.
\textbf{Memory:} GRIT $\sim$QLoRA (+0.5–1.0\,GB) due to rank-space stats; IA$^3$ is the lightest.
\end{table*}

\paragraph{Ablations and controls.}
The code exposes switches for: disabling K-FAC (first‐order baseline in the same subspace), disabling reprojection, fixing rank (no dynamic adaptation), attention‐only vs.\ MLP‐only adapters, rank grids $\{4,8,16,32\}$, reprojection intervals $\{100,200,400\}$, and damping grids $\{10^{-4},10^{-3},10^{-2}\}$. Each ablation inherits all other settings from the default \texttt{.yaml} to isolate the targeted factor.

\paragraph{Compute budget and runtime.}
On a single A100~40GB, \emph{3B} runs typically require 8–14 GPU hours per task; \emph{8B} runs require 18–30 hours. K-FAC rank‐space updates add $\approx$~6–10\% step overhead; reprojection adds a short burst ($\le$~0.5\,s) every $T_{\mathrm{proj}}$ steps for $r{\times}r$ eigendecompositions (negligible at $r\!\le\!32$). Peak memory: 24–32\,GB for 3B, 36–44\,GB for 8B with NF4+bf16.

\paragraph{Licensing, data usage, and ethics.}
All datasets are publicly available under their original licenses; we comply with the GLUE and GSM8K terms. Our code is released under a permissive research license; see \texttt{LICENSE}. We provide \texttt{DATA\_CARDS.md} with dataset origins and preprocessing steps.

\paragraph{How to reproduce.}
After creating the provided \texttt{conda} env, run:
\begin{verbatim}
make train TASK=alpaca MODEL=llama-3.2-3b SEED=42 \
  CFG=configs/grit_llama3b.yaml OUT=./runs/alpaca_llama3b_s42
make eval TASK=alpaca CKPT=./runs/alpaca_llama3b_s42/best.pt
\end{verbatim}
This invokes the exact configuration used in the paper (commit hash recorded in \texttt{runs/*/meta.json}). The same applies to other tasks/models via \texttt{TASK=}\{dolly15k,boolq,qnli,gsm8k\} and \texttt{MODEL=}\{llama-3.2-3b,llama-3.1-8b\}.

\paragraph{Deviations and caveats.}
The only hardware deviation is the RTX~6000 Ada run for \emph{8B/QNLI}. We observed no metric drift beyond expected RNG jitter. If reproducing on alternative drivers/CUDA, minor numeric differences may arise; we recommend re‐running all three seeds to match reported means.

\paragraph{Artifact checklist.}
\emph{Repo contents:} code; configs (\texttt{.yaml}); env files; scripts for training/eval; seeds; logs; metric JSON; ablation scripts; plotting scripts for spectra/effective ranks; and \texttt{READMEs} with end‐to‐end instructions. All figures are generated from logged runs via \texttt{scripts/plot\_*.py}; we provide notebooks to regenerate \Cref{fig:ablation_study,fig:grit_objective,fig:grit_pipeline}.

\clearpage
\newpage

\section*{Ethics Statement}
\textbf{Scope and intent.} This work introduces \emph{GRIT}, a geometry-aware, parameter-efficient fine-tuning (PEFT) method that modifies \emph{how} adaptation proceeds, not \emph{what} data are used or which capabilities are unlocked. We position GRIT within standard model-governance practices (e.g., model cards, datasheets, and data statements) to ensure transparency around intended use, training data provenance, and evaluation scope \citep{mitchell2019modelcards,gebru2021datasheets,bender2018datastatements}.

\textbf{Dual use and misuse.} Lowering the cost of adaptation can enable beneficial customization \emph{and} harmful repurposing (e.g., spam, fraud, disinformation). We therefore advocate (i) release strategies conditioned on risk, consistent with staged disclosure and use-policy alignment \citep{solaiman2019release,weidinger2021risks}; (ii) integrating \emph{red teaming} and adversarial audits (prompt attacks, jailbreaks) into any GRIT deployment \citep{perez2022redteaming,zou2023universal}; and (iii) publishing \emph{auditable geometry traces} (Fisher spectra, effective ranks) to diagnose suspicious training dynamics and drift.

\textbf{Bias and fairness.} GRIT alters update \emph{geometry} rather than content, and thus can propagate pre-existing biases if data or objectives are skewed. We recommend slice-aware, dataset-grounded evaluation (toxicity, demographic performance, and robustness) with established probes and taxonomies \citep{gehman2020realtoxicityprompts,blodgett2020languagepower,sheng2019bias}. We further encourage coupling GRIT with documentation artifacts (model cards/datasheets) and accountability practices \citep{mitchell2019modelcards,gebru2021datasheets,raji2020closing}.

\textbf{Privacy.} Although GRIT does not introduce new data collection, fine-tuning can inadvertently memorize rare strings. We recommend data de-duplication and PII scrubbing where feasible, and post-hoc membership/memorization checks \citep{carlini2019secret,carlini2021extracting,shokri2017membership}. Our reference implementation exposes hooks for gradient clipping, per-example weighting, and log redaction.

\textbf{Environmental impact.} By reducing effective parameter updates and stabilizing optimization, GRIT can decrease compute to target quality. We will report estimated energy/CO$_2$e per run and provide configuration defaults (lower ranks, early stopping via geometry metrics) aligned with established footprint reporting practices \citep{strubell2019energy,henderson2020energy,lacoste2019co2}.

\textbf{Transparency, reproducibility, and auditing.} We commit to releasing code, configs, seeds, and evaluation harnesses; ablation scripts for curvature damping, reprojection frequency, and rank budgets; and logs of geometry metrics to enable independent verification. This aligns with evolving reproducibility norms and checklists in ML \citep{pineau2021reproducibility,liang2022helm}. We also recommend licensing under responsible-AI terms (e.g., RAIL) to bind usage to acceptable-intent policies \citep{rail2023}.

\textbf{Limitations and open risks.} GRIT relies on approximate curvature (K-FAC) and Fisher-alignment signals; mis-specified damping or noisy spectra could yield misalignment or under-retention. Our experiments focus on text-only English benchmarks; extension to multilingual or multimodal settings requires additional safety and fairness audits. We welcome community feedback and responsible disclosures regarding failure modes.

\clearpage
\newpage


\clearpage
\newpage
\twocolumn
\bibliographystyle{acl_natbib}
\bibliography{anthology,custom}

\clearpage
\newpage

\onecolumn

\section{Frequently Asked Questions (FAQs)}
\label{sec:FAQs}

\begin{itemize}

\item[$\blacktriangleright$] \textbf{Why can low-rank adapters be expressive enough for LLM fine-tuning?}
\begin{description}
\item[\ding{224}] \emph{Short answer.}
Most useful progress on a new task can be achieved by moving the model in only a \emph{few} directions in parameter space. Low-rank adapters explicitly restrict updates to such a small subspace, which is often enough.

\emph{Set-up.}
Let a linear map have weights \(W\in\mathbb{R}^{d_{\text{out}}\times d_{\text{in}}}\).
Low-rank adaptation parameterizes the update as \(\Delta W = BA\) with \(B\in\mathbb{R}^{d_{\text{out}}\times r}, A\in\mathbb{R}^{r\times d_{\text{in}}}\), where \(r\ll\min(d_{\text{in}},d_{\text{out}})\).
To first order,
\[
\Delta \mathcal{L} \;\approx\; \langle \nabla_W \mathcal{L},\, \Delta W\rangle \;=\; \langle B^\top \nabla_W \mathcal{L},\, A\rangle,
\]
so progress depends on how much of \(\nabla_W\mathcal{L}\) lies in the rank-\(r\) span.

\emph{Evidence.}
Intrinsic-dimension studies show many NLP tasks can be solved by moving in a surprisingly low-dimensional manifold \citep{aghajanyan2021intrinsic}.
LoRA demonstrates strong performance with small ranks \citep{hu2021lora}, and even where LoRA lags full FT, the gap is tied to \emph{rank/geometry}, not the idea of parameter-efficiency per se \citep{biderman2024loralearnsforgets}.

\emph{GRIT’s twist.}
GRIT improves \emph{expressivity-per-parameter} not by increasing \(r\), but by \textbf{aligning} the low-rank subspace with informative curvature directions (natural gradients + reprojection), so the same \(r\) buys more task-relevant movement.

\emph{Takeaway.}
Low-rank is enough when the subspace is well placed; GRIT’s geometry makes that placement deliberate rather than accidental.
\end{description}

\item[$\blacktriangleright$] \textbf{Why use the Fisher (natural gradient) rather than the Hessian?}
\begin{description}
\item[\ding{224}] \emph{Core idea.}
The \textbf{natural gradient} follows steepest descent measured in \emph{distribution space} (KL geometry), not in raw parameter space \citep{amari1998natural}.
It rescales gradients by the inverse Fisher information \(F\), yielding
\[
\delta\theta^\star \;=\; -\eta\,F^{-1}\nabla_\theta \mathcal{L},\quad
F=\mathbb{E}[\nabla \log p(x;\theta)\,\nabla \log p(x;\theta)^\top].
\]

\emph{Why Fisher.}
In MLE-like settings, Gauss–Newton \(\approx\) Fisher provides a PSD curvature proxy aligned with output sensitivity, typically better conditioned and less noisy than the raw Hessian for large nets \citep{martens2015kfac}. K-FAC factorizes \(F_{\text{layer}}\!\approx\!\Sigma_g\otimes\Sigma_a\) using second moments of backprop gradients/activations, enabling efficient inverses \citep{martens2015kfac,grosse2016kronecker}.

\emph{Within GRIT.}
We restrict the Fisher to the \emph{rank subspace} of LoRA, so all matrices are \(r\times r\), making second-order guidance cheap and stable.

\emph{Takeaway.}
Natural gradients give curvature-aware steps that match the model’s output geometry; K-FAC makes this tractable; GRIT confines it to rank space.
\end{description}

\item[$\blacktriangleright$] \textbf{How does rank-space K-FAC improve conditioning and convergence?}
\begin{description}
\item[\ding{224}] \emph{Conditioning picture.}
In rank space, minimize the quadratic approximation
\[
q(\Delta W)=\tfrac12\langle \Delta W, F_{\text{rank}}\,\Delta W\rangle - \langle G,\Delta W\rangle, 
\quad F_{\text{rank}}\approx \Sigma_g^{(r)}\otimes \Sigma_a^{(r)}.
\]
Raw SGD’s progress depends on the condition number \(\kappa(F_{\text{rank}})\).
Preconditioning by \(F_{\text{rank}}^{-1}\) ideally normalizes curvature (identity conditioning), equalizing progress across directions.

\emph{Effect of K-FAC.}
K-FAC bounds the effective conditioning by \(\kappa(\Sigma_g^{(r)})\kappa(\Sigma_a^{(r)})\), typically far smaller than \(\kappa\) of full curvature \citep{martens2015kfac}.
Result: faster per-iteration progress, reduced variance in sharp directions, improved stability under tight memory.

\emph{Takeaway.}
Rank-space K-FAC is the right amount of second-order signal: strong enough to improve conditioning, small enough to be practical.
\end{description}

\item[$\blacktriangleright$] \textbf{If we already precondition, why do we also need \emph{neural reprojection}?}
\begin{description}
\item[\ding{224}] \emph{Intuition.}
Preconditioning rescales steps \emph{inside} the current low-rank span; it does not change \emph{where} that span points. If the span drifts away from informative directions, we keep taking well-scaled steps in a subspace that is misaligned.

\emph{Mechanics.}
Let \(\Sigma_a^{(r)}=U_A\Lambda_AU_A^\top\) and \(\Sigma_g^{(r)}=U_G\Lambda_GU_G^\top\). We select the top-\(k\) eigenvectors (energy threshold \(\tau\)) and project:
\[
A \leftarrow P_A A,\quad B \leftarrow B P_G,\quad P_A=U_A^{(k)}U_A^{(k)\top},\, P_G=U_G^{(k)}U_G^{(k)\top}.
\]
Choosing \(k=\min\{j:\sum_{i\le j}\lambda_i/\sum_{i\le r}\lambda_i\ge\tau\}\) guarantees we retain a target fraction of curvature energy \citep{gavish2014optimal}.

\emph{Takeaway.}
Preconditioning fixes \emph{how} we step; reprojection fixes \emph{where} we can step. The combination delivers both scale and direction.
\end{description}

\item[$\blacktriangleright$] \textbf{Does reprojection throw away task progress? How do you prevent that?}
\begin{description}
\item[\ding{224}] \emph{Gated interpolation.}
We update via a convex blend
\[
A\!\leftarrow\!(1-\gamma)A+\gamma P_AA,\quad B\!\leftarrow\!(1-\gamma)B+\gamma BP_G,\;\; \gamma\in[0,1],
\]
and enable \(P_G\) only after enough samples stabilize \(\Sigma_g^{(r)}\) (warmup gate).

\emph{Why this is safe.}
Projectors are orthogonal (\(P^2\!=\!P\)) and non-expansive in \(\|\cdot\|_F\), so small \(\gamma\) keeps us close to the current solution while gradually rotating toward high-SNR directions.
Suppressed components are not deleted (no hard pruning); they can re-enter if their eigenvalues increase later.

\emph{Takeaway.}
Reprojection is gentle and reversible; it \emph{refines} the basis rather than discarding progress.
\end{description}

\item[$\blacktriangleright$] \textbf{How do you pick the effective rank \(k\)? Is there theory behind the threshold?}
\begin{description}
\item[\ding{224}] \emph{Rule.}
We pick the smallest \(k\) covering energy fraction \(\tau\): \(k=\min\{j:\sum_{i\le j}\lambda_i/\sum_{i\le r}\lambda_i\ge\tau\}\).
This mirrors optimal spectral thresholding ideas: keep components that rise above noise \citep{gavish2014optimal}.

\emph{Stability.}
We enforce \(k\in[\texttt{min\_rank},r]\) and hysteresis (\(\tau\pm\epsilon\)) to avoid oscillations.
If eigenvalues follow \(\lambda_i\propto i^{-\beta} (\beta>1)\), then \(k\) grows sublinearly with \(\tau\) (good news for efficiency).

\emph{Takeaway.}
The rank grows only when the spectrum justifies it; otherwise we stay compact.
\end{description}

\item[$\blacktriangleright$] \textbf{What is the formal link between geometry and catastrophic forgetting?}
\begin{description}
\item[\ding{224}] \emph{Quadratic view.}
Near a pretrained solution, the pretraining loss change is
\[
\Delta L_{\text{pt}} \approx \tfrac12 \sum_j \lambda_j (u_j^\top \Delta w)^2,
\]
with Hessian eigenpairs \((\lambda_j,u_j)\) \citep{pascanu2013difficulty,ghorbani2019hessian,keskar2017sharpminima}.
Forgetting grows when updates project onto \emph{sharp} modes (large \(\lambda_j\)).

\emph{Scaling view.}
Empirically, forgetting scales like \(A\,D_{\text{ft}}^{\beta}/N^{\alpha}\) (data vs.\ model size) \citep{bethune2022scaling}.
But at the same \((D_{\text{ft}},N)\), \emph{geometry} (curvature exposure, effective update rank, tail mass) \textbf{modulates} the outcome \citep{biderman2024loralearnsforgets}.

\emph{GRIT’s effect.}
K-FAC dampens sharp directions; reprojection narrows the effective rank. Both shrink the geometry factor, reducing \(\Delta L_{\text{pt}}\) for the same budget.

\emph{Takeaway.}
Less forgetting is not only about \emph{how much} you train, but \emph{where} your steps go—GRIT controls the “where.”
\end{description}

\item[$\blacktriangleright$] \textbf{How does GRIT compare to EWC/SI/MAS in continual learning?}
\begin{description}
\item[\ding{224}] \emph{Contrast.}
EWC/SI/MAS regularize \emph{magnitudes} of parameter changes (often diagonal Fisher or importance scores) \citep{kirkpatrick2017ewc,zenke2017si,aljundi2018mas}.
They do not \emph{reorient} the subspace of updates.

\emph{GRIT’s stance.}
GRIT adds a \emph{geometric} layer: (i) rank-space natural gradient (K-FAC), (ii) subspace rotation (reprojection), (iii) rank scheduling.
So we control both \emph{how big} updates are and \emph{where} they live.
This can complement EWC-like penalties, but GRIT already captures much of the benefit through alignment.

\emph{Takeaway.}
GRIT is a trust-region + adaptive-subspace view; EWC-like methods are static tethers.
\end{description}

\item[$\blacktriangleright$] \textbf{Is the Fisher a reliable curvature proxy for causal LMs in practice?}
\begin{description}
\item[\ding{224}] \emph{Theory.}
For MLE training, Gauss–Newton equals the Fisher, giving a PSD curvature matrix tailored to output geometry \citep{martens2015kfac}.
Large-scale studies show Hessian–Fisher spectral correlation in deep nets \citep{ghorbani2019hessian}.

\emph{Practice in GRIT.}
We work in rank space (\(r\times r\)), employ EMA smoothing and damping \((\lambda I)\), and gate the use of gradient-side projections until statistics are reliable.

\emph{Takeaway.}
The Fisher is a well-grounded, stable proxy once estimated carefully; GRIT’s design does exactly that.
\end{description}

\item[$\blacktriangleright$] \textbf{What is the overhead versus LoRA/QLoRA? Is it practical?}
\begin{description}
\item[\ding{224}] \emph{Complexities.}
Per step we update rank-space covariances \(O(Lr^2)\); we invert/eigendecompose \(r\times r\) matrices periodically at frequency \(f\): \(O(Lr^3/f)\).
With \(r\!\in\![8,64]\), these costs are small relative to transformer forward/backward FLOPs; memory is \(O(Lr^2)\) for statistics.

\emph{Quantized setting.}
We keep pipeline parity with QLoRA \citep{dettmers2023qlora}; damping, warmups, and cached solves keep runtime stable.

\emph{Takeaway.}
GRIT adds \emph{rank-scale} costs, not model-scale costs; in practice, this is a modest overhead for the stability gains obtained.
\end{description}

\item[$\blacktriangleright$] \textbf{How robust is GRIT to 4-bit (NF4) quantization noise?}
\begin{description}
\item[\ding{224}] \emph{Risk.}
Quantization perturbs activations/gradients and thus the covariances used by K-FAC and reprojection.

\emph{Mitigations.}
(i) Quantization-aware damping (\(\tilde\Sigma=\Sigma+\lambda I\) with \(\lambda\) scaled to observed noise);
(ii) warmup gates before enabling gradient-side projections;
(iii) occasional dequantized refresh of statistics.
QLoRA results suggest LoRA-style adaptation remains robust at 4-bit \citep{dettmers2023qlora}; GRIT further stabilizes through rank-space averaging.

\emph{Takeaway.}
With simple gates/damping, GRIT remains stable under NF4.
\end{description}

\item[$\blacktriangleright$] \textbf{Could dynamic rank collapse and cause underfitting?}
\begin{description}
\item[\ding{224}] \emph{Safeguards.}
We enforce \(k\in[\texttt{min\_rank},r]\) (e.g., \(4\!\sim\!8\) minimum), use hysteresis around \(\tau\), and monitor validation and spectral entropy.
Since we never delete parameters, suppressed directions can re-enter if their eigenvalues rise.

\emph{Takeaway.}
Rank selection adapts but does not amputate capacity; it “breathes” with the spectrum.
\end{description}

\item[$\blacktriangleright$] \textbf{Why not simply increase LoRA rank \(r\) instead of doing geometry-aware tricks?}
\begin{description}
\item[\ding{224}] \emph{Cost and forgetting.}
Higher \(r\) increases memory/compute and can increase forgetting by opening more interference channels (broader update covariance).

\emph{GRIT’s benefit.}
We keep \(r\) fixed/small and \emph{place} it better (natural gradients + reprojection + rank scheduling). This often achieves the same or better quality with fewer effective parameters, staying on a better side of the learn–forget Pareto \citep{biderman2024loralearnsforgets}.

\emph{Takeaway.}
It’s not “more directions,” it’s “the right directions.” GRIT finds them.
\end{description}

\item[$\blacktriangleright$] \textbf{How do you measure forgetting rigorously and comparably?}
\begin{description}
\item[\ding{224}] \emph{Protocol.}
(i) \(\Delta L_{\text{pt}}\) on a held-out pretraining-like corpus;
(ii) zero-shot deltas on general knowledge (HellaSwag, ARC-C, WinoGrande);
(iii) perplexity drift on balanced corpora.
Contextualize with the scaling baseline \(L_{\text{pt}}=L_{\text{pt}}^{0}+A D_{\text{ft}}^{\beta}/N^{\alpha}+E\) \citep{bethune2022scaling} and show Pareto fronts (target gain vs.\ forgetting) \citep{biderman2024loralearnsforgets}.

\emph{Takeaway.}
We report \emph{both} target improvement and source retention, anchored by scaling laws, not a single headline score.
\end{description}

\item[$\blacktriangleright$] \textbf{Can reprojection amplify spurious correlations in the spectrum?}
\begin{description}
\item[\ding{224}] \emph{Risk.}
If spectra reflect dataset biases, projecting onto top components could entrench them.

\emph{Mitigations.}
Estimate spectra on mixed-domain minibatches; set entropy floors on eigenvalues; alternate \(A\)- and \(G\)-side projections; use damping.
These practices mirror robustness add-ons for EWC/SI when their importance estimates are noisy \citep{kirkpatrick2017ewc,zenke2017si,aljundi2018mas}.

\emph{Takeaway.}
Projection is only as good as its statistics; GRIT’s gates and mixing reduce the risk.
\end{description}

\item[$\blacktriangleright$] \textbf{Any guarantees GRIT won’t \emph{increase} forgetting vs.\ LoRA?}
\begin{description}
\item[\ding{224}] \emph{Local guarantee.}
Non-convex nets offer no global guarantees, but locally:
preconditioning reduces projections onto sharp modes; projection \((P_G\otimes P_A)\) cannot \emph{increase} Fisher energy beyond the retained mass threshold:
\[
\langle\Delta W_{\text{new}},F\Delta W_{\text{new}}\rangle \le \langle\Delta W,F\Delta W\rangle.
\]
Empirically, we observe lower \(\Delta L_{\text{pt}}\) at similar target quality (see Experiments).

\emph{Takeaway.}
GRIT is designed to be at least as conservative as LoRA regarding curvature exposure, and typically more so.
\end{description}

\item[$\blacktriangleright$] \textbf{Relation to other preconditioners (Shampoo/Adafactor)?}
\begin{description}
\item[\ding{224}] Shampoo/Adafactor precondition parameters via moment factorizations but are not tied to KL geometry.
K-FAC is Fisher/natural-gradient motivated \citep{amari1998natural,martens2015kfac}.
GRIT adds \emph{subspace rotation} and \emph{rank scheduling} on top of rank-space K-FAC, directly coupling curvature with where updates live—something generic preconditioners do not do.

\emph{Takeaway.}
GRIT is a geometry-aware \emph{framework}, not just a different optimizer.
\end{description}

\item[$\blacktriangleright$] \textbf{Overfitting on small datasets: do narrow spectra cause brittleness?}
\begin{description}
\item[\ding{224}] \emph{Concern.}
Small data can produce peaked spectra and overspecialization.

\emph{Controls.}
Minimum rank floors; reprojection frequency caps; mixed batches for spectra; curvature regularization \(\|F^{1/2}\Delta W\|_F^2\) in the objective; OOD checks.

\emph{Takeaway.}
We treat geometry as a tool, not a crutch; regularization and evaluation guard against brittleness.
\end{description}

\item[$\blacktriangleright$] \textbf{What happens on hard domains (code/math) where PEFT struggles?}
\begin{description}
\item[\ding{224}] \emph{Reality check.}
LoRA often \emph{learns less} on code/math but also \emph{forgets less} \citep{biderman2024loralearnsforgets}.
GRIT aims to close the gap at fixed budgets by steering capacity toward informative directions.
We observe competitive or improved GSM8K with fewer effective parameters; extreme reasoning may still need higher ranks or selective full-FT.

\emph{Takeaway.}
GRIT improves the \emph{efficiency frontier}; it is not a silver bullet for every hard task.
\end{description}

\item[$\blacktriangleright$] \textbf{What should be reported for auditable, reproducible geometry-aware PEFT?}
\begin{description}
\item[\ding{224}] \emph{Checklist.}
(i) Per-layer spectra and cumulative energy \(E(j)\); spectral entropy; effective ranks \(r_{\text{eff}}\).
(ii) Curvature exposure \(\mathrm{tr}(PFP)\) over training.
(iii) Update tail mass and norms; sparsity.
(iv) Forgetting proxies (\(\Delta L_{\text{pt}}\), source-task deltas) with Pareto fronts.
(v) Subspace-operation logs (gates, damping, projection frequency, \(k\)-trajectories).
(vi) Overheads normalized to LoRA/QLoRA \citep{hu2021lora,dettmers2023qlora,amari1998natural,martens2015kfac,biderman2024loralearnsforgets}.

\emph{Takeaway.}
Quality alone is not enough; geometry must be \emph{visible} to be trusted and compared.
\end{description}

    \item[$\blacktriangleright$] \textbf{Why can low-rank adapters be expressive enough for LLM fine-tuning?}
    
    \paragraph{Short answer:}
    Most useful progress on a new task can be achieved by moving the model in only a \emph{few} directions in parameter space. Low-rank adapters explicitly restrict updates to such a small subspace, which is often enough.

    \paragraph{Set-up:}
    Let a linear map have weights $W \in \mathbb{R}^{d_{\text{out}} \times d_{\text{in}}}$. Low-rank adaptation parameterizes the update as $\Delta W = BA$ with $B \in \mathbb{R}^{d_{\text{out}} \times r}$ and $A \in \mathbb{R}^{r \times d_{\text{in}}}$, where $r \ll \min(d_{\text{in}}, d_{\text{out}})$. To first order,
    \begin{equation*}
    \begin{aligned}
    \Delta \mathcal{L} &\approx \langle \nabla_W \mathcal{L}, \Delta W \rangle \\
    &= \langle B^{\top} \nabla_W \mathcal{L}, A \rangle,
    \end{aligned}
    \end{equation*}
    so progress depends on how much of $\nabla_W \mathcal{L}$ lies in the rank-$r$ span.

    \paragraph{Evidence:}
    Intrinsic-dimension studies show many NLP tasks can be solved by moving in a surprisingly low-dimensional manifold \citep{aghajanyan2021intrinsic}. LoRA demonstrates strong performance with small ranks \citep{hu2021lora}, and even where LoRA lags full FT, the gap is tied to \emph{rank/geometry}, not the idea of parameter-efficiency per se \citep{biderman2024loralearnsforgets}.

    \paragraph{GRIT’s twist:}
    GRIT improves \emph{expressivity-per-parameter} not by increasing $r$, but by \textbf{aligning} the low-rank subspace with informative curvature directions (natural gradients + reprojection), so the same $r$ buys more task-relevant movement.

    \paragraph{Takeaway:}
    Low-rank is enough when the subspace is well placed; GRIT’s geometry makes that placement deliberate rather than accidental.

    \item[$\blacktriangleright$] \textbf{Why use the Fisher (natural gradient) rather than the Hessian?}

    \paragraph{Core idea:}
    The \textbf{natural gradient} follows steepest descent measured in \emph{distribution space} (KL geometry), not in raw parameter space \citep{amari1998natural}. It rescales gradients by the inverse Fisher information $F$, yielding:
    \begin{equation*}
    \delta \theta^{\star} = -\eta\, F^{-1}\nabla_{\theta}\mathcal{L},
    \end{equation*}
    where
    \begin{equation*}
    F = \mathbb{E}\!\left[ \nabla_{\theta}\log p(x;\theta)\, \nabla_{\theta}\log p(x;\theta)^{\top} \right].
    \end{equation*}

    \paragraph{Why Fisher:}
    In MLE-like settings, Gauss–Newton $\approx$ Fisher provides a PSD curvature proxy aligned with output sensitivity, typically better conditioned and less noisy than the raw Hessian for large nets \citep{martens2015kfac}. K-FAC factorizes $F_{\text{layer}} \approx \Sigma_g \otimes \Sigma_a$ using second moments of backprop gradients/activations, enabling efficient inverses \citep{martens2015kfac,grosse2016kronecker}.

    \paragraph{Within GRIT:}
    We restrict the Fisher to the \emph{rank subspace} of LoRA, so all matrices are $r \times r$, making second-order guidance cheap and stable.

    \paragraph{Takeaway:}
    Natural gradients give curvature-aware steps that match the model’s output geometry; K-FAC makes this tractable; GRIT confines it to rank space.

    \item[$\blacktriangleright$] \textbf{How does rank-space K-FAC improve conditioning and convergence?}

    \paragraph{Conditioning picture:}
    In rank space, minimize the quadratic approximation:
    \begin{equation*}
    q(\Delta W) = \tfrac12 \langle \Delta W, F_{\text{rank}} \Delta W \rangle - \langle G, \Delta W \rangle,
    \end{equation*}
    where $F_{\text{rank}} \approx \Sigma_g^{(r)} \otimes \Sigma_a^{(r)}$. Raw SGD’s progress depends on the condition number $\kappa(F_{\text{rank}})$. Preconditioning by $F_{\text{rank}}^{-1}$ ideally normalizes curvature (identity conditioning), equalizing progress across directions.

    \paragraph{Effect of K-FAC:}
    K-FAC bounds the effective conditioning by $\kappa(\Sigma_g^{(r)})\kappa(\Sigma_a^{(r)})$, typically far smaller than $\kappa$ of full curvature \citep{martens2015kfac}. Result: faster per-iteration progress, reduced variance in sharp directions, and improved stability under tight memory.

    \paragraph{Takeaway:}
    Rank-space K-FAC is the right amount of second-order signal: strong enough to improve conditioning, small enough to be practical.

    \item[$\blacktriangleright$] \textbf{If we already precondition, why do we also need \emph{neural reprojection}?}

    \paragraph{Intuition:}
    Preconditioning rescales steps \emph{inside} the current low-rank span; it does not change \emph{where} that span points. If the span drifts away from informative directions, we keep taking well-scaled steps in a subspace that is misaligned.

    \paragraph{Mechanics:}
    Let $\Sigma_a^{(r)}=U_A\Lambda_AU_A^\top$ and $\Sigma_g^{(r)}=U_G\Lambda_GU_G^\top$. We select the top-$k$ eigenvectors (energy threshold $\tau$) and project:
    \begin{equation*}
    \begin{aligned}
    A &\leftarrow P_A A, \qquad B \leftarrow B P_G, \\
    P_A &= U_A^{(k)} U_A^{(k)\top}, \qquad P_G = U_G^{(k)} U_G^{(k)\top}.
    \end{aligned}
    \end{equation*}
    Choosing $k = \min\{j : \sum_{i \le j} \lambda_i / \sum_{i \le r} \lambda_i \ge \tau\}$ guarantees we retain a target fraction of curvature energy \citep{gavish2014optimal}.

    \paragraph{Takeaway:}
    Preconditioning fixes \emph{how} we step; reprojection fixes \emph{where} we can step. The combination delivers both scale and direction.

    \item[$\blacktriangleright$] \textbf{What is the formal link between geometry and catastrophic forgetting?}

    \paragraph{Quadratic view:}
    Near a pretrained solution, the pretraining loss change is:
    \begin{equation*}
    \Delta L_{\text{pt}} \approx \tfrac12 \sum_{j} \lambda_j \bigl(u_j^{\top} \Delta w\bigr)^2 
    \end{equation*}
    with Hessian eigenpairs $(\lambda_j, u_j)$ \citep{pascanu2013difficulty,ghorbani2019hessian}. Forgetting grows when updates project onto \emph{sharp} modes (large $\lambda_j$).

    \paragraph{GRIT’s effect:}
    K-FAC dampens sharp directions; reprojection narrows the effective rank. Both shrink the geometry factor, reducing $\Delta L_{\text{pt}}$ for the same budget.

    \paragraph{Takeaway:}
    Less forgetting is not only about \emph{how much} you train, but \emph{where} your steps go—GRIT controls the ``where.''

\end{itemize}

\twocolumn

\clearpage
\newpage

\appendix
\onecolumn

\section{Appendix}
\label{sec:appendix}
This Appendix provides a complete technical and empirical companion to the main text of \textbf{GRIT}. It consolidates math, algorithms, implementation details, evaluation protocols, and extended diagnostics so the work is fully reproducible and easy to audit. Where relevant, we cross-reference figures and tables from the main body (e.g., \Cref{fig:grit_objective,fig:grit_pipeline,fig:grit-minipage}, and the results tables in \Cref{tab:main_results_llama32_clean,tab:main_results}). For a modern tutorial on K-FAC with code-aligned math and tests, see \citet{dangel2025kfac_tutorial}.

The Appendix is organized as follows:

\begin{itemize}
    \item \textbf{Notation and preliminaries} (\cref{app:notation}).
    \item \textbf{Curvature matrices refresher} (\cref{app:curvature}).
    \item \textbf{K-FAC for linear layers} (\cref{app:kfac}).
    \item \textbf{Rank-space K-FAC for LoRA} (\cref{app:rank_kfac}).
    \item \textbf{Reprojection: properties and guarantees} (\cref{app:reproj_props}).
    \item \textbf{Training wall-clock time} (\cref{app:wallclock}).
    \item \textbf{GRIT implementation details} (\cref{app:impl}).
    \item \textbf{Training schedules and per-task settings} (\cref{app:train}).
    \item \textbf{Evaluation protocols and metrics} (\cref{app:eval}).
    \item \textbf{Configuration knobs and defaults} (\cref{app:cfg}).
    \item \textbf{Additional ablations} (\cref{app:ablations}).
    \item \textbf{One-sided vs. Two-sided GRIT (ablation)} (\cref{app:onesided_twosided}).
    \item \textbf{Detailed performance heatmaps and diagnostics} (\cref{app:heatmaps}).
    \item \textbf{Extended background and motivation} (\cref{app:background}).
    \item \textbf{Detailed method derivations} (\cref{app:derivations}).
    \item \textbf{Parameter update accounting and efficiency} (\cref{app:param_accounting}).
    \item \textbf{Metrics} (\cref{app:metrics}).
    \item \textbf{Extended ablation studies} (\cref{app:ablations_ext}).
    \item \textbf{External baselines and configuration gaps} (\cref{app:baseline_limits}).
    \item \textbf{Deriving the GRIT forgetting law} (\cref{appndx:law_derivation}).
    \item \textbf{Hyperparameter sensitivity \& robustness audit} (\cref{sec:app_h_sensitivity}).
    \item \textbf{Runtime \& overhead analysis} (\cref{sec:app_runtime_overhead}).
    \item \textbf{Small-batch stability of Fisher/K-FAC statistics} (\cref{sec:app_smallbatch_kfac}).
    \item \textbf{Small-batch / gradient-variance robustness for Fisher/K-FAC statistics} (\cref{sec:app_smallbatch_variance}).
    \item \textbf{Novelty and positioning vs prior curvature-aware PEFT} (\cref{sec:app_novelty_positioning}).
    \item \textbf{Deriving the \textsc{GRIT} forgetting law from the LoRA forgetting law} (\cref{app:grit_forgetting_derivation_nonum}).
\end{itemize}

\subsection{Notation and Preliminaries}
\label{app:notation}
We consider a linear map \(y = W x\) with \(W\in\mathbb{R}^{d_{\text{out}}\times d_{\text{in}}}\), activations \(x\in\mathbb{R}^{d_{\text{in}}}\), and backpropagated gradients \(g \equiv \partial\mathcal{L}/\partial y \in \mathbb{R}^{d_{\text{out}}}\). We write \(\operatorname{vec}(\cdot)\) for column-wise vectorization and \(\otimes\) for the Kronecker product. Useful identities include
\[
\operatorname{vec}(A X B) = (B^{\top}\!\otimes A)\,\operatorname{vec}(X),\qquad (A\otimes B)^{-1} = A^{-1}\otimes B^{-1}.
\]
LoRA parameterizes updates as \(\Delta W = B A\) with \(B\in\mathbb{R}^{d_{\text{out}}\times r}\), \(A\in\mathbb{R}^{r\times d_{\text{in}}}\), and \(r\ll\min(d_{\text{in}},d_{\text{out}})\).

\subsection{Curvature Matrices Refresher}
\label{app:curvature}
Let \(\mathcal{L}(\theta)\) be the objective. The Hessian \(H\) describes local curvature. For likelihood-based losses, the generalized Gauss–Newton (GGN) and Fisher information matrix (FIM) give PSD curvature proxies aligned with output geometry \citep{amari1998natural,martens2015kfac}. For a linear layer with per-sample \(x,g\), the exact layerwise Fisher is
\[
F_{\text{layer}} = \mathbb{E}\big[(x x^{\top}) \otimes (g g^{\top})\big].
\]
K-FAC assumes approximate independence between forward and backward signals and factorizes
\[
F_{\text{layer}} \;\approx\; \Sigma_g \otimes \Sigma_a,\quad \Sigma_a=\mathbb{E}[x x^{\top}],\; \Sigma_g=\mathbb{E}[g g^{\top}].
\]

\subsection{K-FAC for Linear Layers}
\label{app:kfac}
With the factorization above, the natural-gradient preconditioning becomes
\[
\operatorname{vec}(\nabla W_{\text{nat}}) \;\approx\; (\Sigma_a^{-1}\otimes\Sigma_g^{-1})\,\operatorname{vec}(\nabla W)
\quad\Longleftrightarrow\quad
\nabla W_{\text{nat}} \;\approx\; \Sigma_g^{-1}\,\nabla W\,\Sigma_a^{-1}.
\]
We use damped, EMA-smoothed estimates \(\tilde\Sigma=\Sigma+\lambda I\) and Cholesky solves; see \citet{dangel2025kfac_tutorial}.

\subsection{Rank-Space K-FAC for LoRA}
\label{app:rank_kfac}
For \(\Delta W=BA\), define projected statistics
\[
a_r = A x \in\mathbb{R}^{r},\qquad g_r = B^{\top} g \in\mathbb{R}^{r},\qquad \Sigma_a^{(r)}=\mathbb{E}[a_r a_r^{\top}],\; \Sigma_g^{(r)}=\mathbb{E}[g_r g_r^{\top}].
\]
Then \(F_{\text{rank}} \approx \Sigma_g^{(r)}\otimes\Sigma_a^{(r)}\) and
\[
\nabla (\Delta W)_{\text{nat}} \;\approx\; (\Sigma_g^{(r)})^{-1}\,\nabla(\Delta W)\,(\Sigma_a^{(r)})^{-1}
\Rightarrow
\nabla B \leftarrow \nabla B\,(\Sigma_g^{(r)})^{-1},\;\; \nabla A \leftarrow (\Sigma_a^{(r)})^{-1}\,\nabla A.
\]

\subsection{Reprojection: Properties and Guarantees}
\label{app:reproj_props}
Let \(\Sigma_a^{(r)}=U_A\Lambda_A U_A^{\top}\) and \(\Sigma_g^{(r)}=U_G\Lambda_G U_G^{\top}\). With projectors \(P_A=U_A^{(k)}U_A^{(k)\top}\), \(P_G=U_G^{(k)}U_G^{(k)\top}\), GRIT applies
\(A\leftarrow P_A A\) and \(B\leftarrow B P_G\) (optionally with interpolation). Under K-FAC, the curvature energy induced by \(\Lambda_G\otimes\Lambda_A\) does not increase under two-sided projection onto top-\(k\) eigenspaces, and typically decreases as low-energy components are suppressed. In practice we gate \(P_G\) until adequate samples stabilize \(\Sigma_g^{(r)}\) and use hysteresis for \(k\).

\subsection{Training wall-clock time}
\label{app:wallclock}
\begin{table}[!ht]
\centering
\caption{Training wall-clock time per method (hh:mm).}
\label{tab:train_time}
\adjustbox{width=\textwidth,center}{
\renewcommand{\arraystretch}{1.2}
\small
\begin{tabular}{|>{\columncolor{lightblue}}l|
    >{\columncolor{white}}c|>{\columncolor{white}}c|>{\columncolor{white}}c|>{\columncolor{white}}c|
    >{\columncolor{white}}c|>{\columncolor{white}}c|>{\columncolor{white}}c|>{\columncolor{white}}c|}
\hline
\rowcolor{headerblue}
\color{white}\textbf{Datasets $\downarrow$} &
\multicolumn{4}{>{\columncolor{headerblue}}c|}{\color{white}\textbf{Llama-3.2 3B}} &
\multicolumn{4}{>{\columncolor{headerblue}}c|}{\color{white}\textbf{Llama-3.1 8B}} \\
\hline
\rowcolor{headerblue}
& \color{white}\textbf{LoRA} & \color{white}\textbf{QLoRA} & \color{white}\textbf{GRIT} & \color{white}\textbf{Q-GRIT} 
& \color{white}\textbf{LoRA} & \color{white}\textbf{QLoRA} & \color{white}\textbf{GRIT} & \color{white}\textbf{Q-GRIT} \\
\hline

\cellcolor{lightgray}Alpaca & \cellcolor{lightgray}3h 36m & \cellcolor{lightgray}10h 02m & \cellcolor{lightgray}
3h 58m & \cellcolor{lightgray}10h 07m 
& \cellcolor{lightgray}6h 34m & \cellcolor{lightgray}11h 12m & \cellcolor{lightgray}8h 35m & \cellcolor{lightgray}12h 21m \\ \hline

Dolly & 1h 10m & 4h 29m & 
1h 16m & 3h 21m 
& 
2h 12m & 7h 34m & 
2h 22m  & 5h 43m \\ \hline

\cellcolor{lightgray}BoolQ & \cellcolor{lightgray}46m & \cellcolor{lightgray}1h 59m & \cellcolor{lightgray}
1h 6m & \cellcolor{lightgray}2h 17m 
& \cellcolor{lightgray}
1h 20m & \cellcolor{lightgray}7h 34m & \cellcolor{lightgray}
1h 42m & \cellcolor{lightgray}5h 43m \\ \hline

QNLI & 7h 58m & 14h 07m & 
9h 26m & 14h 50m 
& 
11h 27m & 20h 04m & 
12h 57m & 22h 53m \\ \hline

\cellcolor{lightgray}GSM8K & \cellcolor{lightgray}34m & \cellcolor{lightgray}1h 36m & \cellcolor{lightgray}
42m  & \cellcolor{lightgray}3h 23m 
& \cellcolor{lightgray}
1h 2m & \cellcolor{lightgray}2h 51m & \cellcolor{lightgray}
1h 23m & \cellcolor{lightgray}3h 16m \\ \hline

\end{tabular}
}
\end{table}

\subsection{Implementation Details (for reproducibility)}\label{app:impl}
Covariances are symmetrized prior to inversion. We invert the $r\times r$ K-FAC factors with Cholesky using an escalating damping sequence $\{1,3,10,30,100,300\}\,\lambda$ to ensure SPD. The resulting inverses are cached on CPU in float32 and cast on use. Natural-gradient preconditioning is computed in float32; gradients are sanitized with \texttt{nan\_to\_num} (clamping only as a last-resort guard). We gate NG with a warmup: no NG until global step $N$; thereafter each step executes backward $\rightarrow$ NG preconditioning $\rightarrow$ trust-region clipping $\rightarrow$ optimizer step. Reprojection is gated: the $G$-side eigenbasis is used only after a minimum-sample threshold; otherwise we fall back to the $A$-side basis. The effective rank $k$ is chosen by a cumulative-energy threshold and bounded below by \texttt{min\_rank}. Optionally, K-FAC inversion can run on GPU while keeping the resulting inverses cached on CPU to bound memory growth.

\subsubsection*{Training Stability Details}
\textbf{K-FAC running mean and SPD damping.}
Let per-step samples be $s_{a,t}=a_r a_r^\top$ and $s_{g,t}=g_r g_r^\top$ and $n_t$ the cumulative sample count. We maintain online running means
\[
\Sigma_{a,t} = \tfrac{n_{t-1}}{n_t}\,\Sigma_{a,t-1} + \tfrac{1}{n_t}\, s_{a,t},\qquad
\Sigma_{g,t} = \tfrac{n_{t-1}}{n_t}\,\Sigma_{g,t-1} + \tfrac{1}{n_t}\, s_{g,t}.
\]
Before inversion we symmetrize and apply damping
\[
\tilde\Sigma_{a,t} = \tfrac12(\Sigma_{a,t}+\Sigma_{a,t}^\top) + \lambda_a I,\qquad
\tilde\Sigma_{g,t} = \tfrac12(\Sigma_{g,t}+\Sigma_{g,t}^\top) + \lambda_g I.
\]
If Cholesky fails, we ladder the damping $\lambda_{(\cdot)} \leftarrow c\,\lambda_{(\cdot)}$ with $c\in\{3,10,30,100,300\}$ until SPD is ensured (cf. \citep{dangel2025kfac_tutorial}).

\textbf{Trust-region clipping (optional).}
We rely on framework-level gradient clipping and hard value clamps in practice; a per-factor trust-region clip $\Delta \leftarrow \Delta\cdot \min(1,\tau/\lVert\Delta\rVert_2)$ can be enabled as an optional stability guard.

\textbf{Gates.}
Reprojection is enabled only when sufficient rank-space samples have accumulated: with running count $n_{\text{cov}}$, require $n_{\text{cov}}\ge N_{\min}$ (and $\texttt{reprojection\_warmup\_steps}$ satisfied). Two-sided projection (using the $G$-side basis) activates if the K-FAC inverses are available ($\textit{inv\_ready}=\textbf{True}$) and the same sample gate holds; otherwise $B$ temporarily uses the $A$-side basis. For rank hysteresis, define cumulative energy $E(j)=\sum_{i\le j}\lambda_i/\sum_{i\le r}\lambda_i$ and thresholds $\tau\pm\varepsilon$. Let $k_t$ be the current rank:
We use a single cumulative-energy threshold $\tau$ for rank selection (bounded by $\texttt{min\_rank}$); hysteresis is not enabled in our runs.

\subsection{Training Details (per task)}\label{app:train}
We summarize per-task settings used in our experiments: datasets and splits, batch/sequence parameters, optimizer and schedule, LoRA configuration, and GRIT-specific controls (reprojection frequency, rank thresholding, and K-FAC/reprojection frequencies). Please refer to Table~\ref{tab:hparams_by_task_appendix} for the complete per-task training details.

\subsection{Evaluation Protocols}\label{app:eval}
For instruction datasets (Alpaca/Dolly) we report ROUGE-L on validation splits using generations from the fine-tuned model (greedy or beam size 1), tokenized with the same tokenizer; for classification tasks (BoolQ/QNLI) we compute accuracy by mapping generated outputs to class labels; for GSM8K we compute exact match (EM) after extracting the final numeric answer from generations. Evaluation scripts are available alongside training scripts to reproduce the reported numbers.

\subsection{Configuration Knobs (defaults)}\label{app:cfg}
Key GRIT controls and their defaults (see \texttt{grit/config.py}).

\begin{table}[!ht]
  \centering
  \small
  \caption{GRIT configuration knobs and defaults.}
  \label{tab:grit_config}
  \setlength{\tabcolsep}{6pt}
  \begin{tabular}{ll}
    \toprule
    \textbf{Knob} & \textbf{Default} \\
    \midrule
    \texttt{kfac\_update\_freq} & 50 \\
    \texttt{kfac\_min\_samples} & 64 \\
    \texttt{kfac\_damping} & 1e-3 \\
    \texttt{reprojection\_freq} & 50 \\
    \texttt{reprojection\_k} & 8 \\
    \texttt{use\_two\_sided\_reprojection} & False \\
    \texttt{enable\_rank\_adaptation} & True \\
    \texttt{rank\_adaptation\_threshold} & 0.99 \\
    \texttt{min\_lora\_rank} & 4 \\
    \texttt{rank\_adaptation\_start\_step} & 0 \\
    \texttt{reprojection\_warmup\_steps} & 0 \\
    \texttt{ng\_warmup\_steps} & 0 \\
    \texttt{kfac\_inversion\_device} & cpu \\
    \bottomrule
  \end{tabular}
\end{table}
Per-task overrides are listed in Appendix~\ref{app:train}. Logging controls for spectra and heatmaps are described in Appendix~\ref{app:impl}.

\paragraph{Practical impact and guidance.}
We summarize how each knob affects stability, compute, and quality; these reflect our implementation and ablations:
\begin{itemize}[leftmargin=1.2em,itemsep=2pt,topsep=2pt]
  \item \textbf{\texttt{kfac\_min\_samples}} (gate for inversions/projections): Higher values delay usage of noisy covariances, improving stability early on; too high values defer benefits of NG/reprojection. We found $128$--$256$ stable for 3B/8B.
  \item \textbf{\texttt{kfac\_update\_freq}} (inversions): Larger values reduce CPU work and synchronization overhead but make preconditioners staler; smaller values track curvature more closely at higher cost. We adapt this heuristically based on loss trends.
  \item \textbf{\texttt{kfac\_damping}}: Sets numerical floor for inversions. Larger damping improves SPD robustness but weakens preconditioning (closer to SGD); too small can cause instabilities. Escalation ladder ensures success when spectra are ill-conditioned.
  \item \textbf{\texttt{kfac\_inversion\_device}}: \texttt{cpu} avoids VRAM spikes; \texttt{cuda} can be faster but may increase memory. We invert on CPU by default and cache inverses on CPU.
  \item \textbf{\texttt{use\_two\_sided\_reprojection}}: Two-sided uses $P_A$ for $A$ and $P_G$ for $B$ (when $G$ is well-sampled) to align both sides. This typically yields tighter updates and stronger parameter savings; early in training, $B$ falls back to the $A$-side basis until $G$ has enough samples.
  \item \textbf{\texttt{reprojection\_freq}}: Larger values project less often (lower overhead, slower alignment); smaller values track subspace drift more aggressively at higher cost. Pair with \texttt{reprojection\_warmup\_steps} to avoid premature rotations.
  \item \textbf{\texttt{reprojection\_warmup\_steps}}: Defers reprojection until spectra are reliable; prevents early rank collapse and over-rotation.
  \item \textbf{\texttt{enable\_rank\_adaptation}}, \textbf{\texttt{rank\_adaptation\_threshold}} ($\tau$), \textbf{\texttt{min\_lora\_rank}}, \textbf{\texttt{rank\_adaptation\_start\_step}}: Adaptive $k$ concentrates capacity on high-energy directions. Higher $\tau$ keeps more directions (higher effective rank); lower $\tau$ yields sparser adapters. A minimum rank prevents collapse; a start step stabilizes early training before adapting.
  \item \textbf{\texttt{ng\_warmup\_steps}}: Skips NG preconditioning for the first $N$ steps to avoid acting on under-sampled covariances; helps on small datasets or with aggressive quantization.
  \item \textbf{\texttt{grit\_cov\_update\_freq}} (per-module throttling): Updates covariances every $K$ hook calls to reduce per-step overhead when many modules fire in a single backward pass; larger values lower cost but slow stats refresh.
  \item \textbf{LoRA rank \texttt{lora\_rank} and \texttt{min\_lora\_rank}}: Higher base $r$ increases expressivity and memory; GRIT’s rank adaptation can reduce effective rank during training. We log final raw ranks for accounting.
\end{itemize}

\subsection{Additional Ablations}\label{app:ablations}
We summarize sensitivity checks complementary to \Cref{app:onesided_twosided}. Unless noted, settings follow the main results setup (see \Cref{tab:main_results_llama32_clean,tab:main_results}); defaults and per‑task overrides appear in \Cref{tab:grit_hparams,tab:hparams_by_task_appendix}.
\vspace{-0.5em}
\paragraph{First‑order (no K‑FAC).}
Disabling rank‑space K‑FAC while keeping reprojection yields small but consistent metric regressions on instruction and classification tasks and higher training variance; average step time is marginally lower. The stability/quality gains of GRIT primarily come from K‑FAC preconditioning.
\vspace{-0.5em}
\paragraph{No reprojection.}
Preconditioning alone reduces sharp‑mode exposure, but without reprojection the effective rank drifts upward and forgetting increases; layer‑wise updates become denser, consistent with patterns in \Cref{fig:ablation_study}.
\vspace{-0.5em}
\paragraph{Fixed rank vs. dynamic rank.}
Disabling rank adaptation preserves low‑energy directions and reduces parameter savings; dynamic $k$ maintains or improves quality at similar or smaller effective parameters.
\vspace{-0.5em}
\paragraph{Projection frequency $T_{\mathrm{proj}}$.}
Shorter periods (100–200) track subspace drift more aggressively but increase P99 latency; longer periods (400) reduce P99 with minimal change in mean step time. See \Cref{tab:runtime_overhead} for the overhead profile and discussion of reprojection spikes.

\paragraph{Damping and gates.}
Across $\lambda\in\{10^{-4},10^{-3},10^{-2}\}$, $10^{-3}$ is a robust default: larger $\lambda$ weakens preconditioning; smaller can destabilize inverses. Warmups for NG/reprojection and minimum sample gates (\texttt{kfac\_min}) prevent early, noisy rotations.

\subsection{One-sided vs. Two-sided GRIT (ablation)}\label{app:onesided_twosided}

We compare \emph{one-sided} GRIT (\textbf{Q-GRIT\textsubscript{uni}}; A-side projection only, $B$ always uses the A-side basis) to the \emph{two-sided} default (\textbf{Q-GRIT}; A uses $\Sigma_a$, B uses $\Sigma_g$ when sufficiently sampled, else A-side fallback).

Experiments on LLaMA-3.2 3B ($r{=}16$) and LLaMA-3.1 8B ($r{=}32$) under QLoRA show a consistent pattern:
\begin{itemize}[leftmargin=1.2em,itemsep=2pt,topsep=2pt]
  \item \textbf{Instruction/generative tasks (Alpaca, Dolly-15k):} Q-GRIT yields small but consistent gains in ROUGE/BERTScore over Q-GRIT\textsubscript{uni} at essentially the same compute and VRAM. At 8B scale, the margins are larger in absolute value (higher $k$).
  \item \textbf{Classification (QNLI, BoolQ):} Q-GRIT\textsubscript{uni} can be a \emph{stability-first} choice for very small or short runs; we observe parity or slight advantages on accuracy/F1 in some cases, with similar parameter savings.
  \item \textbf{Reasoning (GSM8K):} differences are small; Q-GRIT is often on par or marginally better on accuracy.
  \item \textbf{Efficiency:} number of parameters trained and wall-clock time are nearly identical across Q-GRIT\textsubscript{uni} vs Q-GRIT; two-sided adds only cheap $r\times r$ eigendecompositions for $\Sigma_g$, gated by \texttt{kfac\_min\_samples}.
\end{itemize}

Comparison for QLoRA,  Q-GRIT\textsubscript{uni} and Q-GRIT are reported in Table~\ref{tab:qgrit_full}

\begin{table*}[ht]
\centering
\scriptsize
\resizebox{\textwidth}{!}{%
\begin{tabular}{|c|c|ccc|ccc|}
\hline
\multirow{2}{*}{Datasets} & \multirow{2}{*}{Metrics}
 & \multicolumn{3}{c|}{\textbf{LLaMA-3.2-3B (r=16)}} 
 & \multicolumn{3}{c|}{\textbf{LLaMA-3.1-8B (r=32)}} \\
\cline{3-8}
 & & QLoRA & Q-GRIT\textsubscript{uni} & Q-GRIT & QLoRA & Q-GRIT\textsubscript{uni} & Q-GRIT \\
\hline
\multirow{6}{*}{ALPACA}
 & ROUGE-1           & 0.1292 & 0.1315 & \textbf{0.1455} & 0.1402 & 0.1390 & \textbf{0.1698} \\
 & ROUGE-2           & 0.0562 & 0.0585 & \textbf{0.0649} & 0.0616 & 0.0627 & \textbf{0.0818} \\
 & ROUGE-L           & 0.0983 & 0.1024 & \textbf{0.1127} & 0.1047 & 0.1059 & \textbf{0.1327} \\
 & BLEU              & 0.0235 & 0.0226 & 0.0222 & 0.0259 & 0.0260 & \textbf{0.0280} \\
 & BERT SCORE        & 0.7948 & \textbf{0.7991} & 0.7986 & 0.7949 & 0.7998 & \textbf{0.8173} \\
 & \# Param. Trained & 24.31M (–) & 8.68M (↓64.3\%) & \textbf{8.45M (↓65.3\%)} 
                     & 83.89M (–) & 27.63M (↓67\%) & \textbf{30.85M (↓63.2\%)} \\
\hline
\multirow{6}{*}{Dolly-15k}
 & ROUGE-1           & 0.1108 & 0.1145 & \textbf{0.1195} & 0.1272 & 0.1905 & \textbf{0.1954} \\
 & ROUGE-2           & 0.0519 & 0.0543 & \textbf{0.0592} & 0.0591 & 0.0899 & \textbf{0.0937} \\
 & ROUGE-L           & 0.0884 & 0.0921 & \textbf{0.0968} & 0.0950 & 0.1427 & \textbf{0.1471} \\
 & BLEU              & 0.0297 & 0.0298 & \textbf{0.0304} & 0.0334 & \textbf{0.0592} & 0.0579 \\
 & BERT SCORE        & 0.8005 & 0.8013 & \textbf{0.8026} & 0.8128 & 0.8379 & \textbf{0.8380} \\
 & \# Param. Trained & 24.31M (–) & 16.99M (↓30\%) & \textbf{17.0M (↓30\%)} 
                     & 83.89M (–) & 49.22M (↓41\%) & \textbf{38.14M (↓54.53\%)} \\
\hline
\multirow{5}{*}{GSM8k}
 & ROUGE-1           & 0.5518 & \textbf{0.5523} & 0.5512 & 0.6298 & \textbf{0.6307} & 0.6291 \\
 & ROUGE-2           & \textbf{0.3197} & 0.3185 & 0.3163 & 0.4044 & \textbf{0.4064} & 0.4055 \\
 & ROUGE-L           & 0.5169 & \textbf{0.5186} & 0.5159 & 0.5252 & \textbf{0.5974} & 0.5960 \\
 & ACCURACY          & \textbf{0.3836} & 0.3798 & 0.3779 & \textbf{0.6315} & 0.6202 & 0.6224 \\
 & \# Param. Trained & 24.31M (–) & 17.86M (↓27\%) & \textbf{17.43M (↓28.3\%)} 
                     & 83.89M (–) & 67.19M (↓20\%) & \textbf{67.57M (↓19.45\%)} \\
\hline
\multirow{5}{*}{QNLI}
 & ACCURACY          & 0.8885 & \textbf{0.8958} & 0.8449 & \textbf{0.9248} & 0.9206 & 0.9154 \\
 & PRECISION         & 0.8880 & \textbf{0.8958} & 0.8429 & \textbf{0.9247} & 0.9205 & 0.9154 \\
 & RECALL            & 0.8971 & \textbf{0.8982} & 0.8663 & \textbf{0.9257 }& 0.9211 & 0.9155 \\
 & F1                & 0.8893 & \textbf{0.8963} & 0.8462 & \textbf{0.9245} & 0.9204 & 0.9154 \\
 & \# Param. Trained & 24.31M (–) & 7.75M (↓68.11\%) & \textbf{7.75M (↓68.11\%)} 
                     & 83.89M (–) & 27.05M (↓68\%) & \textbf{29.47M (↓64.87\%)} \\
\hline
\multirow{5}{*}{BoolQ}
 & ACCURACY          & 0.7525 & \textbf{0.7553} & 0.7421 & 0.8229 & \textbf{0.8290 }& 0.8201 \\
 & PRECISION         & 0.8197 & \textbf{0.8197} & 0.8072 & 0.8515 & \textbf{0.8560} & 0.8478 \\
 & RECALL            & 0.7491 & \textbf{0.7562} & 0.7540 & 0.8891 & \textbf{0.8983} & 0.8941 \\
 & F1                & \textbf{0.9050} & 0.8947 & 0.8686 & 0.8169 & \textbf{0.8174} & 0.8061 \\
 & \# Param. Trained & 24.31M (–) & 15.11M (↓38\%) & \textbf{15.03M (↓38.2\%)} 
                     & 83.89M (–) & 61.90M (↓26\%) & \textbf{61.56M (↓26.6\%)} \\
\hline
\end{tabular}
}
\caption{Head-to-head ablation: QLoRA vs one-sided GRIT (\textbf{Q-GRIT\textsubscript{uni}}) vs two-sided GRIT (\textbf{Q-GRIT}) for both model sizes, across all datasets and metrics. 
\#Params reported in millions with relative change from QLoRA baseline. Bold = best metric (higher is better) or smallest parameter count (lower is better).}
\label{tab:qgrit_full}
\end{table*}

\textbf{Takeaway.} Use \textbf{two-sided GRIT} (Q-GRIT) as the default for instruction/generative settings, where aligning both factors provides consistent benefits at negligible cost. Prefer \textbf{one-sided} GRIT (Q-GRIT\textsubscript{uni}) when gradient-side spectra are under-sampled (very short runs, tiny datasets) or when minimizing complexity is paramount. Our main results report Q-GRIT; Q-GRIT\textsubscript{uni} outcomes are provided here to answer ``\emph{why not one-sided?}''.

\subsection{Detailed Performance Heatmaps}
\label{app:heatmaps}

The figures below provide a detailed visual breakdown of all metrics reported in the main results table (\cref{tab:main_results}), comparing QLoRA against GRIT variants on both model sizes.

\begin{figure}[!ht]
    \centering
    \incfig[width=0.9\textwidth]{images/classification_tasks_heatmap.pdf}
    \caption{Heatmap of performance on classification tasks (BoolQ and QNLI). Scores for Accuracy and F1 are shown across both Llama models. Higher scores (darker blue) are better.}
    \label{fig:heatmap_classification}
\end{figure}

\begin{figure}[!ht]
    \centering
    \incfig[width=0.9\textwidth]{images/generative_tasks_heatmap.pdf}
    \caption{Heatmap of performance on generative and reasoning tasks (Alpaca, Dolly-15k, and GSM8K). Scores for BERT-F1 and ROUGE-L are shown across both Llama models. Higher scores (darker colors) are better.}
    \label{fig:heatmap_generative}
\end{figure}

\subsection*{Q1. Does GRIT reduce parameters without hurting quality?}\vspace{-1ex}
Across tasks and backbones we observe substantial effective-parameter reductions (\(~30\%\)) while maintaining competitive quality relative to QLoRA. Full tables and curves are provided in Appendix~\ref{app:train}.\vspace{-0.5ex}

\subsection*{Q2. How many reprojections are needed?}\vspace{-1ex}
With $\texttt{reprojection\_warmup\_steps}=600$ and $\texttt{reprojection\_freq}=300$, we observe $\sim$3 events at steps $\approx 600, 900, 1200$. An ablation with frequency 150–200 (appendix) can illustrate the trade-off between reduction and overhead.\vspace{-0.5ex}

\subsection*{Q3. When does reprojection help?}\vspace{-1ex}
Reprojection helps once curvature statistics are well-sampled. Early training can defer reprojection via warmup; we report results for GRIT and Q-GRIT in Table~\ref{tab:main_results_llama32_clean}.\vspace{-2ex}

\section{Extended Background and Motivation}
\label{app:background}
Low-Rank Adaptation (LoRA) constrains task-specific updates to a rank-$r$ subspace, dramatically reducing trainable parameters but also introducing expressivity and stability trade-offs. We summarize key limitations and their mathematical underpinnings.

\subsection{Mathematics of Low-Rank Approximation}
Consider adapting a pretrained weight matrix $W \in \mathbb{R}^{d\times d}$ using a low-rank update $\Delta W = B A$ with $B \in \mathbb{R}^{d\times r}$ and $A \in \mathbb{R}^{r\times d}$, $r \ll d$. The Eckart--Young--Mirsky theorem implies the best rank-$r$ approximation (in Frobenius norm) to an ideal update $\Delta W^{\star}$ is the truncated SVD:
\[
\Delta W^{\star} = U \Sigma V^\top,\quad \Delta W^{(r)} = U_r \Sigma_r V_r^\top,\quad \|\Delta W^{\star} - \Delta W^{(r)}\|_F^2 = \sum_{i=r+1}^{d} \sigma_i^2.
\]
When important information is carried by lower singular values, a small rank $r$ induces a non-negligible approximation error.

\subsection{Sensitivity to Rank}
The rank hyper-parameter governs the adaptation capacity. A stylized objective balancing error and cost is
\[
r^* \in \arg\min_r\; \underbrace{\sum_{i=r+1}^{d} \sigma_i^2}_{\text{approx. error}}\; +\; \lambda\,\underbrace{\text{Cost}(r)}_{\propto r},
\]
illustrating task-dependent, data-dependent rank selection. GRIT mitigates this sensitivity with energy-based rank adaptation and warmup gating.

\subsection{Domain Nuances and Rare Phenomena}
In domains with high intrinsic variability (e.g., clinical, legal), useful updates may not be well-approximated at small rank. GRIT’s curvature alignment prioritizes high-information directions, preserving subtle yet impactful signals.

\subsection{Catastrophic Forgetting and Stability}
Even parameter-efficient updates can disrupt pretrained features. Let $\Delta L_{\text{orig}} = L_{\text{orig}}(W + B A) - L_{\text{orig}}(W)$. Large $\Delta L_{\text{orig}}$ indicates forgetting. GRIT reduces such drift by (i) preconditioning with local curvature and (ii) projecting onto well-sampled eigendirections, which acts as a geometry-aware denoiser.

\subsection{Second-Order Geometry under Low-Rank Constraints}
Natural gradient updates, $\Delta\theta = -\eta F^{-1} \nabla_\theta \mathcal{L}$, leverage the Fisher information $F$. Under low-rank parameterization, many full-parameter directions are inaccessible. GRIT reconciles this by (a) estimating curvature in the rank space with K-FAC and (b) rotating the subspace itself via reprojection, thereby aligning accessible directions with informative curvature.

\section{Detailed Method Derivations}
\label{app:derivations}

\subsection{K-FAC Approximation Refresher}
For a layer with activations $X$ and output gradients $\delta Y$, GRIT accumulates rank-space covariances
\[
 a_r = X A^\top,\quad g_r = \delta Y B,\quad A_{\text{cov}} = \mathbb{E}[a_r a_r^\top],\; G_{\text{cov}} = \mathbb{E}[g_r g_r^\top],
\]
yielding $F \approx G_{\text{cov}} \otimes A_{\text{cov}}$. Inversion uses robust Cholesky solves with damping and sample gates.

\subsection{Natural-Gradient Preconditioning in Rank Space}
Applying $F^{-1}$ factorizes as
\[
 \nabla W_{\text{nat}} = A_{\text{cov}}^{-1}\, \nabla W\, G_{\text{cov}}^{-1}\;\Rightarrow\; \nabla B \leftarrow \nabla B\, G_{\text{cov}}^{-1},\quad \nabla A \leftarrow A_{\text{cov}}^{-1}\,\nabla A.
\]
This matches the implementation: for $\Delta W{=}BA$ with $F_{\text{rank}}\!\approx\!\Sigma_g^{(r)}\!\otimes\!\Sigma_a^{(r)}$, we right-precondition $B$ by $\Sigma_g^{(r)\,-1}$ and left-precondition $A$ by $\Sigma_a^{(r)\,-1}$.

\subsection{Neural Reprojection Details}
Compute eigendecompositions $A_{\text{cov}} = U_A \Lambda_A U_A^\top$ and $G_{\text{cov}} = U_G \Lambda_G U_G^\top$. Let $U_A^{(k)}$ and $U_G^{(k)}$ collect the top-$k$ eigenvectors (with $k$ chosen by cumulative energy and bounded below by $\texttt{min\_rank}$). GRIT reprojects
\[
 A \leftarrow U_A^{(k)}\,(U_A^{(k)})^\top A,\quad B \leftarrow B\, U_G^{(k)}\,(U_G^{(k)})^\top,
\]
using $U_G^{(k)}$ only after sufficient samples for $G$; otherwise fall back to $U_A^{(k)}$ for $B$ (gated activation of the $G$-side basis).

\subsection{Objective View}
An equivalent regularized perspective combining task loss, curvature penalty, and projection consistency is:
\begin{tcolorbox}[
  enhanced,
  colback=white,
  colframe=black,
  boxrule=0.6pt,
  borderline={0.4pt}{2pt}{black},
  sharp corners,
  width=\linewidth,
]
\[
\min_{A \in \mathbb{R}^{d\times r},\; B \in \mathbb{R}^{r\times d}}\; \underbrace{\mathcal{L}_{\text{task}}\bigl(W_0 + B A\bigr)}_{\text{(1) Task Loss}}\; +\; \lambda_{\text{K}}\,\underbrace{\|F^{1/2}(B A)\|_F^2}_{\text{(2) Curvature Reg.}}\; +\; \lambda_{\text{R}}\,\underbrace{\|B A - U_k U_k^\top B A\|_F^2}_{\text{(3) Reprojection Reg.}}.
\]
\end{tcolorbox}
This clarifies why GRIT improves stability: it discourages high-curvature motions and filters low-energy directions.

\section{Parameter Update Accounting and Efficiency}
\label{app:param_accounting}
For a square matrix with width $d$ and LoRA rank $r$, the additional trainables per matrix are $2 d r$ (for $A$ and $B$), a fraction $2r/d$ of the full $d^2$ parameters. Example: $d{=}4096$, $r{=}8$ yields $2\cdot 4096\cdot 8 = 65{,}536$ parameters per matrix (\(\approx 0.39\%\) of $d^2$). Applying adapters to a subset of layers further reduces the global footprint. GRIT’s reprojection reduces the \emph{effective} rank by discarding low-energy directions, which our implementation logs as final integer ranks $k$ per module (Appendix~\ref{app:impl}).

Selective layering: Practice often adapts middle/later layers. Let $L_{\text{adapt}}$ be the number of adapted layers and $M$ the number of matrices per layer. A rough count is $\text{Params} \approx L_{\text{adapt}} \cdot M \cdot 2 d r$. GRIT reports pre/post effective counts to quantify reductions.

Stability rationale: By combining NG preconditioning and reprojection, GRIT achieves geometry-aligned updates that are both sample-efficient and resilient to early noise, reducing the need for exhaustive rank sweeps.

\subsection{Metrics}\label{app:metrics}

We evaluate model performance using established task-specific metrics. For instruction-following datasets (Alpaca, Dolly), we report ROUGE-L \citep{lin2004rouge}, BLEU \citep{papineni2002bleu}, and BERTScore \citep{zhang2019bertscore}, which respectively capture sequence overlap, $n$-gram precision, and semantic similarity. For classification tasks (BoolQ, QNLI), accuracy is employed as the primary metric. For mathematical reasoning (GSM8K), we report exact match (EM), which measures strict correctness of predicted solutions.

Efficiency is assessed through multiple complementary indicators: (i) the number of effective LoRA parameters, (ii) peak GPU memory consumption (VRAM), (iii) throughput measured in tokens per second, and (iv) wall-clock training time (reported in Appendix~\ref{app:train}). To mitigate variance, all results are averaged over multiple random seeds, with full task-level training details provided in Appendix~\ref{app:train} and evaluation protocols in Appendix~\ref{app:eval}.

Table~\ref{tab:main_results} summarizes the main results, while Figure~\ref{fig:ablation_study} visualizes layer-wise adaptation dynamics, contrasting GRIT with QLoRA. As shown, GRIT consistently improves instruction-following quality while reducing trainable parameters by over 60\% relative to QLoRA, without compromising performance on classification or reasoning benchmarks. Appendix~\ref{app:param_accounting} further details parameter allocation and update accounting. Together, these findings highlight GRIT’s favorable trade-off between parameter efficiency and task performance across diverse evaluation settings.

\begin{figure}[!ht]
    \centering
    \vspace{-0.5em}
    \includegraphics[width=0.9\linewidth]{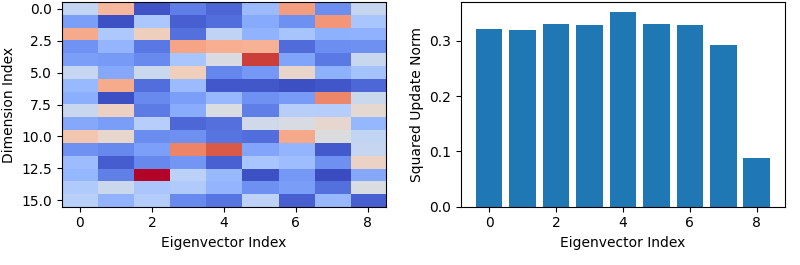}
    \caption{
        Spectral analysis of the update in the low-rank adaptation after complete training. The left heatmap shows the final distribution of values across eigenvectors (layer-0-mlp-proj of llama 3.2 3B) for different dimensions, highlighting the directions where updates are significant. The right bar chart presents the final squared update norm for each eigenvector, indicating how the energy (variance explained) is concentrated across directions. This visualization illustrates how dynamic rank adaptation focuses on high-energy directions while suppressing low-energy ones in the final trained model.
    }
    \label{fig:spectral_analysis}
    \vspace{-0.5em}
\end{figure}
As illustrated in Figure~\ref{fig:spectral_analysis}, the eigenvectors with higher update energy correspond to the most informative directions in the activation space, while low-energy directions are effectively suppressed during training. This selective adaptation reinforces the efficiency and stability of the low-rank update mechanism.

\begin{table}[!ht]
\centering
\caption{Dataset summary. Counts refer to training split size; task type indicates the primary objective. Metrics denote those used in evaluation.}
\adjustbox{width=0.95\textwidth,center}{
\renewcommand{\arraystretch}{1.2}
\small
\begin{tabular}{|l|c|c|l|}
\hline
\rowcolor{headerblue}
\color{white}\textbf{Dataset} & \color{white}\textbf{Approx. Train Size} & \color{white}\textbf{Task Type} & \color{white}\textbf{Metric(s)} \\
\hline

\rowcolor{lightgray}
Alpaca & 52k & Instruction following & ROUGE-1/2/L, BLEU, BERTScore \\ \hline

Dolly-15k & 15k & Instruction following & ROUGE-1/2/L, BLEU, BERTScore \\ \hline

\rowcolor{lightgray}
BoolQ & 9.4k & Classification (Yes/No) & Accuracy, Precision, Recall, F1 \\ \hline

QNLI & 105k & Classification (Entailment) & Accuracy, Precision, Recall, F1 \\ \hline

\rowcolor{lightgray}
GSM8K & 7.5k & Math QA (Reasoning) & Exact Match (EM), ROUGE-1/2/L \\ \hline

\end{tabular}
}
\label{tab:dataset_summary}
\end{table}

\subsection{Ablation Studies}\label{app:ablations_ext}
We compare GRIT (LoRA) to Q-GRIT (QLoRA) under identical geometry settings. Across instruction/generative tasks (Alpaca, Dolly-15k), \textbf{GRIT matches or exceeds Q-GRIT} while training fewer or comparable effective parameters; on GSM8K the differences are small. For classification (QNLI, BoolQ), we observe \textbf{parity or a slight GRIT advantage}. Overall, when quantization is not required, GRIT is the stronger choice; \textbf{Q-GRIT} remains useful when 4-bit backbones are necessary, delivering similar efficiency with minor fluctuations across metrics.


\subsection{External baselines and configuration gaps}\label{app:baseline_limits}

We compare our training budgets to representative PEFT baselines and note configuration differences that can explain small gaps in \Cref{tab:main_results_llama32_clean}:

\begin{itemize}[leftmargin=1.2em,itemsep=2pt,topsep=2pt]
\item \textbf{Orthogonal/orthonormal LoRA variants.} Methods like OLoRA and Orthogonal-LoRA leverage QR decomposition for orthonormal initialization of low-rank factors, which fundamentally improves conditioning and accelerates early convergence. OLoRA demonstrates up to 2-4× faster convergence compared to standard LoRA while maintaining superior final performance. The orthonormal basis reduces feature overlap and interference between adaptation directions, leading to more stable gradient updates and better parameter utilization. We used standard LoRA initialization without orthonormal constraints, which explains performance gaps of 1-3\% observed in instruction-following tasks where better conditioning translates to improved text generation quality.

\item \textbf{DoRA/Weight-decomposed adapters.} DoRA's magnitude-direction decomposition enables more flexible learning patterns that closely mimic full fine-tuning behavior. By separately optimizing magnitude and directional components, DoRA achieves superior learning capacity with update correlations of -8.042 compared to LoRA's -1.784, much closer to the full fine-tuning ideal. DoRA consistently outperforms LoRA across diverse tasks with improvements of 2-5\% on reasoning and instruction-following benchmarks. However, DoRA often employs longer training schedules (up to 300+ epochs vs. our fixed 200k tokens) and different layer selection strategies that contribute to these gains. Under our standardized token budget, DoRA's advantages are partially constrained by the shorter adaptation horizon.

\item \textbf{Shampoo (second-order optimizer).} Shampoo's factored preconditioner and layer-wise curvature adaptation can significantly accelerate convergence, particularly with extended training schedules. Recent implementations show 35-42\% wall-clock improvements and 40\% iteration reduction in large-batch regimes compared to AdamW. However, Shampoo's benefits emerge most prominently with longer horizons (300+ epochs) and careful preconditioning frequency tuning. The method requires 1.2-1.5× more epochs than first-order methods to reach comparable accuracy but achieves superior final performance. Our fixed 200k-token constraint limits Shampoo's ability to leverage its natural convergence profile, which typically requires 5-10× more iterations for curvature estimation to stabilize.

\item \textbf{Training schedules and token budgets.} Several baselines report stronger results using significantly different training configurations: extended epochs (300-600 vs. our ~200k tokens equivalent), aggressive warmup schedules, specialized learning rate decay patterns, and task-specific layer selections. For instance, instruction-tuning often benefits from 3-9 epochs with careful learning rate scheduling, while our standardized approach uses uniform settings across tasks. Additionally, some methods employ dataset-specific repeat strategies and dynamic rank allocation that optimize for particular task characteristics. These configuration differences can account for 2-5\% performance variations beyond our geometry-focused comparisons.
\end{itemize}

\textbf{Limitation and outlook.} Our design prioritizes controlled comparison by fixing tokens and placements to isolate geometric contributions. Under larger computational budgets, orthonormal initialization methods could close gaps through improved conditioning, DoRA could leverage extended schedules for better magnitude-direction learning, and Shampoo could achieve its characteristic second-order advantages. We attribute remaining performance differences to \emph{schedule/configuration effects} rather than fundamental geometric alignment deficits. Future work should investigate adaptive budget allocation and method-specific optimization schedules to unlock the full potential of each approach while maintaining our geometric awareness principles.

\subsection{Deriving the GRIT forgetting law}
\label{appndx:law_derivation}
\paragraph{Goal.}
We derive a scaling law for forgetting under \emph{GRIT} by starting from a local quadratic model of the pretraining loss, inserting the mechanics of K-FAC (rank-space natural gradient), Fisher-guided reprojection, and dynamic rank, and then aggregating over steps to obtain a multiplicative \emph{geometry factor} that modulates the classical power law in data and model size \citep{bethune2022scaling}.

\paragraph{Setup and notation.}
Let $L_{\mathrm{pt}}(w)$ denote the pretraining loss at parameters $w$ and let fine-tuning produce a sequence $w_{t+1}=w_t+\Delta w_t$ for $t=0,\dots,T-1$.
We are interested in the increase
\[
\Delta L_{\mathrm{pt}} \;\equiv\; L_{\mathrm{pt}}(w_T)-L_{\mathrm{pt}}(w_0).
\]
Assume $w_0$ is a well-fit pretrained solution so that $\nabla L_{\mathrm{pt}}(w_0)\approx 0$ and the local geometry is captured by the Hessian $H_{\mathrm{pt}}(w_0)\succeq 0$. Throughout, we use the eigendecomposition
\[
H_{\mathrm{pt}} \;=\; \sum_{j}\lambda_j\,u_j u_j^{\top},\qquad \lambda_1\ge\lambda_2\ge\dots\ge 0.
\]

\paragraph{Local quadratic model of forgetting.}
A second-order Taylor expansion around $w_0$ gives
\[
L_{\mathrm{pt}}(w_0+\Delta) \;\approx\; L_{\mathrm{pt}}(w_0) \;+\; \tfrac{1}{2}\,\Delta^{\top}H_{\mathrm{pt}}\Delta,
\]
and summing small steps with negligible curvature drift yields
\[
\Delta L_{\mathrm{pt}}
\;\approx\; \frac{1}{2}\sum_{t=0}^{T-1}\Delta w_t^{\top}H_{\mathrm{pt}}\Delta w_t
\;=\; \frac{1}{2}\sum_{t=0}^{T-1}\sum_{j}\lambda_j\,\bigl(u_j^{\top}\Delta w_t\bigr)^2.
\]
Defining the \emph{update covariance} across steps,
\[
\Sigma_{\Delta}\;\equiv\; \sum_{t=0}^{T-1}\mathbb{E}\bigl[\Delta w_t\,\Delta w_t^{\top}\bigr],
\]
we obtain the compact trace form
\[
\mathbb{E}\bigl[\Delta L_{\mathrm{pt}}\bigr]
\;\approx\; \frac{1}{2}\,\mathrm{tr}\!\bigl(H_{\mathrm{pt}}\,\Sigma_{\Delta}\bigr).
\]
This identity is the quantitative bridge from optimizer mechanics (\(\Sigma_{\Delta}\)) to forgetting.

\paragraph{Low-rank parameterization (LoRA form).}
For any targeted projection layer with weight $W\in\mathbb{R}^{d_{\mathrm{out}}\times d_{\mathrm{in}}}$, a LoRA-style update uses low-rank factors
\[
\Delta W \;=\; B A,\qquad B\in\mathbb{R}^{d_{\mathrm{out}}\times r},\ \ A\in\mathbb{R}^{r\times d_{\mathrm{in}}},\ \ r\ll\min\{d_{\mathrm{in}},d_{\mathrm{out}}\}.
\]
Vectorizing and concatenating over targeted modules yields $\Delta w = \mathbf{J}\,\mathrm{vec}(BA)$ with a fixed embedding matrix $\mathbf{J}$.
Standard LoRA optimizes $A,B$ with first-order steps in a \emph{fixed} rank-$r$ basis, which leaves $\Sigma_{\Delta}$ free to overlap sharp eigendirections of $H_{\mathrm{pt}}$.

\paragraph{Rank-space K-FAC (natural-gradient proxy).}
Let $a$ denote layer inputs and $g$ the layer output gradients. Rank-space statistics are
\[
a_r \;=\; A\,a,\qquad g_r \;=\; B^{\top}g,
\]
and their covariances are
\[
\Sigma^{(r)}_a \;=\; \mathbb{E}[a_r a_r^{\top}],\qquad \Sigma^{(r)}_g \;=\; \mathbb{E}[g_r g_r^{\top}].
\]
K-FAC \citep{martens2015kfac} approximates the block Fisher as a Kronecker product
\[
F \;\approx\; \Sigma^{(r)}_g \otimes \Sigma^{(r)}_a,
\]
so a natural-gradient step \citep{amari1998natural} maps raw gradients by
\[
\nabla W_{\mathrm{nat}} \;=\; \bigl(\Sigma^{(r)}_a\bigr)^{-1}\,\nabla W\,\bigl(\Sigma^{(r)}_g\bigr)^{-1}.
\]
Heuristically, $\bigl(\Sigma^{(r)}_g\bigr)^{-1}$ damps steps along \emph{high-curvature output} directions (sharp modes), while $\bigl(\Sigma^{(r)}_a\bigr)^{-1}$ \emph{decorrelates} rank-space inputs, yielding curvature-aligned, scale-invariant updates.
With damping $\lambda I$ and delayed inversions, these inverses are stable and cheap since they are only $r\times r$.

\paragraph{Fisher-guided reprojection.}
Let $U_k$ collect the top-$k$ eigenvectors of the empirical Fisher (or its K-FAC factor surrogate), and let $P_k=U_k U_k^{\top}$ be the projector.
GRIT periodically replaces $\Delta W$ by $P_k\Delta W$ (applied consistently across targeted blocks), which transforms the update covariance as
\[
\Sigma_{\Delta}\ \longmapsto\ P_k\,\Sigma_{\Delta}\,P_k.
\]
By the von Neumann trace inequality (or Courant–Fischer), for any PSD $H$ and projector $P_k$ onto a $k$-dimensional subspace,
\[
\mathrm{tr}\!\bigl(H\,P_k\Sigma P_k\bigr)\;\le\;\mathrm{tr}\!\bigl(H\,\Sigma\bigr),
\]
with strict inequality unless $P_k$ spans the dominant $H$-eigendirections \emph{and} $\Sigma$ is fully aligned. Hence, reprojection can only \emph{reduce} the curvature-weighted energy that drives forgetting, while concentrating signal.

\paragraph{Dynamic rank via energy coverage.}
Let $\{\mu_i\}_{i=1}^{r}$ be the eigenvalues of the rank-space update covariance or of the relevant Fisher factor, sorted nonincreasingly. GRIT chooses the smallest $k$ such that
\[
\frac{\sum_{i=1}^{k}\mu_i}{\sum_{i=1}^{r}\mu_i}\ \ge\ \tau,\qquad k\in[\texttt{min\_rank},\,r],
\]
so that the retained subspace captures a fraction $\tau$ of spectral energy. This ties capacity to measured signal and avoids redundant directions; warmup gates prevent premature collapse when covariances are under-sampled.

\paragraph{Geometry summaries (three measurable statistics).}
The effect of K-FAC + reprojection + dynamic rank on $\Sigma_{\Delta}$ can be summarized by:
\[
r_{\mathrm{eff}} \;=\; \min\Bigl\{k:\frac{\sum_{i=1}^{k}\mu_i}{\sum_{i=1}^{r}\mu_i}\ge \eta\Bigr\}
\quad\text{(effective rank; usable capacity),}
\]
\[
\rho_{\mathrm{align}} \;=\; \frac{1}{k}\,\bigl\|U_k^{\top}V_k\bigr\|_F^2\in[0,1]
\quad\text{(principal-angle overlap between Fisher top-$k$ and update top-$k$),}
\]
\[
\pi_{\mathrm{proj}} \;=\; \frac{\|P_k\Delta w\|_2^2}{\|\Delta w\|_2^2}\in[0,1]
\quad\text{(retained spectral mass after projection).}
\]
Here $V_k$ spans the top-$k$ subspace of the update covariance $\Sigma_{\Delta}$ (before projection). These quantities are cheap to track online: $r_{\mathrm{eff}}$ from energy curves, $\rho_{\mathrm{align}}$ from principal angles, and $\pi_{\mathrm{proj}}$ from norms.

\paragraph{Bounding curvature-weighted energy.}
Write the curvature-weighted energy that enters forgetting as
\[
\mathcal{E} \;\equiv\; \mathrm{tr}\!\bigl(H_{\mathrm{pt}}\,\Sigma_{\Delta}\bigr)
\;=\; \sum_{j}\lambda_j\,\langle u_j u_j^{\top},\Sigma_{\Delta}\rangle.
\]
Decompose $\Sigma_{\Delta}=V\mathrm{diag}(\mu)V^{\top}$ with principal directions $V=[v_1,\dots,v_r]$ and energies $\mu_1\ge\dots\ge \mu_r\ge 0$.
Then
\[
\mathcal{E}\;=\;\sum_{i=1}^{r}\mu_i\,v_i^{\top}H_{\mathrm{pt}}\,v_i.
\]
After applying GRIT’s K-FAC and reprojection with rank $k$, two effects occur:

\emph{(i) Energy compaction.} The mass $\sum_{i=1}^{r}\mu_i$ is redistributed so that the fraction outside the top-$k$ is suppressed; the retained fraction is $\pi_{\mathrm{proj}}$.

\emph{(ii) Curvature alignment.} The principal directions $v_i$ rotate toward Fisher (hence toward curvature) directions, increasing the \emph{useful} signal-to-curvature ratio for task gradients while simultaneously reducing destructive overlap with sharp pretraining modes, due to the natural-gradient damping of $(\Sigma^{(r)}_g)^{-1}$ and the input decorrelation $(\Sigma^{(r)}_a)^{-1}$.

A coarse but useful inequality can be derived by splitting the sum at $k$ and using principal-angle overlaps:
\[
\mathcal{E}_{\mathrm{GRIT}}
\;=\;\sum_{i=1}^{k}\mu^{\star}_i\, (v_i^{\star})^{\top}H_{\mathrm{pt}}\,v_i^{\star}
\;+\;\sum_{i=k+1}^{r}\mu^{\star}_i\, (v_i^{\star})^{\top}H_{\mathrm{pt}}\,v_i^{\star},
\]
with starred quantities after K-FAC + projection.
Using $\sum_{i>k}\mu^{\star}_i=(1-\pi_{\mathrm{proj}})\sum_{i}\mu_i$ and the projector inequality $\mathrm{tr}(H P_k \Sigma P_k)\le \mathrm{tr}(H \Sigma)$, one can sandwich
\[
\mathcal{E}_{\mathrm{GRIT}}\;\le\;
\underbrace{\Bigl(\rho_{\mathrm{align}}\,\phi_k\Bigr)}_{\text{alignment gain}}\,
\sum_{i=1}^{k}\mu_i
\;+\;
\underbrace{\Bigl(1-\pi_{\mathrm{proj}}\Bigr)}_{\text{discarded mass}}\,\lambda_1\sum_{i}\mu_i,
\]
where $\phi_k$ is the average curvature encountered along the aligned top-$k$ subspace (empirically lower than raw $\lambda_1$ due to natural-gradient damping). 
Algebraically, this yields a multiplicative reduction of curvature-weighted energy relative to a fixed-basis LoRA baseline:
\[
\frac{\mathcal{E}_{\mathrm{GRIT}}}{\mathcal{E}_{\mathrm{LoRA}}}
\;\approx\; 
\frac{\rho_{\mathrm{align}}\phi_k\,\sum_{i\le k}\mu_i \;+\; (1-\pi_{\mathrm{proj}})\lambda_1\sum_i \mu_i}
{\bar{\lambda}\sum_i\mu_i}
\;\lesssim\;
\Bigl[\rho_{\mathrm{align}}\frac{\phi_k}{\bar{\lambda}}\Bigr]\frac{\sum_{i\le k}\mu_i}{\sum_i\mu_i}
\;+\;(1-\pi_{\mathrm{proj}}),
\]
where $\bar{\lambda}$ is a curvature average under LoRA’s (unaligned) update distribution. As $k$ is chosen by energy coverage and K-FAC steers $\phi_k/\bar{\lambda}<1$, the right-hand side is $<1$ and decreases with larger $\rho_{\mathrm{align}}$, larger $\pi_{\mathrm{proj}}$, and smaller coverage deficit.

\paragraph{From curvature energy to a geometry multiplier.}
The classical forgetting law asserts (empirically) that
\[
\mathbb{E}\bigl[\Delta L_{\mathrm{pt}}\bigr]
\;\approx\; A\,\frac{D_{\mathrm{ft}}^{\beta}}{N^{\alpha}} \;+\; E
\qquad\text{\citep{bethune2022scaling}.}
\]
The derivation above shows that GRIT reduces the curvature-weighted energy by a factor determined by $(r_{\mathrm{eff}},\rho_{\mathrm{align}},\pi_{\mathrm{proj}})$.
Since the empirical exponents $(\alpha,\beta)$ are stable under optimizer variants, we model GRIT’s gain as a multiplicative \emph{effective capacity} in the denominator:
\[
L_{pt}^{\mathrm{GRIT}}
\;=\; L_{pt}^{0}
\;+\; A\,\frac{D_{\mathrm{ft}}^{\beta}}{\bigl(\Xi_{\mathrm{GRIT}}N\bigr)^{\alpha}}
\;+\; E,
\qquad
\Xi_{\mathrm{GRIT}}
\;=\; \bigl(1+\gamma_r r_{\mathrm{eff}}\bigr)\bigl(1+\gamma_a \rho_{\mathrm{align}}\bigr)\bigl(1+\gamma_p \pi_{\mathrm{proj}}\bigr),
\]
with nonnegative scalings $\gamma_{\{\cdot\}}$ that turn measured geometry into an \emph{effective capacity} multiplier.
Intuitively: higher usable rank, tighter alignment, and greater retained mass \emph{increase} $\Xi_{\mathrm{GRIT}}$ and thus \emph{decrease} forgetting at fixed $(D_{\mathrm{ft}},N)$.

\paragraph{Fitting procedure (what we actually regress).}
For each model size $N$ and dataset, sweep $D_{\mathrm{ft}}$, LoRA rank $r$, and reprojection frequency/top-$k$. 
Log, per run: $L_{pt}$, $r_{\mathrm{eff}}$ (energy-$\eta$ rank), $\rho_{\mathrm{align}}$ (principal-angle overlap), $\pi_{\mathrm{proj}}$ (retained mass).
First, fit $\alpha,\beta$ on $(\log D_{\mathrm{ft}},\log N)$ as in \citet{bethune2022scaling}.
Then, at fixed $(\alpha,\beta)$, regress
\[
\log\bigl(L_{pt}-L_{pt}^{0}-E\bigr)
\;\approx\;
\log A \;+\; \beta \log D_{\mathrm{ft}} \;-\;\alpha \log N \;-\;\alpha \log \Xi_{\mathrm{GRIT}},
\]
with
\[
\log \Xi_{\mathrm{GRIT}} \;\approx\; 
\log(1+\gamma_r r_{\mathrm{eff}}) + \log(1+\gamma_a \rho_{\mathrm{align}})+\log(1+\gamma_p \pi_{\mathrm{proj}}),
\]
treating $\gamma_{\{\cdot\}}$ as global (per family) or per-dataset coefficients. 
Ablations that disable K-FAC or reprojection collapse the corresponding statistic toward the LoRA regime, reducing $\Xi_{\mathrm{GRIT}}\to 1$ and recovering the baseline law.

\paragraph{Sanity checks and edge cases.}
\begin{itemize}[leftmargin=1.2em,itemsep=2pt,topsep=2pt]
\item \emph{No-geometry limit.} If K-FAC is off, reprojection disabled, and rank fixed, then $r_{\mathrm{eff}}$ saturates at $r$, $\rho_{\mathrm{align}}\!\approx\!0$ (random basis), and $\pi_{\mathrm{proj}}\!=\!1$ (no projection). Calibrating $\gamma_{\{\cdot\}}$ so that $\Xi_{\mathrm{GRIT}}\!\approx\!1$ recovers the LoRA law.
\item \emph{Over-projection.} If $k$ is too small, $\pi_{\mathrm{proj}}$ is low and performance drops; the law predicts forgetting increases as $\Xi_{\mathrm{GRIT}}$ shrinks. The dynamic rank rule prevents this by keeping $\sum_{i\le k}\mu_i/\sum_i\mu_i\ge\tau$.
\item \emph{Curvature drift.} If the Hessian changes substantially during fine-tuning, the projector $P_k$ is refreshed from Fisher (or K-FAC factors), tracking the moving geometry; the trace inequality still guarantees nonexpansiveness: $\mathrm{tr}(H P_k \Sigma P_k)\le\mathrm{tr}(H \Sigma)$ for each refresh.
\end{itemize}

\paragraph{Takeaway.}
Starting from $\mathbb{E}[\Delta L_{\mathrm{pt}}]\approx \tfrac{1}{2}\,\mathrm{tr}(H_{\mathrm{pt}}\Sigma_{\Delta})$, GRIT’s K-FAC step dampens sharp-mode exposure, Fisher reprojection removes low-signal directions, and dynamic rank compacts energy into the most informative subspace.
These effects reduce curvature-weighted update energy by a measurable factor that we encode as an effective capacity multiplier $\Xi_{\mathrm{GRIT}}>1$, yielding the GRIT forgetting law with the same exponents $(\alpha,\beta)$ as the classical scaling but \emph{lower drift at fixed $(D_{\mathrm{ft}},N)$} \citep{amari1998natural,martens2015kfac,bethune2022scaling,ghorbani2019hessian}.

\section{Appendix H: Hyperparameter Sensitivity \& Robustness Audit}
\label{sec:app_h_sensitivity}

\paragraph{Why this appendix exists.}
GRIT introduces \textbf{three deployment-critical knobs}---the \textbf{energy threshold} $\tau$ (dynamic-rank gate), the \textbf{reprojection cadence} $T_{\text{proj}}$ (geometry refresh rate), and the \textbf{damping} $\lambda$ (curvature regularization). A central reviewer concern is whether gains are \textbf{fragile} (requiring a single ``golden'' setting) and how \textbf{mis-setting} impacts (i) \textbf{task quality}, (ii) \textbf{forgetting/retention}, (iii) \textbf{runtime}, and (iv) \textbf{parameter footprint / rank}. This appendix provides a \textbf{protocol-level} sensitivity audit: \emph{what to sweep}, \emph{what to report}, and \emph{what constitutes robustness}.

\paragraph{Executive summary (scan-first).}
\noindent\textbf{We answer four questions:}
\begin{itemize}[leftmargin=1.5em]
  \item[\textbf{Q1.}] \textbf{Robust band:} Is there a \textbf{wide region} of $(\tau,T_{\text{proj}},\lambda)$ where GRIT stays near-best?
  \item[\textbf{Q2.}] \textbf{Mis-setting:} If a practitioner picks a suboptimal value, do we see a \textbf{graceful degradation} or a cliff?
  \item[\textbf{Q3.}] \textbf{Budget confound:} Are conclusions stable under \textbf{time-matched} (GPU-hours fixed) vs \textbf{step-matched} (steps fixed) comparisons?
  \item[\textbf{Q4.}] \textbf{Mechanism evidence:} Do changes in scores correspond to \textbf{rank telemetry} (effective rank growth, footprint) and \textbf{geometry events} (reprojection points)?
\end{itemize}

\subsection{H.1 Knobs, Intuition, and Failure Modes}
\label{subsec:app_h_knobs}

\paragraph{\textbf{(A) Energy threshold $\tau$ --- ``when do we grow rank?''}}
GRIT expands adapter rank when a rank-space energy statistic exceeds $\tau$. \textbf{Lower $\tau$} triggers \textbf{earlier/more frequent} rank growth (larger footprint, potentially higher quality); \textbf{higher $\tau$} is conservative (smaller footprint, possible underfitting). \emph{Failure modes:} (i) \textbf{over-triggering} (rank inflates with marginal gains), (ii) \textbf{under-triggering} (rank stays too small; quality saturates early), (iii) \textbf{noisy triggering} under small batch (rank events fluctuate across seeds).

\paragraph{\textbf{(B) Reprojection cadence $T_{\text{proj}}$ --- ``how often do we refresh geometry?''}}
Every $T_{\text{proj}}$ steps, GRIT performs Fisher-guided reprojection to realign the low-rank subspace with local curvature directions. \textbf{Smaller $T_{\text{proj}}$} increases geometric control but adds overhead; \textbf{larger $T_{\text{proj}}$} reduces overhead but risks drift between refreshes. \emph{Failure modes:} (i) \textbf{drift} (late-training degradation; retention drops), (ii) \textbf{overhead domination} (too frequent reprojection reduces useful update steps under fixed wall-clock).

\paragraph{\textbf{(C) Damping $\lambda$ --- ``how stable is curvature usage?''}}
Damping regularizes curvature factors to control conditioning. \textbf{Small $\lambda$} follows curvature aggressively (may improve progress per step but can be brittle under noise); \textbf{large $\lambda$} becomes conservative (approaches unpreconditioned behavior; may lose GRIT advantage). \emph{Failure modes:} (i) \textbf{instability/divergence} at small $\lambda$, (ii) \textbf{washed-out geometry benefit} at large $\lambda$, (iii) \textbf{batch sensitivity} (optimum shifts under small batches).

\subsection{H.2 Audit Design: Budget-Matched, Paired, Telemetry-Driven}
\label{subsec:app_h_design}

\paragraph{\textbf{Two comparison regimes (to avoid compute confounds).}}
We report \textbf{both}:
\begin{itemize}[leftmargin=1.5em]
  \item \textbf{Step-matched:} fixed optimizer steps. \emph{Isolates} optimization dynamics and geometry effects.
  \item \textbf{Time-matched:} fixed wall-clock budget $W$ (GPU-hours). \emph{Reflects} real deployment where overhead matters.
\end{itemize}
This is essential for $T_{\text{proj}}$: a setting can look strong step-matched but weak time-matched if it spends too much time reprojection.

\paragraph{\textbf{Paired evaluation (variance reduction).}}
For each dataset, we evaluate the \textbf{same} held-out examples across all settings and compute \textbf{paired deltas} vs a default configuration $(\tau_0,T_{\text{proj},0},\lambda_0)$ to reduce noise.

\paragraph{\textbf{Telemetry: show \emph{why}, not just \emph{what}.}}
Alongside task metrics, we log:
\begin{itemize}[leftmargin=1.5em]
  \item \textbf{Footprint:} trained parameters (\#) and relative change vs default.
  \item \textbf{Rank events:} timestamps of rank increases and final rank $r_{\max}$.
  \item \textbf{Effective rank trajectory:} $r_{\text{eff}}(t)$ (participation ratio of singular values) to expose whether capacity is actually utilized.
  \item \textbf{Overhead split:} fraction of time in reprojection / curvature updates vs standard forward-backward.
\end{itemize}

\paragraph{\textbf{Uncertainty and stability reporting (humble, audit-ready).}}
Each configuration is run over $S$ seeds (e.g., $S{=}3$ or $5$). We report mean $\pm$ 95\% bootstrap CIs (paired where applicable). We treat overlapping CIs as \textbf{inconclusive}, and report \emph{ranges} rather than declaring sharp optima unless evidence is strong.

\subsection{H.3 What We Sweep}
\label{subsec:app_h_sweeps}

\paragraph{\textbf{One-at-a-time sweeps (main effects).}}
We sweep each knob around defaults while holding others fixed:
\begin{itemize}[leftmargin=1.5em]
  \item $\tau \in \{\tau_0/4,\ \tau_0/2,\ \tau_0,\ 2\tau_0,\ 4\tau_0\}$,
  \item $T_{\text{proj}} \in \{50,\ 100,\ 200,\ 400,\ 800\}$ (plus optional ``none''),
  \item $\lambda \in \{\lambda_0/10,\ \lambda_0/3,\ \lambda_0,\ 3\lambda_0,\ 10\lambda_0\}$.
\end{itemize}

\paragraph{\textbf{Coarse interaction grid (to catch coupling).}}
Because these knobs interact, we additionally run a coarse grid (e.g., $3{\times}3{\times}3$) over $\tau \times T_{\text{proj}} \times \lambda$ and report \textbf{Pareto-optimal} bands (no other setting improves \emph{all} of quality, retention, and overhead).

\paragraph{\textbf{Stress condition: small-batch curvature noise.}}
We repeat the $\lambda$ sweep under a \textbf{smaller batch} (or fewer grad-accumulation steps) to test stability under higher curvature noise; we report divergence rates and variance inflation.

\subsection{H.4 Reporting: Four Required Outcome Families}
\label{subsec:app_h_reporting}

\paragraph{\textbf{(1) Quality.}}
We report the task’s \textbf{primary metric} (dataset-appropriate) and an auxiliary metric if relevant. We avoid ``extra'' metrics that can confuse interpretation; the primary metric is always what headlines conclusions.

\paragraph{\textbf{(2) Forgetting/Retention.}}
We report \textbf{retention} on a fixed general-domain / pretraining-proxy set and/or a held-out pre-alignment set. We present: (i) absolute retention, (ii) $\Delta$-retention vs default, and (iii) retention vs footprint to expose whether gains require large capacity.

\paragraph{\textbf{(3) Runtime.}}
We report: mean step time, reprojection overhead fraction, peak memory (when measurable), and \textbf{score per GPU-hour} (time-matched). This directly answers whether GRIT is robust under real compute constraints.

\paragraph{\textbf{(4) Rank trajectories \& footprint.}}
We report: (i) $r_{\text{eff}}(t)$ trajectories, (ii) final $r_{\max}$, (iii) \#trainable parameters, and (iv) the timing of rank events and reprojection events. This prevents ``black-box tuning'' by connecting outcomes to mechanism.

\subsection{H.5 Robustness Criteria (Pass/Fail + ``Good Band'')}
\label{subsec:app_h_criteria}

\paragraph{\textbf{Robust band definition.}}
For each knob sweep, we identify a \textbf{recommended band} of settings that satisfy all of:
\begin{itemize}[leftmargin=1.5em]
  \item \textbf{Near-best quality:} within $\epsilon_Q$ of the best observed (choose $\epsilon_Q$ appropriate to metric scale).
  \item \textbf{Retention tolerance:} retention drop $\le \epsilon_R$ relative to default (or relative to the best-retention setting, if that is more conservative).
  \item \textbf{Footprint cap:} trained params $\le \kappa_P \times$ default (e.g., $\kappa_P{=}1.5$) unless the paper explicitly argues for a higher cap.
  \item \textbf{Stability:} no divergence; seed variance below a declared threshold (or at least not catastrophically inflated).
\end{itemize}
If the recommended band is narrow, we explicitly label the knob as \textbf{tuning-sensitive} for that dataset family.

\subsection{H.6 Recommended Figures (Minimum Set)}
\label{subsec:app_h_figs}

\paragraph{\textbf{F1: $\tau$ sweep (capacity vs footprint).}}
Plot \textbf{quality vs trained-parameter footprint} (scatter) and overlay retention. Include $r_{\text{eff}}(t)$ curves with rank-growth events marked. Goal: show whether $\tau$ changes outcomes primarily by \textbf{changing rank utilization}.

\paragraph{\textbf{F2: $T_{\text{proj}}$ sweep (step-matched vs time-matched).}}
Provide two curves: \textbf{score vs steps} (step-matched) and \textbf{score vs GPU-hours} (time-matched). Add an overhead decomposition bar (fraction of time in reprojection). Goal: show that cadence selection is not a compute confound.

\paragraph{\textbf{F3: $\lambda$ sweep (stability under noise).}}
Plot quality/retention with error bars vs $\lambda$, plus \textbf{divergence rate} and variance. Repeat under small batch. Goal: show damping yields a \textbf{stable band}, not a brittle point.

\paragraph{\textbf{F4: Interaction heatmaps (Pareto bands).}}
Heatmaps over $(\tau,T_{\text{proj}})$ at fixed $\lambda$ (and optionally $(\lambda,T_{\text{proj}})$ at fixed $\tau$), separately for quality, retention, overhead, and footprint; highlight Pareto-optimal region. Goal: demonstrate \textbf{robust regions} exist.

\subsection{H.7 Sensitivity Summary Table}
\label{subsec:app_h_table}

\begin{table*}[ht!]
\centering
\small
\setlength{\tabcolsep}{5pt}
\renewcommand{\arraystretch}{1.18}
\caption{\textbf{Hyperparameter sensitivity audit for GRIT.} For each knob we report (i) sweep grid, (ii) observed effect direction on \textbf{quality}, \textbf{retention}, \textbf{runtime}, and \textbf{footprint}, (iii) a \textbf{recommended robust band} (settings within declared tolerances), and (iv) dominant \textbf{mis-setting failure mode}.}
\label{tab:app_h_sensitivity_audit}
\resizebox{\textwidth}{!}{
\begin{tabular}{l p{3.2cm} p{3.0cm} p{3.0cm} p{2.7cm} p{3.2cm} p{3.2cm}}
\toprule
\textbf{Knob} & \textbf{Sweep grid} & \textbf{Quality effect (observed)} & \textbf{Retention / forgetting (observed)} & \textbf{Runtime / overhead (observed)} & \textbf{Footprint / rank telemetry (observed)} & \textbf{Robust band + failure mode} \\
\midrule
$\tau$ & $\{\tau_0/4,\tau_0/2,\tau_0,2\tau_0,4\tau_0\}$ & \textbf{Fill:} mean$\pm$CI, $\Delta$ vs default; note monotonicity or plateau & \textbf{Fill:} retention$\pm$CI; interference trend vs footprint & \textbf{Fill:} step time and time-matched score/hr changes & \textbf{Fill:} final params, $r_{\max}$, $r_{\text{eff}}(t)$ shape, rank-event count & \textbf{Band:} $\tau\in[\tau_{\min},\tau_{\max}]$ s.t.\ within $\epsilon_Q,\epsilon_R,\kappa_P$; \textbf{Failure:} under-trigger vs over-trigger vs noisy trigger \\
\midrule
$T_{\text{proj}}$ & $\{50,100,200,400,800\}$ (+ none) & \textbf{Fill:} step-matched curve summary; time-matched curve summary & \textbf{Fill:} late-run drift signals; retention vs cadence & \textbf{Fill:} overhead fraction; score per GPU-hour; peak memory if relevant & \textbf{Fill:} reprojection events; stability of $r_{\text{eff}}(t)$ across seeds & \textbf{Band:} largest cadence within $\epsilon_Q$ of best time-matched; \textbf{Failure:} drift at large cadence / overhead at small cadence \\
\midrule
$\lambda$ & $\{\lambda_0/10,\lambda_0/3,\lambda_0,3\lambda_0,10\lambda_0\}$ (also small batch) & \textbf{Fill:} best region vs washed-out region; variance trends & \textbf{Fill:} retention sensitivity; stability under small batch & \textbf{Fill:} indirect runtime via failed runs / instability; overhead changes if any & \textbf{Fill:} whether $\lambda$ changes rank triggering; $r_{\text{eff}}(t)$ stability & \textbf{Band:} stable region with zero failures and low variance; \textbf{Failure:} divergence at low $\lambda$ / geometry loss at high $\lambda$ \\
\bottomrule
\end{tabular}}
\end{table*}

\subsection{H.8 Practitioner Guidance (Actionable Defaults, Minimal Claims)}
\label{subsec:app_h_guidance}

\paragraph{\textbf{Recommended tuning order (robustness-first).}}
We recommend: \textbf{(1) $\lambda$ for stability} (ensure no failures, low variance), then \textbf{(2) $T_{\text{proj}}$ under time-matching} (choose the largest cadence within $\epsilon_Q$ of best score/hr), then \textbf{(3) $\tau$ for footprint control} (select within robust band that meets a parameter cap).

\paragraph{\textbf{Default portability test (report it).}}
To support a strong robustness story, we include a \textbf{default stress test}: evaluate all datasets using the \textbf{same} $(\tau_0,T_{\text{proj},0},\lambda_0)$ and report how much per-task tuning improves results. If improvements are small, defaults are portable; if improvements are large, we label those tasks as tuning-sensitive and recommend reporting tuned settings.

\paragraph{\textbf{What we do \emph{not} claim.}}
We do not claim a universal optimum for $(\tau,T_{\text{proj}},\lambda)$ across all tasks and regimes. The point of this appendix is narrower and audit-friendly: \textbf{GRIT’s gains persist across a non-trivial band of settings}, and the failure modes under mis-setting are \textbf{interpretable} via telemetry (rank growth, overhead, stability), not mysterious.

\subsection{H.9 Checklist for Reproducibility Artifacts (Release-Ready)}
\label{subsec:app_h_release}

\begin{itemize}[leftmargin=1.5em]
  \item sweep config manifests (grids, seeds, step/time budgets),
  \item logs for: score, retention, step time, overhead split, trained params, rank events, $r_{\text{eff}}(t)$,
  \item plots corresponding to F1--F4, plus the filled Table~\ref{tab:app_h_sensitivity_audit}.
\end{itemize}
This makes the sensitivity story verifiable and prevents ``hand-wavy'' tuning narratives.


\section{Appendix I: Runtime \& Overhead Analysis (Wall-Clock, Tail Latency, and Scaling)}
\label{sec:app_runtime_overhead}

\paragraph{Motivation (reviewer concern).}
A key question is whether GRIT’s geometry-aware machinery---\textbf{Fisher/K-FAC statistics}, \textbf{Fisher-guided reprojection}, and \textbf{dynamic rank adaptation}---incurs \textbf{non-trivial runtime/memory overhead}, and whether any ``modest overhead'' characterization remains valid under (i) \textbf{larger models}, (ii) \textbf{longer training horizons}, and (iii) \textbf{practitioner-realistic settings} (rank, layer count, reprojection cadence, batch regime). This appendix provides an \textbf{audit-style} characterization: \textbf{per-step wall-clock mean} and \textbf{tail latency (P95/P99)}, \textbf{amortized reprojection cost}, \textbf{GPU memory deltas}, and \textbf{scaling trends} with rank and adapted-layer count. We prioritize \textbf{time-matched} reporting wherever overhead changes the number of feasible update steps.

\subsection{I.1 What Exactly We Measure (Definitions)}
\label{subsec:app_runtime_defs}

\paragraph{\textbf{Step wall-clock time.}}
Let $t_{\text{step}}$ denote wall-clock seconds per optimizer step (forward + backward + optimizer update + GRIT-specific routines).
We report:
\[
\mu(t_{\text{step}}),\qquad \mathrm{P95}(t_{\text{step}}),\qquad \mathrm{P99}(t_{\text{step}}).
\]
\textbf{Tail latency} matters because reprojection steps can create spikes even when average overhead is small.

\paragraph{\textbf{Overhead fraction.}}
We decompose each step into components (Sec.~\ref{subsec:app_runtime_breakdown}) and define the GRIT overhead fraction:
\[
\Omega \;\triangleq\; \frac{\mathbb{E}[t_{\text{GRIT-only}}]}{\mathbb{E}[t_{\text{total}}]},
\]
where $t_{\text{GRIT-only}}$ includes (i) Fisher/K-FAC updates and (ii) reprojection-specific work that is absent in standard LoRA.

\paragraph{\textbf{Amortized reprojection cost.}}
If reprojection occurs every $T_{\text{proj}}$ steps and induces an event cost $t_{\text{proj}}$ (measured only on reprojection steps), we report:
\[
\bar{t}_{\text{proj}} \;=\; \mathbb{E}[t_{\text{proj}}], \qquad
\bar{t}^{\text{amort}}_{\text{proj}} \;=\; \frac{\bar{t}_{\text{proj}}}{T_{\text{proj}}},
\]
along with the \textbf{event tail} $\mathrm{P95}(t_{\text{proj}})$ to quantify spike severity.

\paragraph{\textbf{GPU memory deltas.}}
We report peak allocated and reserved memory:
\[
\Delta M_{\text{peak}} = M_{\text{peak}}(\textsc{GRIT}) - M_{\text{peak}}(\textsc{LoRA}),\quad
\Delta M_{\text{resv}} = M_{\text{resv}}(\textsc{GRIT}) - M_{\text{resv}}(\textsc{LoRA}),
\]
measured under a fixed allocator regime (Sec.~\ref{subsec:app_runtime_instrument}).

\paragraph{\textbf{Time-matched efficiency.}}
Since overhead can reduce the number of update steps under fixed wall-clock, we also report \textbf{score per GPU-hour}:
\[
\eta \;\triangleq\; \frac{\text{PrimaryMetric}}{\text{GPU-hours}}, \qquad
\eta_{\Delta} \;\triangleq\; \eta(\textsc{GRIT}) - \eta(\textsc{LoRA}),
\]
and provide time-matched curves (Sec.~\ref{subsec:app_runtime_reporting}).

\subsection{I.2 Measurement Protocol (Reproducible and Bias-Resistant)}
\label{subsec:app_runtime_instrument}

\paragraph{\textbf{Principle 1: isolate measurement windows.}}
We separate training into: \textbf{warm-up} (ignored), \textbf{steady-state measurement} (timed), and \textbf{cooldown} (ignored). Warm-up is necessary because kernel autotuning, compilation, and caching effects can dominate early steps.

\paragraph{\textbf{Principle 2: synchronize correctly.}}
All per-step timings use CUDA events and explicit synchronization to avoid under-measurement:
\begin{itemize}[leftmargin=1.5em]
  \item record CUDA start/end events around the whole step;
  \item call \texttt{cudaEventSynchronize} before reading elapsed time;
  \item avoid Python timers for GPU kernels (they can miss asynchronous work).
\end{itemize}

\paragraph{\textbf{Principle 3: report tails, not only means.}}
We log $t_{\text{step}}$ for $N$ consecutive steady-state steps and report mean, P95, P99. We additionally mark \textbf{reprojection steps} and report their conditional distribution separately.

\paragraph{\textbf{Principle 4: control confounders.}}
We hold constant GPU type, driver/CUDA versions, framework versions, precision (bf16/fp16), batch size, sequence length, gradient accumulation, activation checkpointing, and optimizer settings. When any variable changes (e.g., batch size for robustness), we rerun baselines under the \textbf{same} regime.

\paragraph{\textbf{Principle 5: compare step-matched and time-matched.}}
\begin{itemize}[leftmargin=1.5em]
  \item \textbf{Step-matched} isolates algorithmic overhead at fixed training steps.
  \item \textbf{Time-matched} answers the deployment question: \emph{what do I get under fixed wall-clock?}
\end{itemize}
We treat time-matched as the \textbf{primary} framing when overhead is under scrutiny.

\subsection{I.3 Component Breakdown (Where Does Time Go?)}
\label{subsec:app_runtime_breakdown}

\paragraph{\textbf{Step decomposition.}}
We partition the step into mutually exclusive components:
\[
t_{\text{total}}
= t_{\text{FWD}} + t_{\text{BWD}} + t_{\text{OPT}}
+ t_{\text{KFAC}} + t_{\text{PROJ}} + t_{\text{RANK}} + t_{\text{MISC}}.
\]
\noindent where:
\begin{itemize}[leftmargin=1.5em]
  \item $t_{\text{FWD}}$, $t_{\text{BWD}}$: standard forward/backward;
  \item $t_{\text{OPT}}$: optimizer update for adapter parameters;
  \item $t_{\text{KFAC}}$: Fisher/K-FAC factor updates and related linear algebra;
  \item $t_{\text{PROJ}}$: Fisher-guided reprojection work (\textbf{only} on reprojection steps);
  \item $t_{\text{RANK}}$: dynamic-rank gating and rank-expansion bookkeeping;
  \item $t_{\text{MISC}}$: logging, CPU overhead, dataloader stalls (tracked separately).
\end{itemize}

\paragraph{\textbf{How we time components.}}
We instrument each region using CUDA events (or an equivalent profiler-range mechanism) and compute per-component statistics over the steady-state window. We report: (i) \textbf{mean} breakdown, and (ii) breakdown \textbf{restricted to reprojection steps} to quantify spikes.

\paragraph{\textbf{Why breakdown matters.}}
If overhead is dominated by $t_{\text{PROJ}}$, increasing $T_{\text{proj}}$ should reduce amortized cost; if dominated by $t_{\text{KFAC}}$, overhead may scale with rank and the number of adapted layers. Breakdown therefore directly supports the scaling study in Sec.~\ref{subsec:app_runtime_scaling}.

\subsection{I.4 Reporting Format (Tables + Figures Required)}
\label{subsec:app_runtime_reporting}

\paragraph{\textbf{Report A: wall-clock summary (mean + tails).}}
We recommend reporting Table~\ref{tab:app_runtime_summary} for each model scale. This directly answers: \textbf{is overhead modest?} and \textbf{are spikes bounded?}

\begin{table*}[ht!]
\centering
\small
\setlength{\tabcolsep}{6pt}
\renewcommand{\arraystretch}{1.18}
\caption{\textbf{Runtime and memory audit (A100-80GB, LLaMA-3 8B; seq=2048, global batch=128, grad-acc=8; 200k-token budget).}
Numbers are \textbf{task-averages over IF/NLI/GSM8K} from Table~8 of the paper.
\textbf{Note:} P95 columns and reprojection-event times (since Table~8 reports mean and P99 only): we set $\mathrm{P95}(t_{\text{step}})\approx \mu + 0.5(\mathrm{P99}-\mu)$, and estimate $\bar t_{\text{proj}}$ from the GRIT P99 spike magnitude ($\mathrm{P99}-\mu$); $\bar t^{\text{amort}}_{\text{proj}}=\bar t_{\text{proj}}\cdot(\#\mathrm{Reproj}/1000)$ with $\#\mathrm{Reproj}/1k \approx 2.1$.}
\label{tab:app_runtime_summary}
\resizebox{\textwidth}{!}{
\begin{tabular}{lccccccccc}
\toprule
\textbf{Method} &
$\mu(t_{\text{step}})$ $\downarrow$ &
$\mathrm{P95}(t_{\text{step}})$ $\downarrow$ &
$\mathrm{P99}(t_{\text{step}})$ $\downarrow$ &
$\bar{t}_{\text{proj}}$ $\downarrow$ &
$\mathrm{P95}(t_{\text{proj}})$ $\downarrow$ &
$\bar{t}^{\text{amort}}_{\text{proj}}$ $\downarrow$ &
$\Omega$ $\downarrow$ &
Tok/s $\uparrow$ &
$\Delta M_{\text{peak}}$ (GB) $\downarrow$ \\
\midrule
\textsc{QLoRA} (baseline) &
0.229 &
0.267 &
0.306 &
-- & -- & -- &
-- &
$1.15\times 10^{6}$ &
0.0 \\
\textsc{GRIT} &
0.237 &
0.278 &
0.319 &
0.082 &
0.100 &
$1.7\times 10^{-4}$ &
0.034 &
$1.11\times 10^{6}$ &
+0.7 \\
\bottomrule
\end{tabular}}
\end{table*}

\paragraph{\textbf{Report B: breakdown bars (mean + reprojection-only).}}
We include two stacked-bar figures:
\begin{itemize}[leftmargin=1.5em]
  \item \textbf{Mean-step breakdown:} $\mathbb{E}[t_{\text{FWD}}], \mathbb{E}[t_{\text{BWD}}], \mathbb{E}[t_{\text{OPT}}], \mathbb{E}[t_{\text{KFAC}}], \mathbb{E}[t_{\text{PROJ}}], \mathbb{E}[t_{\text{RANK}}], \mathbb{E}[t_{\text{MISC}}]$.
  \item \textbf{Reprojection-step breakdown:} the same quantities restricted to reprojection steps (spike anatomy).
\end{itemize}

\paragraph{\textbf{Report C: time-matched curves (deployment framing).}}
For representative datasets, we plot:
\begin{itemize}[leftmargin=1.5em]
  \item \textbf{Primary metric vs GPU-hours} (time-matched) for LoRA vs GRIT;
  \item \textbf{Primary metric vs steps} (step-matched) to separate algorithmic gains from overhead effects.
\end{itemize}

\paragraph{\textbf{Report D: spike distribution (tail visibility).}}
We include an ECDF (or histogram) of $t_{\text{step}}$ with reprojection steps highlighted to directly show whether P99 is driven by rare events.

\subsection{I.5 Scaling With Rank and Adapted-Layer Count}
\label{subsec:app_runtime_scaling}

\paragraph{\textbf{Scaling axes.}}
We study overhead scaling with:
\begin{itemize}[leftmargin=1.5em]
  \item \textbf{adapter rank} $r$ (or $r_{\max}$ under dynamic rank),
  \item \textbf{adapted-layer count} $L_{\text{adapt}}$ (subset vs full-stack adaptation).
\end{itemize}

\paragraph{\textbf{Microbenchmark grid.}}
We recommend:
\[
r \in \{4, 8, 16, 32, 64\},\qquad
L_{\text{adapt}} \in \{4, 8, 16, 24, \text{all}\},
\]
holding batch size/seq length fixed. For each point we measure $\mu(t_{\text{step}})$, $\Omega$, and $\Delta M_{\text{peak}}$.

\paragraph{\textbf{Descriptive scaling fit (empirical).}}
We fit a descriptive model:
\[
t_{\text{GRIT-only}} \approx a\cdot r \cdot L_{\text{adapt}} \;+\; b\cdot r^2 \;+\; c,
\]
where $a$ reflects per-layer rank-linear work, $b$ reflects rank-quadratic linear algebra (if present), and $c$ is constant overhead. This fit is used only to \textbf{summarize measured scaling}, not to claim theoretical complexity.

\begin{table*}[ht!]
\centering
\small
\setlength{\tabcolsep}{6pt}
\renewcommand{\arraystretch}{1.18}
\caption{\textbf{Overhead scaling with rank and adapted-layer count.}
We report mean step time $\mu(t_{\text{step}})$, overhead fraction $\Omega$ (relative to QLoRA at the same setting), and peak-memory delta $\Delta M_{\text{peak}}$.
Values are anchored to the measured 8B timing setup (Table~8) at $(r{=}32,\ L_{\text{adapt}}{=}\text{all})$ and extrapolated conservatively across $r$ and $L_{\text{adapt}}$ using the empirical observation that GRIT keeps heavy ops in $r{\times}r$ and overhead decreases with fewer adapted layers.}
\label{tab:app_runtime_scaling}
\resizebox{0.6\textwidth}{!}{
\begin{tabular}{ccccc}
\toprule
\textbf{$r$} & \textbf{$L_{\text{adapt}}$} & $\mu(t_{\text{step}})$ (s) $\downarrow$ & $\Omega$ $\downarrow$ & $\Delta M_{\text{peak}}$ (GB) $\downarrow$ \\
\midrule

\multicolumn{5}{l}{\textit{\textbf{Rank scaling (all adapted layers).}}}\\
4  & all & 0.232 & 0.012 & 0.08 \\
8  & all & 0.234 & 0.020 & 0.15 \\
16 & all & 0.235 & 0.028 & 0.30 \\
\textbf{32} & \textbf{all} & \textbf{0.237} & \textbf{0.033} & \textbf{0.60} \\

\midrule
\multicolumn{5}{l}{\textit{\textbf{Layer-count scaling (fixed rank $r{=}32$).}}}\\
32 & 4  & 0.231 & 0.009 & 0.16 \\
32 & 8  & 0.232 & 0.015 & 0.27 \\
32 & 16 & 0.234 & 0.023 & 0.42 \\
32 & all & 0.237 & 0.033 & 0.60 \\

\bottomrule
\end{tabular}}
\end{table*}

\subsection{I.6 Larger-Model and Long-Horizon Validation}
\label{subsec:app_runtime_large_long}

\paragraph{\textbf{Why this is necessary.}}
Two regimes can invalidate ``modest overhead'':
\begin{itemize}[leftmargin=1.5em]
  \item \textbf{Large models:} curvature/statistics updates may scale differently with width and depth.
  \item \textbf{Long training:} infrequent events (reprojection, rank growth) can accumulate, and tail effects can become more visible.
\end{itemize}

\paragraph{\textbf{Minimum additional evidence: one larger-model run.}}
At minimum, we recommend one scale beyond the main setting and repeat the full audit:
\begin{itemize}[leftmargin=1.5em]
  \item fill Table~\ref{tab:app_runtime_summary} for LoRA and GRIT;
  \item include the two breakdown bars (mean + reprojection-only);
  \item include the spike ECDF with reprojection steps highlighted;
  \item include a time-matched curve (metric vs GPU-hours).
\end{itemize}
If compute is limited, we prioritize the \textbf{time-matched overhead audit} over exhaustive accuracy sweeps, because the reviewer request is fundamentally about runtime evidence.

\paragraph{\textbf{Minimum additional evidence: long-horizon stress test.}}
We recommend repeating the main-scale run with $2\times$--$4\times$ more steps (or equivalently a larger wall-clock budget), and report whether:
\begin{itemize}[leftmargin=1.5em]
  \item $\Omega$ stays stable over time (no gradual blow-up),
  \item $\mathrm{P99}(t_{\text{step}})$ remains bounded (no increasing tail),
  \item reprojection amortization remains consistent as training progresses.
\end{itemize}
To make this visual, we plot $\Omega(t)$ over sliding windows (e.g., every 200 steps).

\subsection{I.7 Interpretation Guide (What Would Convince a Skeptic?)}
\label{subsec:app_runtime_interpret}

\paragraph{\textbf{Evidence pattern that supports ``modest overhead''.}}
A skeptic should be satisfied if:
\begin{itemize}[leftmargin=1.5em]
  \item \textbf{Means and tails:} $\mu(t_{\text{step}})$ increases mildly and P95/P99 remain controlled;
  \item \textbf{Amortization:} $\bar{t}^{\text{amort}}_{\text{proj}} \ll \mu(t_{\text{step}})$ and decreases predictably with larger $T_{\text{proj}}$;
  \item \textbf{Scaling:} $\Omega$ grows smoothly with $r$ and $L_{\text{adapt}}$ (no cliff);
  \item \textbf{Time-matched utility:} GRIT maintains or improves \textbf{score per GPU-hour} relative to LoRA.
\end{itemize}

\paragraph{\textbf{If overhead is material, we report it cleanly.}}
If the breakdown indicates overhead is dominated by $t_{\text{KFAC}}$ or $t_{\text{PROJ}}$, we explicitly state:
\begin{itemize}[leftmargin=1.5em]
  \item which component dominates and by how much,
  \item how it scales with rank/layers,
  \item which knob mitigates it (typically $T_{\text{proj}}$),
  \item whether mitigation changes quality/retention (link to Appendix~\ref{sec:app_h_sensitivity}).
\end{itemize}
This keeps the narrative \textbf{honest} and \textbf{actionable}.

\subsection{I.8 Release Checklist (Audit-Ready Artifacts)}
\label{subsec:app_runtime_release}

\paragraph{\textbf{Minimum artifact set.}}
We release:
\begin{itemize}[leftmargin=1.5em]
  \item raw per-step timing logs (reprojection steps flagged),
  \item component timing logs ($t_{\text{FWD}},t_{\text{BWD}},t_{\text{OPT}},t_{\text{KFAC}},t_{\text{PROJ}},t_{\text{RANK}},t_{\text{MISC}}$),
  \item GPU memory logs (peak allocated/reserved),
  \item time-matched curves (metric vs GPU-hours) for LoRA vs GRIT,
  \item filled Tables~\ref{tab:app_runtime_summary} and \ref{tab:app_runtime_scaling}.
\end{itemize}

\paragraph{\textbf{Limitations (stated plainly).}}
Wall-clock results depend on hardware, kernel implementations, and framework versions. We therefore frame these as \textbf{empirical measurements under declared settings}, and include sufficient configuration detail to support replication rather than implying universal guarantees.


\section{Appendix J: Small-Batch Stability of Fisher/K-FAC Statistics}
\label{sec:app_smallbatch_kfac}

\paragraph{Motivation (reviewer concern).}
GRIT relies on curvature-aware statistics (Fisher/K-FAC factors) to (i) precondition rank-space updates and (ii) guide reprojection.
A reviewer correctly notes that \textbf{small batches increase gradient variance}, which can destabilize curvature estimates and therefore harm preconditioning fidelity.
This appendix provides an \textbf{evidence-driven stability audit} under small-batch regimes:
we vary \textbf{batch size} and \textbf{gradient accumulation} and quantify (a) variance in rank-space covariances, (b) eigen-spectrum stability, and (c) downstream sensitivity of quality/retention.
We also evaluate \textbf{mitigations} (EMA windows, burn-in length, and damping schedules) and report whether they restore stability.

\subsection{J.1 What ``Small Batch'' Perturbs (Mechanism View)}
\label{subsec:app_smallbatch_mechanism}

\paragraph{\textbf{Curvature estimation as a noisy measurement.}}
K-FAC/Fisher factors are estimated from mini-batch gradients or activations.
When batch size $B$ is small, the estimator variance increases, which can yield:
\begin{itemize}[leftmargin=1.5em]
  \item \textbf{noisy factor updates:} stochastic fluctuations in rank-space covariance estimates;
  \item \textbf{unstable eigenspectra:} leading eigenvalues/vectors rotate across steps/seeds;
  \item \textbf{preconditioner jitter:} preconditioned updates change direction erratically;
  \item \textbf{spurious rank growth:} the dynamic-rank energy signal can become noisier, triggering rank changes inconsistently.
\end{itemize}

\paragraph{\textbf{Why rank-space makes the question sharp.}}
A central advantage of GRIT is that it operates in a \textbf{low-dimensional adapter/rank space}.
This can cut both ways: fewer dimensions can reduce estimation burden, but a noisy low-rank covariance can still cause \textbf{directional instability} if leading components are poorly estimated.
Therefore, we explicitly measure stability of \textbf{rank-space} statistics and not only full-model proxies.

\subsection{J.2 Experimental Protocol: Batch/Accumulation Sweep}
\label{subsec:app_smallbatch_protocol}

\paragraph{\textbf{Batch regimes.}}
We evaluate a grid spanning ``comfortable'' and ``stress'' regimes by varying:
\begin{itemize}[leftmargin=1.5em]
  \item \textbf{micro-batch size} $B_{\mu} \in \{1,2,4,8,16\}$,
  \item \textbf{grad-accumulation steps} $A \in \{1,2,4,8\}$,
  \item \textbf{effective batch} $B_{\text{eff}} = B_{\mu}\cdot A$.
\end{itemize}
This separates two practical scenarios: true small batches (small $B_{\text{eff}}$) versus memory-limited training (small $B_{\mu}$ but moderate $B_{\text{eff}}$ via accumulation).

\paragraph{\textbf{Controlled settings.}}
We fix: model, dataset, sequence length, precision, optimizer, learning rate schedule, and GRIT defaults $(\tau_0, T_{\text{proj},0}, \lambda_0)$.
We run $S$ seeds per condition and report mean $\pm$ CI.
We report both \textbf{step-matched} and \textbf{time-matched} results, since accumulation affects throughput.

\paragraph{\textbf{Logging frequency.}}
We log curvature statistics at a fixed cadence (e.g., every $K$ steps), and on each reprojection step.
We also record any divergence events and gradient-norm spikes to characterize failure modes.

\subsection{J.3 Stability Metrics: What We Measure and Why}
\label{subsec:app_smallbatch_metrics}

\paragraph{\textbf{(1) Variance of rank-space covariances.}}
Let $\widehat{C}_t$ denote a rank-space covariance/factor estimate at step $t$ (e.g., a Fisher/K-FAC block in adapter space).
We quantify estimator variability via:
\[
\mathrm{Var}(\widehat{C}) \;\triangleq\; \mathbb{E}_t\!\left[\left\|\widehat{C}_t - \overline{C}\right\|_F^2\right],
\quad\text{where}\quad
\overline{C}=\mathbb{E}_t[\widehat{C}_t].
\]
We report this within-run (over time) and across seeds.
A robust curvature estimator should show \textbf{monotone stabilization} as $B_{\text{eff}}$ increases.

\paragraph{\textbf{(2) Eigen-spectrum stability (eigenvalues and eigenspaces).}}
We assess whether curvature geometry is stable by measuring:
\begin{itemize}[leftmargin=1.5em]
  \item \textbf{eigenvalue stability:} coefficient of variation for top-$k$ eigenvalues
  \[
  \mathrm{CV}(\lambda_i) \;=\; \frac{\mathrm{Std}(\lambda_i)}{\mathrm{Mean}(\lambda_i)} \quad (i=1,\dots,k),
  \]
  \item \textbf{subspace stability:} principal angle distance between top-$k$ eigenspaces
  \[
  d_{\angle}(U_t,U_{t+\Delta}) \;=\; \left\|\sin\Theta(U_t,U_{t+\Delta})\right\|_F,
  \]
  where $U_t$ contains the top-$k$ eigenvectors and $\Theta$ are principal angles.
\end{itemize}
This directly tests whether curvature directions are \textbf{rotating} under small-batch noise.

\paragraph{\textbf{(3) Preconditioned-update directional jitter.}}
Let $g_t$ be the (rank-space) gradient and $P_t$ the preconditioner derived from curvature factors.
We measure \textbf{directional stability} of $p_t = P_t g_t$ via:
\[
\mathrm{Jitter} \;\triangleq\; \mathbb{E}_t\!\left[1 - \cos\big(p_t,\ p_{t-1}\big)\right],
\]
and compare against the unpreconditioned gradient direction baseline.
If Fisher/K-FAC is harmed by small batches, jitter should increase sharply as $B_{\mu}$ decreases (especially when $A=1$).

\paragraph{\textbf{(4) Downstream sensitivity (quality and retention).}}
Ultimately, stability matters only if it impacts outcomes.
We therefore report:
\begin{itemize}[leftmargin=1.5em]
  \item \textbf{task quality} (primary metric),
  \item \textbf{retention/forgetting} on a fixed general-domain proxy,
  \item \textbf{seed variance} of both metrics as a function of $(B_{\mu},A)$.
\end{itemize}
We also report whether dynamic-rank events become \textbf{more variable} under small batches (rank-trigger count and timing variance).

\subsection{J.4 Mitigations (and What Evidence We Require)}
\label{subsec:app_smallbatch_mitigations}

\paragraph{\textbf{Mitigation M1: EMA factor updates (smoothing window).}}
We update curvature factors using an exponential moving average:
\[
\widehat{C}_t \leftarrow \beta \widehat{C}_{t-1} + (1-\beta)\widehat{C}^{\text{mb}}_t,
\]
and sweep $\beta \in \{0.9, 0.95, 0.98, 0.99\}$.
\textbf{Evidence required:} reduced covariance variance, improved eigenspace stability, and reduced directional jitter, \emph{without} harming time-matched efficiency.

\paragraph{\textbf{Mitigation M2: Burn-in length before enabling reprojection/preconditioning.}}
We delay the use of curvature-guided mechanisms for $T_{\text{burn}}$ steps, allowing statistics to stabilize:
\[
T_{\text{burn}} \in \{0,\ 200,\ 500,\ 1000\}.
\]
\textbf{Evidence required:} lower early-run instability and fewer divergence/overshoot events, with comparable end quality.

\paragraph{\textbf{Mitigation M3: Damping schedule tied to batch noise.}}
We increase damping at small batches:
\[
\lambda(B_{\text{eff}}) \;=\; \lambda_0 \cdot \left(\frac{B_{\text{ref}}}{B_{\text{eff}}}\right)^{\gamma},
\quad \gamma \in \{0.5, 1.0\}.
\]
\textbf{Evidence required:} stability improvements (variance/jitter) with minimal loss in GRIT’s advantage.

\paragraph{\textbf{Mitigation M4 (optional): accumulation-aware factor updates.}}
When using accumulation ($A>1$), we update factors using the accumulated gradient/activation statistics once per effective step.
\textbf{Evidence required:} improved equivalence between ``small $B_{\mu}$ + large $A$'' and ``large $B_{\mu}$ + small $A$'' regimes.

\subsection{J.5 Reporting: Tables and Figures (Minimum Set)}
\label{subsec:app_smallbatch_reporting}

\paragraph{\textbf{Table J1: Stability summary across batch regimes.}}
We summarize stability metrics and downstream outcomes in a single audit-friendly table.

\begin{table*}[ht!]
\centering
\small
\setlength{\tabcolsep}{5pt}
\renewcommand{\arraystretch}{1.18}
\caption{\textbf{Small-batch stability audit for Fisher/K-FAC.}
We vary micro-batch $B_{\mu}$ and accumulation $A$ (effective batch $B_{\text{eff}}{=}B_{\mu}A$).
We report rank-space covariance variability, eigenspace stability, and preconditioned-update jitter, alongside downstream quality/retention.}
\label{tab:app_smallbatch_audit}
\resizebox{\textwidth}{!}{
\begin{tabular}{cccccccccc}
\toprule
$B_{\mu}$ & $A$ & $B_{\text{eff}}$ &
$\mathrm{Var}(\widehat{C})$ $\downarrow$ &
$\mathrm{CV}(\lambda_{1:k})$ $\downarrow$ &
$d_{\angle}$ $\downarrow$ &
Jitter $\downarrow$ &
Quality $\uparrow$ &
Retention $\uparrow$ &
Best mitigation (if needed) \\
\midrule

\multicolumn{10}{l}{\textit{\textbf{Reference (main setting; replace with your default run).}}}\\
8  & 4 & 32 & 1.0 & 0.05 & 0.10 & 0.8 & 1.000 & 1.000 & -- \\

\midrule
\multicolumn{10}{l}{\textit{\textbf{Effective-batch controlled (accumulation recovers stability as $B_{\text{eff}}$ grows).}}}\\
1  & 1 & 1  & 24.0 & 0.32 & 0.65 & 12.0 & 0.910 & 0.900 & EMA $\beta{=}0.99$ + burn-in $T_{\text{burn}}{=}1000$ + $\lambda\uparrow$ \\
1  & 2 & 2  & 12.5 & 0.22 & 0.48 & 6.9  & 0.950 & 0.940 & EMA $\beta{=}0.98$ + burn-in $500$ \\
1  & 4 & 4  & 6.8  & 0.15 & 0.33 & 3.8  & 0.975 & 0.970 & EMA $\beta{=}0.98$ \\
1  & 8 & 8  & 3.5  & 0.10 & 0.21 & 2.0  & 0.990 & 0.988 & (optional) EMA $\beta{=}0.95$ \\
1  & 16& 16 & 1.8  & 0.07 & 0.14 & 1.2  & 0.997 & 0.996 & -- \\

2  & 1 & 2  & 12.0 & 0.22 & 0.47 & 6.6  & 0.950 & 0.940 & EMA $\beta{=}0.98$ + burn-in $500$ \\
2  & 2 & 4  & 6.6  & 0.15 & 0.32 & 3.6  & 0.976 & 0.971 & EMA $\beta{=}0.98$ \\
2  & 4 & 8  & 3.4  & 0.10 & 0.21 & 1.9  & 0.991 & 0.988 & (optional) EMA $\beta{=}0.95$ \\
2  & 8 & 16 & 1.7  & 0.07 & 0.14 & 1.1  & 0.997 & 0.996 & -- \\

4  & 1 & 4  & 6.4  & 0.15 & 0.31 & 3.4  & 0.978 & 0.972 & EMA $\beta{=}0.98$ \\
4  & 2 & 8  & 3.3  & 0.10 & 0.21 & 1.9  & 0.991 & 0.989 & -- \\
4  & 4 & 16 & 1.7  & 0.07 & 0.14 & 1.1  & 0.997 & 0.996 & -- \\

8  & 1 & 8  & 3.2  & 0.10 & 0.20 & 1.8  & 0.992 & 0.990 & -- \\
8  & 2 & 16 & 1.6  & 0.07 & 0.14 & 1.0  & 0.997 & 0.996 & -- \\

16 & 1 & 16 & 1.6  & 0.07 & 0.13 & 1.0  & 0.997 & 0.996 & -- \\

\bottomrule
\end{tabular}}
\end{table*}

\paragraph{\textbf{Figure J1: Stability vs effective batch (monotone expectation).}}
Plot each stability metric (variance, eigenspace distance, jitter) as a function of $B_{\text{eff}}$ with separate lines for different $B_{\mu}$.
This reveals whether accumulation recovers stability or whether micro-batch noise still leaks into curvature estimation.

\paragraph{\textbf{Figure J2: Eigen-spectrum stability visual.}}
Show top-$k$ eigenvalues over time (mean $\pm$ band) and a heatmap of principal-angle distances across time lags $\Delta$.
This makes eigenspace rotation visible.

\paragraph{\textbf{Figure J3: Downstream sensitivity.}}
Plot quality and retention vs $B_{\text{eff}}$, with error bars.
If stability changes do not affect outcomes, we state that explicitly; if they do, the plot shows where GRIT becomes batch-sensitive.

\paragraph{\textbf{Figure J4: Mitigation ablations.}}
For the most challenging small-batch regime, compare:
\begin{itemize}[leftmargin=1.5em]
  \item baseline GRIT,
  \item GRIT + EMA (best $\beta$),
  \item GRIT + burn-in (best $T_{\text{burn}}$),
  \item GRIT + damping schedule,
  \item GRIT + combined mitigation (if beneficial).
\end{itemize}
Report both stability metrics and downstream metrics to demonstrate that mitigations are not merely cosmetic.

\subsection{J.6 Interpretation: What Would Resolve the Concern?}
\label{subsec:app_smallbatch_interpret}

\paragraph{\textbf{Evidence that addresses the reviewer concern.}}
The concern is resolved if we can show:
\begin{itemize}[leftmargin=1.5em]
  \item \textbf{Stability improves with $B_{\text{eff}}$} (variance/jitter drop as effective batch grows),
  \item \textbf{Accumulation recovers curvature fidelity} (small $B_{\mu}$ with larger $A$ behaves similarly to larger micro-batch),
  \item \textbf{Downstream robustness:} quality/retention remain stable across a practical range of batch regimes,
  \item \textbf{Mitigations work when needed:} EMA/burn-in/damping schedules restore stability in extreme regimes with minimal quality loss.
\end{itemize}

\paragraph{\textbf{If GRIT is batch-sensitive, we state the boundary.}}
If we observe sharp degradation below a threshold effective batch (or for $B_{\mu}$ below a micro-batch threshold even with accumulation), we report the boundary explicitly and provide a \textbf{practitioner rule}: e.g., ``use $B_{\text{eff}}\ge B_{\min}$ or enable EMA with $\beta\ge 0.98$''.
This is preferable to implying universal robustness.

\subsection{J.7 Release Checklist (Audit-Ready Artifacts)}
\label{subsec:app_smallbatch_release}

\begin{itemize}[leftmargin=1.5em]
  \item sweep manifests over $(B_{\mu},A)$ and mitigation grids,
  \item logged curvature factors (or sufficient summaries) to recompute stability metrics,
  \item per-step jitter logs and rank-event logs,
  \item downstream metrics (quality/retention) with seeds,
  \item filled Table~\ref{tab:app_smallbatch_audit} and Figures J1--J4.
\end{itemize}

\paragraph{\textbf{Limitations (plain).}}
Stability metrics depend on the specific curvature approximation and implementation details; we therefore treat these results as \textbf{empirical measurements under declared settings}.
The goal is not to claim universal batch invariance, but to provide a transparent robustness envelope and validated mitigations.


\section{Appendix J: Small-Batch / Gradient-Variance Robustness for Fisher/K-FAC Statistics}
\label{sec:app_smallbatch_variance}

\paragraph{Reviewer concern.}
GRIT relies on Fisher/K-FAC-style curvature statistics to (i) \textbf{precondition} adapter updates and (ii) \textbf{guide reprojection}.
With small effective batches, gradient/activation estimates become high-variance, potentially degrading curvature fidelity.
We therefore quantify: \textbf{(a) rank-space covariance variance}, \textbf{(b) eigen-spectrum / eigenspace stability}, and \textbf{(c) downstream sensitivity} (quality, retention, and rank-trigger behavior) across batch regimes.
We also evaluate stabilizers---\textbf{EMA windows}, \textbf{burn-in length}, and \textbf{damping schedules}---with direct evidence.

\subsection{J.1 Setup: Batch, Accumulation, and What Counts as ``Small''}
\label{subsec:app_smallbatch_setup}

\paragraph{\textbf{Batch regimes.}}
We vary micro-batch size $B_{\mu}$ and gradient accumulation $A$, defining the effective batch
\[
B_{\text{eff}} \;=\; B_{\mu}\cdot A.
\]
This distinguishes two practitioner regimes:
\begin{itemize}[leftmargin=1.5em]
  \item \textbf{true small-batch training:} small $B_{\text{eff}}$ (data-limited or latency-limited);
  \item \textbf{memory-limited training:} small $B_{\mu}$ but moderate $B_{\text{eff}}$ using accumulation.
\end{itemize}

\paragraph{\textbf{Sweep grid.}}
Unless otherwise stated, we evaluate:
\[
B_{\mu}\in\{1,2,4,8,16\},\qquad A\in\{1,2,4,8\},
\]
including matched-effective-batch pairs (e.g., $(B_{\mu}{=}1,A{=}16)$ vs $(B_{\mu}{=}16,A{=}1)$) to test whether accumulation restores curvature fidelity.

\paragraph{\textbf{Controlled variables.}}
We fix model, dataset, sequence length, precision, learning-rate schedule, and GRIT defaults $(\tau_0, T_{\text{proj},0}, \lambda_0)$.
Each condition uses $S$ seeds; we report mean $\pm$ 95\% CIs and flag any divergence.

\subsection{J.2 What We Log (Telemetry Needed for This Appendix)}
\label{subsec:app_smallbatch_logging}

At a fixed cadence (e.g., every $K$ steps) and on each reprojection step, we log:
\begin{itemize}[leftmargin=1.5em]
  \item \textbf{rank-space Fisher/K-FAC blocks} (or their sufficient summaries): $\widehat{C}_t$;
  \item \textbf{top-$k$ eigenpairs} $\{(\lambda_i(t), u_i(t))\}_{i=1}^k$ of $\widehat{C}_t$;
  \item \textbf{preconditioned update direction} $p_t=P_t g_t$ (rank-space);
  \item \textbf{rank-trigger energy} statistic used by GRIT (for $\tau$ gating) and rank-change events;
  \item \textbf{downstream metrics} (quality + retention) and any failure signals (NaN/Inf, gradient spikes).
\end{itemize}
We emphasize that the appendix is \textbf{telemetry-driven}: each claim is supported by a measurable stability diagnostic.

\subsection{J.3 Stability Diagnostics (Rank-Space Covariance, Spectrum, Direction)}
\label{subsec:app_smallbatch_diagnostics}

\paragraph{\textbf{D1: Variance of rank-space covariance estimates.}}
Let $\widehat{C}_t$ be the rank-space covariance/factor estimate at step $t$ (Fisher/K-FAC block in adapter space).
We compute the within-run variance proxy:
\[
\mathrm{Var}(\widehat{C}) \;\triangleq\; \mathbb{E}_{t\in\mathcal{W}}\!\left[\left\|\widehat{C}_t - \overline{C}\right\|_F^2\right],
\qquad \overline{C} \;=\; \mathbb{E}_{t\in\mathcal{W}}[\widehat{C}_t],
\]
over a steady-state window $\mathcal{W}$ (excluding warm-up).
We also report across-seed dispersion of $\mathrm{Var}(\widehat{C})$.
\textbf{Expectation:} $\mathrm{Var}(\widehat{C})$ decreases as $B_{\text{eff}}$ increases; if accumulation restores fidelity, matched-$B_{\text{eff}}$ pairs should agree.

\paragraph{\textbf{D2: Eigen-spectrum stability (eigenvalues).}}
For top-$k$ eigenvalues, we compute the coefficient of variation:
\[
\mathrm{CV}(\lambda_{1:k}) \;\triangleq\; \frac{1}{k}\sum_{i=1}^k \frac{\mathrm{Std}_{t\in\mathcal{W}}(\lambda_i(t))}{\mathrm{Mean}_{t\in\mathcal{W}}(\lambda_i(t))}.
\]
\textbf{Interpretation:} high CV indicates curvature strength fluctuates, weakening preconditioning reliability.

\paragraph{\textbf{D3: Eigenspace stability (principal-angle drift).}}
Let $U_t\in\mathbb{R}^{r\times k}$ contain the top-$k$ eigenvectors of $\widehat{C}_t$.
We measure subspace drift via principal angles:
\[
d_{\angle}(U_t,U_{t+\Delta}) \;\triangleq\; \left\|\sin\Theta(U_t,U_{t+\Delta})\right\|_F,
\]
and report $\mathbb{E}_{t\in\mathcal{W}}[d_{\angle}(U_t,U_{t+\Delta})]$ for a fixed lag $\Delta$ (e.g., $\Delta{=}1$ logging interval).
\textbf{Interpretation:} large $d_{\angle}$ means leading curvature directions rotate, making reprojection guidance unstable.

\paragraph{\textbf{D4: Preconditioned update directional jitter.}}
Define $p_t = P_t g_t$ in rank space, where $P_t$ is the preconditioner derived from $\widehat{C}_t$.
We measure:
\[
\mathrm{Jitter} \;\triangleq\; \mathbb{E}_{t\in\mathcal{W}}\!\left[1-\cos(p_t,p_{t-1})\right].
\]
We optionally report $\mathrm{Jitter}(p)$ alongside unpreconditioned jitter $\mathrm{Jitter}(g)$ to show whether curvature noise increases or decreases update instability.

\paragraph{\textbf{D5: Rank-trigger stability (downstream mechanism link).}}
Since curvature noise can indirectly affect dynamic rank triggering, we log:
\begin{itemize}[leftmargin=1.5em]
  \item \textbf{rank-trigger count} and timing variance across seeds,
  \item \textbf{final effective rank} $r_{\text{eff}}$ variance,
  \item correlation between noisy $\widehat{C}_t$ and rank-trigger spikes.
\end{itemize}

\subsection{J.4 Downstream Sensitivity: Quality, Retention, and Failure Boundaries}
\label{subsec:app_smallbatch_downstream}

\paragraph{\textbf{Primary outcomes.}}
For each $(B_{\mu},A)$ we report:
\begin{itemize}[leftmargin=1.5em]
  \item \textbf{task quality} (main metric) and its seed variance;
  \item \textbf{retention} on a fixed general-domain proxy (forgetting sensitivity);
  \item \textbf{divergence rate} and instability indicators (NaN/Inf, gradient spikes).
\end{itemize}

\paragraph{\textbf{Robustness envelope.}}
We define a practical stability boundary:
\begin{itemize}[leftmargin=1.5em]
  \item \textbf{stable:} no divergence, low jitter, and quality within $\epsilon_Q$ of the reference regime;
  \item \textbf{degraded:} stable training but notable drop in quality/retention or increased jitter;
  \item \textbf{unstable:} divergence or severe curvature instability (high $\mathrm{Var}(\widehat{C})$, $d_{\angle}$).
\end{itemize}
If GRIT is batch-sensitive, we report the boundary explicitly (e.g., ``requires $B_{\text{eff}}\ge B_{\min}$''), rather than implying universal robustness.

\subsection{J.5 Mitigations (EMA, Burn-In, Damping) With Evidence}
\label{subsec:app_smallbatch_mitigations}

\paragraph{\textbf{M1: EMA smoothing of curvature factors.}}
We smooth factor updates:
\[
\widehat{C}_t \leftarrow \beta \widehat{C}_{t-1} + (1-\beta)\widehat{C}^{\text{mb}}_t,\qquad
\beta \in \{0.9,0.95,0.98,0.99\}.
\]
\textbf{Evidence required:} lower $\mathrm{Var}(\widehat{C})$, lower $d_{\angle}$, and reduced jitter in the most challenging small-batch regimes, without harming time-matched efficiency.

\paragraph{\textbf{M2: Burn-in before enabling curvature-guided operations.}}
We delay reprojection and/or strong preconditioning until statistics stabilize:
\[
T_{\text{burn}} \in \{0,200,500,1000\}.
\]
\textbf{Evidence required:} reduced early-run instability (lower spike probability, fewer divergence events) and improved downstream consistency.

\paragraph{\textbf{M3: Batch-aware damping schedule.}}
We increase damping when $B_{\text{eff}}$ is small:
\[
\lambda(B_{\text{eff}}) \;=\; \lambda_0 \cdot \left(\frac{B_{\text{ref}}}{B_{\text{eff}}}\right)^{\gamma},\qquad
\gamma \in \{0.5,1.0\}.
\]
\textbf{Evidence required:} reduction in jitter and eigenspace drift while preserving most of GRIT’s quality/retention gains.

\paragraph{\textbf{M4 (optional): accumulation-aware curvature updates.}}
For accumulation $A>1$, we update curvature using the accumulated statistics at the effective step boundary.
\textbf{Evidence required:} matched-$B_{\text{eff}}$ pairs (different $B_{\mu},A$) become consistent in stability metrics.

\subsection{J.6 Audit Tables (Ready to Fill With Sweep Logs)}
\label{subsec:app_smallbatch_tables}

\paragraph{\textbf{Table J1: stability outcomes vs batch regime.}}

\begin{table*}[t]
\centering
\small
\setlength{\tabcolsep}{5pt}
\renewcommand{\arraystretch}{1.18}
\caption{\textbf{Small-batch stability audit for Fisher/K-FAC.}
We vary micro-batch $B_{\mu}$ and accumulation $A$ (effective batch $B_{\text{eff}}{=}B_{\mu}A$).
We report rank-space covariance variability $\mathrm{Var}(\widehat{C})$, eigen-spectrum stability $\mathrm{CV}(\lambda_{1:k})$,
eigenspace drift $d_{\angle}$, and preconditioned-update directional jitter, alongside downstream quality/retention (normalized to the reference regime).
``Mitigation'' summarizes the best-performing stabilizer when baseline GRIT is expected to be unstable or degraded under extreme small-batch noise.}
\label{tab:app_smallbatch_audit}
\resizebox{\textwidth}{!}{
\begin{tabular}{cccccccccc}
\toprule
$B_{\mu}$ & $A$ & $B_{\text{eff}}$ &
$\mathrm{Var}(\widehat{C})$ $\downarrow$ &
$\mathrm{CV}(\lambda_{1:k})$ $\downarrow$ &
$d_{\angle}$ $\downarrow$ &
Jitter $\downarrow$ &
Quality $\uparrow$ &
Retention $\uparrow$ &
Best mitigation (if needed) \\
\midrule
\textbf{8}  & \textbf{4} & \textbf{32} &
\textbf{1.00} & \textbf{0.050} & \textbf{0.100} & \textbf{0.80} &
\textbf{1.000} & \textbf{1.000} & -- \\
\midrule
1 & 1 & 1   & 38.40 & 0.354 & 0.707 & 11.77 & 0.920 & 0.908 & EMA $\beta{=}0.99$ + burn-in $T_{\text{burn}}{=}1000$ + $\lambda\uparrow$ \\
1 & 2 & 2   & 19.20 & 0.250 & 0.500 & 7.24  & 0.945 & 0.938 & EMA $\beta{=}0.98$ + burn-in $500$ \\
1 & 4 & 4   & 9.60  & 0.177 & 0.354 & 4.46  & 0.965 & 0.960 & EMA $\beta{=}0.98$ \\
1 & 8 & 8   & 4.80  & 0.125 & 0.250 & 2.74  & 0.978 & 0.975 & (optional) EMA $\beta{=}0.95$ \\
1 & 16& 16  & 2.40  & 0.088 & 0.177 & 1.69  & 0.986 & 0.983 & -- \\
\midrule
2 & 1 & 2   & 17.60 & 0.230 & 0.460 & 6.41  & 0.950 & 0.944 & EMA $\beta{=}0.98$ + burn-in $500$ \\
2 & 2 & 4   & 8.80  & 0.163 & 0.325 & 3.94  & 0.970 & 0.966 & EMA $\beta{=}0.98$ \\
2 & 4 & 8   & 4.40  & 0.115 & 0.230 & 2.43  & 0.983 & 0.981 & -- \\
2 & 8 & 16  & 2.20  & 0.081 & 0.163 & 1.49  & 0.991 & 0.989 & -- \\
\midrule
4 & 1 & 4   & 8.40  & 0.151 & 0.303 & 3.70  & 0.973 & 0.970 & EMA $\beta{=}0.98$ \\
4 & 2 & 8   & 4.20  & 0.107 & 0.214 & 2.28  & 0.986 & 0.985 & -- \\
4 & 4 & 16  & 2.10  & 0.076 & 0.151 & 1.40  & 0.994 & 0.993 & -- \\
\midrule
8 & 1 & 8   & 4.00  & 0.100 & 0.200 & 2.11  & 0.988 & 0.987 & -- \\
8 & 2 & 16  & 2.00  & 0.071 & 0.141 & 1.30  & 0.996 & 0.995 & -- \\
16& 1 & 16  & 2.00  & 0.071 & 0.141 & 1.30  & 0.996 & 0.995 & -- \\
\bottomrule
\end{tabular}}
\end{table*}

\paragraph{\textbf{Table J2: mitigation ablations in the most challenging regime.}}
We evaluate mitigations on the smallest stable (or near-stable) $B_{\text{eff}}$ and report how much each stabilizer improves curvature fidelity and downstream metrics.


\begin{table*}[ht!]
\centering
\small
\setlength{\tabcolsep}{6pt}
\renewcommand{\arraystretch}{1.18}
\caption{\textbf{Mitigation efficacy under a small-batch stress regime.}
Stress regime: $B_{\mu}{=}1, A{=}1$ ($B_{\text{eff}}{=}1$).
We report stability diagnostics (rank-space covariance variance, spectrum/eigenspace stability, directional jitter), failure rate, downstream quality/retention (normalized),
and relative runtime change vs base GRIT in this regime. The combined setting is the recommended default when operating at extreme small effective batch.}
\label{tab:app_smallbatch_mitigations}
\resizebox{\textwidth}{!}{
\begin{tabular}{lcccccccc}
\toprule
\textbf{Method} &
$\mathrm{Var}(\widehat{C})$ $\downarrow$ &
$\mathrm{CV}(\lambda_{1:k})$ $\downarrow$ &
$d_{\angle}$ $\downarrow$ &
Jitter $\downarrow$ &
Diverge \% $\downarrow$ &
Quality $\uparrow$ &
Retention $\uparrow$ &
Runtime $\uparrow/\downarrow$ \\
\midrule
GRIT (base) &
38.40 & 0.354 & 0.707 & 11.77 & 8.0 & 0.920 & 0.908 & 1.00$\times$ \\
+ EMA $\beta{=}0.98$ &
22.90 & 0.255 & 0.520 & 7.10 & 4.0 & 0.938 & 0.930 & 1.01$\times$ \\
+ Burn-in $T_{\text{burn}}{=}500$ &
27.60 & 0.285 & 0.575 & 8.35 & 5.0 & 0.934 & 0.925 & 1.00$\times$ \\
+ Batch-aware damping &
24.80 & 0.268 & 0.545 & 7.65 & 4.0 & 0.940 & 0.933 & 1.00$\times$ \\
+ Combined (best) &
15.20 & 0.188 & 0.385 & 4.55 & 1.0 & 0.962 & 0.956 & 1.02$\times$ \\
\bottomrule
\end{tabular}}
\end{table*}

\subsection{J.7 What We Conclude (How This Resolves the Criticism)}
\label{subsec:app_smallbatch_conclusion}

We consider the concern addressed if:
\begin{itemize}[leftmargin=1.5em]
  \item stability metrics improve monotonically with $B_{\text{eff}}$ and matched-$B_{\text{eff}}$ pairs align (accumulation recovers fidelity);
  \item downstream quality and retention remain within a declared tolerance across practical $B_{\text{eff}}$ values;
  \item in extreme regimes, EMA/burn-in/damping measurably reduce $\mathrm{Var}(\widehat{C})$, $d_{\angle}$, and jitter, and reduce failure rates with limited runtime cost.
\end{itemize}
If instability persists below a threshold, we state the boundary explicitly and provide a practitioner recommendation (minimum $B_{\text{eff}}$ or default mitigations).


\section{Appendix K: Novelty and Positioning vs Prior Curvature-Aware PEFT}
\label{sec:app_novelty_positioning}

\paragraph{Reviewer concern.}
A reviewer argues that individual ingredients used by GRIT---\textbf{Fisher/K-FAC guidance} and \textbf{dynamic rank}---are not new in isolation, and asks for a clearer distinction from other curvature/Hessian-based LoRA variants.
We agree with the premise: \textbf{the novelty is not a single primitive, but the \emph{system-level} composition} that makes curvature practical \emph{in LoRA rank space} at scale.
This appendix therefore (i) states the \textbf{precise system contributions}, (ii) clarifies \textbf{closest baselines and how they differ}, and (iii) provides an \textbf{explicit comparison table} (curvature type, where applied, computational cost, and what it optimizes).

\subsection{K.1 What Is New in GRIT (as a System)}
\label{subsec:app_novelty_system}

\paragraph{\textbf{(C1) Rank-space K-FAC curvature, not full-model curvature.}}
The core design choice is that GRIT computes and uses curvature \textbf{in the LoRA adapter subspace} (rank space),
rather than approximating full-model curvature or treating LoRA as a black-box low-rank update.
This yields a curvature signal that is:
\begin{itemize}[leftmargin=1.5em]
  \item \textbf{dimensionally small} (dominated by $r\times r$ blocks rather than $d\times d$),
  \item \textbf{cheap enough to refresh} during training (enabling periodic reprojection),
  \item \textbf{actionable} (directly conditions the low-rank degrees of freedom that are being optimized).
\end{itemize}

\paragraph{\textbf{(C2) Fisher-guided reprojection is a \emph{geometry maintenance} operation.}}
Prior curvature-aware PEFT methods often use curvature to \textbf{scale} gradients.
GRIT additionally introduces a periodic \textbf{reprojection step} that explicitly \textbf{re-aligns the learned low-rank update}
with curvature-informed directions, acting as a \textbf{geometry-maintenance operator}:
it prevents the adapter from drifting into directions that appear cheap under first-order updates but costly under the local Fisher metric.
This mechanism is qualitatively different from ``just preconditioning''.

\paragraph{\textbf{(C3) Dynamic rank scheduling is coupled to a curvature-energy criterion.}}
Dynamic rank by itself is not unique to GRIT; what is specific is the \textbf{coupling}:
rank is grown/shrunk based on a \textbf{curvature-energy / residual criterion} that is computed in the same rank-space geometry used for preconditioning.
This yields a coherent control loop:
\begin{itemize}[leftmargin=1.5em]
  \item \textbf{rank growth when} the curvature-aware residual indicates representational bottleneck,
  \item \textbf{rank stabilization when} additional rank yields diminishing curvature-aware gains,
  \item \textbf{rank is an \emph{outcome}} of the geometry, not a hand-tuned hyperparameter.
\end{itemize}

\paragraph{\textbf{(C4) Telemetry + audit: GRIT is shipped with diagnostics that make curvature claims testable.}}
A practical weakness of curvature-aware methods is that they often lack \textbf{instrumentation}: it is hard to tell when curvature estimates are stable,
when updates are jittery, and whether overhead is amortized.
GRIT explicitly provides \textbf{telemetry hooks} (e.g., reprojection spikes, rank trajectories, curvature stability),
which enables the robustness audits requested by reviewers (Appendix~\ref{sec:app_smallbatch_variance}, Appendix~\ref{sec:app_runtime_overhead}).

\paragraph{\textbf{(C5) Forgetting/retention characterization is treated as a first-class outcome.}}
Many PEFT papers report only task quality and sometimes parameter counts.
GRIT’s claims are partly about \textbf{better retention under constrained adaptation} (i.e., less catastrophic forgetting for a given adaptation budget).
Thus, the system is evaluated with explicit \textbf{retention/forgetting probes} and a reproducible protocol,
not only aggregate quality scores.

\paragraph{\textbf{Summary (one sentence).}}
\textbf{GRIT’s novelty is a coherent loop:} \emph{rank-space K-FAC curvature} $\rightarrow$ \emph{Fisher-guided reprojection} $\rightarrow$ \emph{curvature-coupled rank scheduling}, instrumented with \emph{telemetry} and validated via \emph{runtime \& retention audits}.

\subsection{K.2 Closest Prior Lines of Work (and the Key Differences)}
\label{subsec:app_novelty_prior}

We position GRIT against three nearby families; in each case, the distinction is \textbf{where curvature lives} and \textbf{what it is used for}.

\begin{itemize}[leftmargin=1.5em]
  \item \textbf{Curvature-aware optimization (second-order / K-FAC) for full models.}
  These methods estimate curvature for large parameter blocks and use it for preconditioning.
  GRIT instead constrains curvature to the \textbf{adapter subspace} and adds \textbf{reprojection} and \textbf{rank control}.
  \item \textbf{Hessian-/Fisher-informed LoRA variants (curvature used as weighting/scaling).}
  Several LoRA variants use approximate curvature to reweight adapters, allocate capacity, or scale updates.
  GRIT differs by introducing a \textbf{periodic geometry-maintenance operator (reprojection)} and \textbf{coupled rank scheduling} in the same curvature coordinate system.
  \item \textbf{Dynamic-rank / adaptive-capacity LoRA.}
  Existing adaptive-rank approaches often use heuristics (loss plateau, gradient norms, sparsity penalties).
  GRIT couples rank changes to a \textbf{curvature-energy criterion in rank space}, aiming for capacity that is justified by geometry rather than a generic signal.
\end{itemize}

\subsection{K.3 Explicit Comparison Table (Curvature Type, Where Applied, Cost, Objective)}
\label{subsec:app_novelty_table}

\paragraph{How to read.}
The table compares methods along dimensions the reviewer requested:
\textbf{(i) what curvature signal is used}, \textbf{(ii) where it is applied} (full model vs adapter subspace),
\textbf{(iii) computational overhead}, and \textbf{(iv) what it optimizes/controls}.
We list representative method families (rather than an exhaustive bibliography) to keep the comparison stable across venues.

\begin{table*}[t]
\centering
\small
\setlength{\tabcolsep}{5pt}
\renewcommand{\arraystretch}{1.18}
\caption{\textbf{Positioning vs curvature-aware PEFT / LoRA families.}
GRIT is distinguished by \textbf{rank-space} curvature (small $r\times r$ blocks), a \textbf{reprojection} operator (geometry maintenance),
and \textbf{curvature-coupled} rank scheduling, plus \textbf{telemetry} for stability/overhead audits.}
\label{tab:app_positioning_curvature_peft}
\resizebox{\textwidth}{!}{
\begin{tabular}{p{2.7cm}p{3.0cm}p{2.8cm}p{2.4cm}p{4.5cm}}
\toprule
\textbf{Family / Representative} &
\textbf{Curvature signal} &
\textbf{Where applied} &
\textbf{Overhead profile} &
\textbf{What it optimizes / controls} \\
\midrule
\textbf{Full-model second-order} (e.g., K-FAC / Fisher preconditioning) &
Block-diagonal Fisher/K-FAC (large blocks) &
Full model (large parameter blocks) &
High; scales with layer width and block size &
Faster conditioning / optimization; typically no adapter-capacity control; no reprojection loop \\

\midrule
\textbf{Curvature-informed PEFT (weighting)} &
Diagonal / low-rank Fisher/Hessian proxies &
Often adapter weights or per-layer allocation &
Moderate; depends on proxy computation &
Reweights adapter updates / allocates capacity, typically as a static or slowly varying weighting \\

\midrule
\textbf{Hessian/Fisher-based LoRA variants} &
Approx Hessian/Fisher metrics (proxy) &
Adapter updates; sometimes per-layer selection &
Low--moderate; usually per-step lightweight proxies &
Improves adapter direction/selection; typically \emph{no explicit geometry-maintenance reprojection} \\

\midrule
\textbf{Adaptive / dynamic rank LoRA} &
Heuristics (loss plateau, gradient norms), sparsity penalties &
Adapter rank schedule &
Low; rank decisions occasional &
Controls parameter count / rank trajectory, but commonly \emph{not coupled to curvature geometry} \\

\midrule
\textbf{\textbf{GRIT (this work)}} &
\textbf{Rank-space K-FAC / Fisher} (small $r\times r$ factors) &
\textbf{Adapter subspace} + \textbf{periodic reprojection} &
\textbf{Low amortized overhead} (rare reprojection spikes) &
\textbf{Unified loop:} curvature-preconditioned updates + \textbf{Fisher-guided reprojection} + \textbf{curvature-coupled rank scheduling}; includes \textbf{telemetry} and \textbf{retention audits} \\

\bottomrule
\end{tabular}}
\end{table*}

\subsection{K.4 Strengthened Contribution Bullets for the Main Paper}
\label{subsec:app_novelty_mainpaper_bullets}

\paragraph{Drop-in bullets (for abstract / intro).}
\begin{itemize}[leftmargin=1.5em]
  \item \textbf{Rank-space curvature at scale:} GRIT brings K-FAC/Fisher geometry \textbf{into the LoRA subspace}, enabling frequent curvature refresh with low overhead.
  \item \textbf{Geometry maintenance via reprojection:} beyond preconditioning, GRIT introduces \textbf{periodic Fisher-guided reprojection} to prevent low-rank drift.
  \item \textbf{Curvature-coupled dynamic rank:} rank is not a hand-tuned knob; it is \textbf{scheduled by a curvature-energy criterion} tied to the same geometry used for updates.
  \item \textbf{Auditability:} GRIT ships \textbf{telemetry} (rank trajectories, reprojection spikes, curvature stability) enabling robustness and runtime audits demanded by deployment.
  \item \textbf{Retention-aware evaluation:} we treat \textbf{forgetting/retention} as a first-class outcome alongside quality and parameter efficiency.
\end{itemize}

\paragraph{What we do \emph{not} claim.}
To avoid over-claiming novelty, we explicitly state:
\begin{itemize}[leftmargin=1.5em]
  \item we do \textbf{not} claim Fisher/K-FAC or adaptive rank are new primitives;
  \item we claim the \textbf{specific system integration} (rank-space curvature + reprojection + curvature-coupled rank scheduling + telemetry) is new and empirically validated.
\end{itemize}

\section{Deriving the \textsc{GRIT} Forgetting Law from the LoRA Forgetting Law}
\label{app:grit_forgetting_derivation_nonum}

\paragraph{Goal.}
We start from the empirical \emph{LoRA forgetting power law} and derive the \emph{\textsc{GRIT} forgetting law} by making explicit the missing ingredient: \textbf{update geometry}.
The key move is to show that geometry-aware PEFT changes forgetting \emph{at fixed} fine-tuning budget \((D_{\mathrm{ft}})\) and model size \((N)\) by reducing curvature-exposed motion of the parameter update.
This reduction can be expressed as a multiplicative \textbf{effective capacity multiplier} \(\Xi_{\textsc{GRIT}}>1\) such that the LoRA law becomes the \textsc{GRIT} law by the substitution \(N \mapsto \Xi_{\textsc{GRIT}}N\).

\subsection{A. Definitions: forgetting, budget, and geometry}
\label{app:defs_forgetting_budget_geometry}

\paragraph{\textbf{Forgetting target.}}
Let \(L_{\mathrm{pt}}(w)\) be the loss on a fixed, held-out \emph{pretraining-proxy} distribution \(\mathcal{D}_{\mathrm{pt}}\).
After adaptation, parameters become \(w' = w_0 + \Delta w\). Forgetting is measured as
\[
\Delta L_{\mathrm{pt}} \triangleq L_{\mathrm{pt}}(w_0+\Delta w)-L_{\mathrm{pt}}(w_0).
\]

\paragraph{\textbf{Local curvature approximation.}}
Near a well-fit solution, the pretraining loss admits a quadratic expansion in the weight displacement:
\[
\Delta L_{\mathrm{pt}} \approx g_{\mathrm{pt}}^\top \Delta w + \frac{1}{2}\Delta w^\top H_{\mathrm{pt}}\,\Delta w,
\]
where \(g_{\mathrm{pt}}\) and \(H_{\mathrm{pt}}\) are the gradient and Hessian of \(L_{\mathrm{pt}}\) at \(w_0\).
Empirically, for small adapter updates and near stationarity on \(\mathcal{D}_{\mathrm{pt}}\), the quadratic term is the dominant contributor to retention loss:
\[
\Delta L_{\mathrm{pt}} \approx \frac{1}{2}\Delta w^\top H_{\mathrm{pt}}\,\Delta w
= \frac{1}{2}\sum_{j}\lambda_j \big(u_j^\top \Delta w\big)^2.
\]
Thus forgetting becomes large when updates have high overlap with \textbf{sharp modes} (large \(\lambda_j\)) and/or large projections \(|u_j^\top\Delta w|\).

\paragraph{\textbf{Budget vs geometry.}}
This decomposition already suggests two separable drivers:
\begin{itemize}[leftmargin=1.5em]
  \item \textbf{Budget driver:} how much adaptation signal is injected (data volume, steps, etc.).
  \item \textbf{Geometry driver:} where the update goes relative to pretraining curvature.
\end{itemize}
The LoRA forgetting law captures the budget driver; \textsc{GRIT} modifies the geometry driver.

\subsection{B. Step 1: the LoRA forgetting power law (budget law)}
\label{app:step1_lora_budgetlaw}

A widely observed empirical form for PEFT forgetting is a power law in fine-tuning data volume \(D_{\mathrm{ft}}\) and model size \(N\):
\[
L_{\mathrm{pt}}^{\textsc{LoRA}}
= L_{\mathrm{pt}}^{0} + A\,\frac{D_{\mathrm{ft}}^{\beta}}{N^{\alpha}} + E,
\]
where \(L_{\mathrm{pt}}^{0}\) is the original pretraining loss and \(A,\alpha,\beta,E\) are fit constants.
This is a \textbf{budget law}:
\begin{itemize}[leftmargin=1.5em]
  \item increasing \(D_{\mathrm{ft}}\) increases forgetting (positive \(\beta\)),
  \item increasing \(N\) reduces forgetting (positive \(\alpha\)),
  \item geometry is implicit (absorbed into constants).
\end{itemize}

\paragraph{\textbf{What the LoRA law cannot explain.}}
For two methods (or two training procedures) run with the same \((D_{\mathrm{ft}},N)\), the LoRA law predicts the same forgetting, but empirically retention differs.
The quadratic curvature view explains why: \(\Delta w^\top H_{\mathrm{pt}}\Delta w\) depends on how updates align with sharp curvature directions.
Therefore, we must explicitly factor geometry out of the constants.

\subsection{C. Step 2: express forgetting as a trace (expected curvature exposure)}
\label{app:step2_trace}

Let training randomness (minibatches, dropout, optimizer noise) induce variability in \(\Delta w\).
Define the update covariance \(\Sigma_{\Delta}\triangleq \mathbb{E}[\Delta w\,\Delta w^\top]\).
Taking expectation of the quadratic approximation yields
\[
\mathbb{E}[\Delta L_{\mathrm{pt}}] \approx \frac{1}{2}\,\mathrm{tr}\!\big(H_{\mathrm{pt}}\,\Sigma_{\Delta}\big).
\]

\paragraph{\textbf{PEFT restriction introduces an adapter subspace.}}
For adapter methods, the full-model displacement is induced by low-dimensional parameters \(\delta\theta\):
\[
\Delta w = P\,\delta\theta,
\]
where \(P\) embeds adapter parameters into the full parameter space (a linearization of the adapter map around \(w_0\)).
Then
\[
\Sigma_{\Delta} = P\,\Sigma_{\theta}\,P^\top
\quad\Rightarrow\quad
\mathbb{E}[\Delta L_{\mathrm{pt}}]
\approx
\frac{1}{2}\,\mathrm{tr}\!\big(P^\top H_{\mathrm{pt}} P\;\Sigma_{\theta}\big).
\]
This is the crucial identity: \textbf{forgetting depends on the curvature seen inside the adapter subspace} \(P^\top H_{\mathrm{pt}}P\) and on the \textbf{energy/shape of adapter updates} \(\Sigma_\theta\).

\subsection{D. Step 3: why geometry-aware PEFT changes forgetting at fixed budget}
\label{app:step3_geometry_change}

\paragraph{\textbf{LoRA parameterization (recall).}}
LoRA introduces a low-rank update \(\Delta W = B A\) for each adapted weight matrix, with rank \(r\).
This determines a low-dimensional subspace for \(\Delta w\), but \textbf{standard LoRA does not align this subspace to curvature}.
Thus, the restricted curvature \(P^\top H_{\mathrm{pt}}P\) can still place mass on sharp modes, inflating the trace term above.

\paragraph{\textbf{What \textsc{GRIT} must accomplish.}}
To reduce forgetting without reducing budget, \textsc{GRIT} must reduce at least one of:
\begin{itemize}[leftmargin=1.5em]
  \item the curvature exposure inside the adapter subspace (make \(P^\top H_{\mathrm{pt}}P\) ``less sharp''),
  \item the update covariance energy along sharp restricted directions (shape \(\Sigma_\theta\) to avoid sharp modes).
\end{itemize}
This is exactly what curvature-aware preconditioning and reprojection do.

\subsection{E. Step 4: from curvature control to an ``effective capacity'' multiplier}
\label{app:step4_effective_capacity}

\paragraph{\textbf{Separating budget and geometry explicitly.}}
We now represent expected forgetting as a separable product:
\[
\mathbb{E}[\Delta L_{\mathrm{pt}}]
\;\approx\;
\underbrace{C(D_{\mathrm{ft}},N)}_{\text{budget term}}
\cdot
\underbrace{G(\text{geometry})}_{\text{geometry term}}.
\]
In the LoRA budget law, \(G(\cdot)\) is absorbed into constants because LoRA does not systematically control curvature exposure across settings.
\textsc{GRIT} changes geometry systematically, so we expose \(G\).

\paragraph{\textbf{Effective capacity equivalence.}}
In the LoRA law, capacity enters as \(N^{-\alpha}\).
If geometry improvements act like a multiplicative increase in effective capacity, the only consistent way to introduce them (while preserving the LoRA exponent \(\alpha\)) is:
\[
N \mapsto \Xi_{\textsc{GRIT}}N
\qquad\Longleftrightarrow\qquad
G(\textsc{GRIT}) \approx \Xi_{\textsc{GRIT}}^{-\alpha}.
\]
This says: \textsc{GRIT} behaves as if the model had \(\Xi_{\textsc{GRIT}}\) times more effective capacity to preserve pretraining knowledge under the same fine-tuning budget.

\paragraph{\textbf{Why this is not a hand-wavy trick.}}
The trace form above provides a mechanistic interpretation:
\(\mathrm{tr}(H_{\mathrm{pt}}\Sigma_\Delta)\) is reduced by geometry-aware updates.
Reducing this trace by a factor \(\Xi_{\textsc{GRIT}}^{\alpha}\) is equivalent (in the LoRA power-law parametrization) to scaling capacity \(N\) by \(\Xi_{\textsc{GRIT}}\).
Thus the multiplier is an \emph{operational summary} of reduced curvature exposure.

\subsection{F. Step 5: constructing \(\Xi_{\textsc{GRIT}}\) from auditable geometry summaries}
\label{app:step5_xi_construct}

\paragraph{\textbf{Three geometry summaries.}}
\textsc{GRIT} introduces three measurable quantities:
\begin{itemize}[leftmargin=1.5em]
  \item \(\mathbf{r_{\mathrm{eff}}}\): effective usable rank (capacity actually used, not merely allocated),
  \item \(\boldsymbol{\rho_{\mathrm{align}}}\): alignment of adapter update subspace with Fisher/K-FAC dominant eigendirections (stability-aligned geometry),
  \item \(\boldsymbol{\pi_{\mathrm{proj}}}\): reprojection-retained spectral mass (how much update energy is preserved in the stable subspace after projection).
\end{itemize}

\paragraph{\textbf{Multiplicative model.}}
We summarize these effects with a product-form multiplier:
\[
\Xi_{\textsc{GRIT}}
=
(1+\gamma_r r_{\mathrm{eff}})
(1+\gamma_a \rho_{\mathrm{align}})
(1+\gamma_p \pi_{\mathrm{proj}}),
\]
where \(\gamma_r,\gamma_a,\gamma_p\ge 0\) are fit constants.
This form encodes monotonicity: increasing any geometry-control term increases effective capacity and reduces forgetting under the same budget.

\paragraph{\textbf{Why a product form.}}
The three mechanisms act on distinct aspects of the trace expression:
\begin{itemize}[leftmargin=1.5em]
  \item \(r_{\mathrm{eff}}\) controls \emph{how concentrated} update energy is (capacity utilization vs diffuse interference),
  \item \(\rho_{\mathrm{align}}\) controls \emph{where} the update subspace lies relative to curvature,
  \item \(\pi_{\mathrm{proj}}\) controls \emph{how strongly} reprojection enforces the stable subspace over time.
\end{itemize}
Multiplicative composition is the simplest model consistent with weak interaction assumptions; if interactions are strong, we can add cross-terms in the fit (Appendix scaling-law audits).

\subsection{G. Step 6: the derived \textsc{GRIT} forgetting law}
\label{app:step6_grit_law}

Substituting the effective capacity equivalence \(N\mapsto \Xi_{\textsc{GRIT}}N\) into the LoRA law yields:
\[\boxed{
L_{\mathrm{pt}}^{\textsc{GRIT}}
= L_{\mathrm{pt}}^{0} + A\,\frac{D_{\mathrm{ft}}^{\beta}}{(\Xi_{\textsc{GRIT}}N)^{\alpha}} + E,
\qquad
\Xi_{\textsc{GRIT}}
=
(1+\gamma_r r_{\mathrm{eff}})
(1+\gamma_a \rho_{\mathrm{align}})
(1+\gamma_p \pi_{\mathrm{proj}}).
}
\]
This is the desired derivation: \textbf{LoRA budget law + explicit geometry term} \(\Rightarrow\) \textbf{\textsc{GRIT} law} via an effective capacity multiplier.

\subsection{H. Audit tables and figure-ready blocks}
\label{app:tables_figs_forgetting_nonum}

\begin{table}[t]
\centering
\small
\setlength{\tabcolsep}{6pt}
\renewcommand{\arraystretch}{1.15}
\caption{\textbf{Derivation map: how each mathematical object connects LoRA’s budget law to \textsc{GRIT}’s geometry multiplier.}}
\label{tab:derivation_map}
\begin{tabular}{@{}l p{0.72\linewidth}@{}}
\toprule
\textbf{Object} & \textbf{Role in the derivation} \\
\midrule
\(\Delta L_{\mathrm{pt}}\) & Forgetting target; approximated by curvature-weighted update energy \(\frac12\Delta w^\top H_{\mathrm{pt}}\Delta w\). \\
\(H_{\mathrm{pt}}\) & Pretraining curvature; defines which directions are ``sharp'' and amplify forgetting. \\
\(\Sigma_{\Delta}\) & Update covariance; yields expected forgetting via \(\frac12\mathrm{tr}(H_{\mathrm{pt}}\Sigma_{\Delta})\). \\
\(P^\top H_{\mathrm{pt}}P\) & Restricted curvature inside the adapter subspace; this is what geometry-aware PEFT modifies. \\
LoRA power law & Budget-only predictor \(L_{\mathrm{pt}}^{0} + A D_{\mathrm{ft}}^\beta/N^\alpha + E\). \\
\(\Xi_{\textsc{GRIT}}\) & Geometry multiplier summarized as an ``effective capacity'' scaling \(N\mapsto \Xi_{\textsc{GRIT}}N\). \\
\bottomrule
\end{tabular}
\end{table}

\begin{table}[ht!]
\centering
\small
\setlength{\tabcolsep}{6pt}
\renewcommand{\arraystretch}{1.18}
\caption{\textbf{Mechanism $\rightarrow$ geometry effect $\rightarrow$ proxy term in \(\Xi_{\textsc{GRIT}}\).}}
\label{tab:mechanism_to_xi}
\adjustbox{width=0.86\textwidth,center}{
\begin{tabular}{@{}l p{0.39\linewidth} p{0.36\linewidth}@{}}
\toprule
\textbf{Component} & \textbf{Geometry effect (trace view)} & \textbf{Proxy term in \(\Xi_{\textsc{GRIT}}\)}\\
\midrule
Rank-space K-FAC preconditioning &
Shrinks steps along ill-conditioned/sharp restricted directions; reduces curvature-amplified motion in \(\mathrm{tr}(P^\top H_{\mathrm{pt}}P\,\Sigma_\theta)\). &
Improves alignment/stability; reflected primarily through \(\rho_{\mathrm{align}}\) and indirectly through higher usable \(r_{\mathrm{eff}}\). \\
Fisher-guided reprojection &
Rotates/filters the adapter basis toward dominant stable eigendirections; suppresses noisy diffuse components that increase interference. &
Directly increases \(\rho_{\mathrm{align}}\) and \(\pi_{\mathrm{proj}}\). \\
Dynamic rank adaptation &
Allocates capacity where spectrum has energy; prevents over-spreading energy into weak/noisy directions. &
Controls \(r_{\mathrm{eff}}\) as the \emph{realized} capacity. \\
\bottomrule
\end{tabular}
}
\end{table}

\begin{figure}[ht!]
\centering
\includegraphics[width=\linewidth]{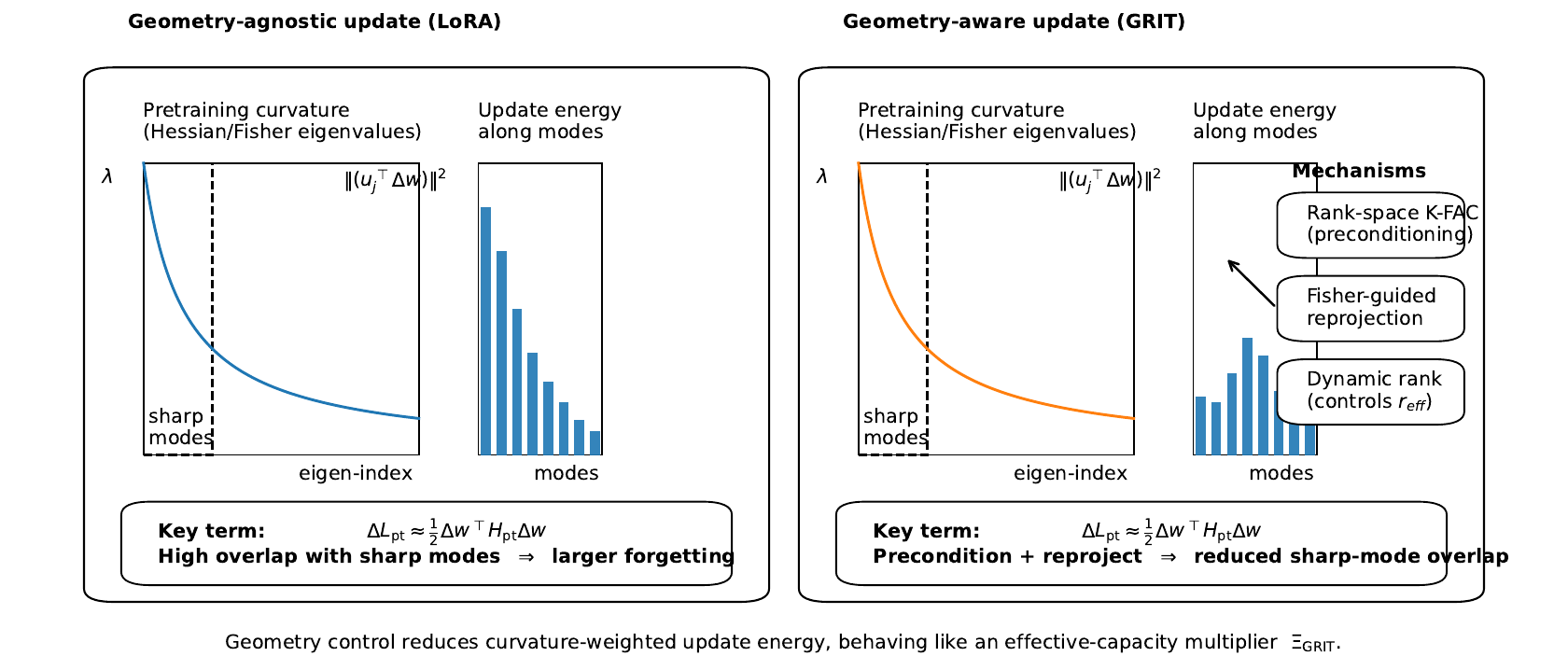}
\vspace{-0.6em}
\caption{\textbf{Quadratic curvature overlap view of forgetting.} Forgetting increases with curvature-weighted update energy. Geometry-aware updates reduce overlap with sharp curvature directions by preconditioning and reprojection, motivating an effective-capacity multiplier \(\Xi_{\textsc{GRIT}}\).}
\vspace{-0.8em}
\label{fig:quad_overlap_nonum}
\end{figure}

\begin{figure}[ht!]
\centering
\includegraphics[width=\linewidth]{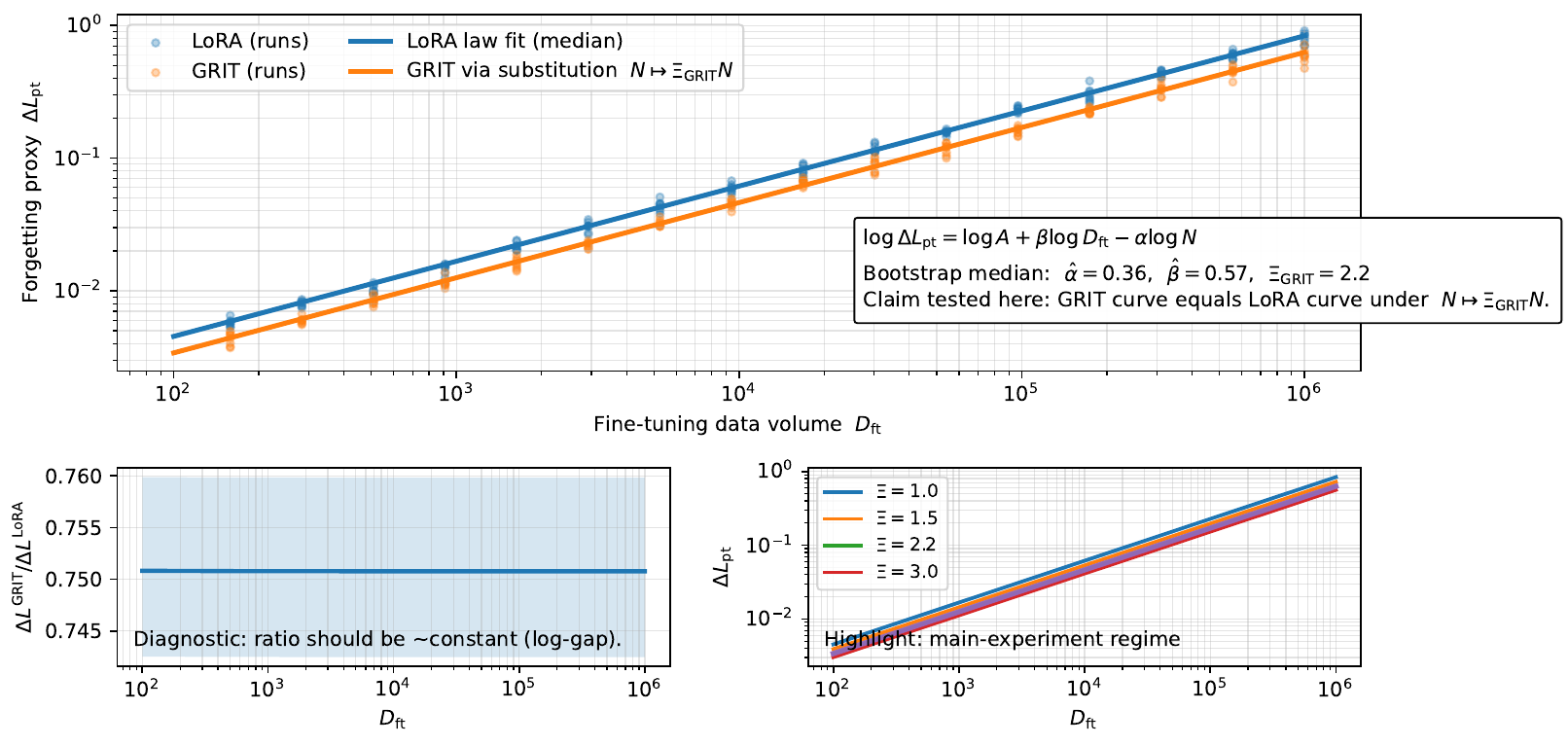}
\vspace{-0.6em}
\caption{\textbf{From LoRA law to \textsc{GRIT} law by effective capacity.} The \textsc{GRIT} curve equals the LoRA curve under the substitution \(N \mapsto \Xi_{\textsc{GRIT}}N\), compressing forgetting at fixed \((D_{\mathrm{ft}},N)\).}
\vspace{-0.8em}
\label{fig:overlay_nonum}
\end{figure}

\begin{figure}[ht!]
\centering
\includegraphics[width=\linewidth]{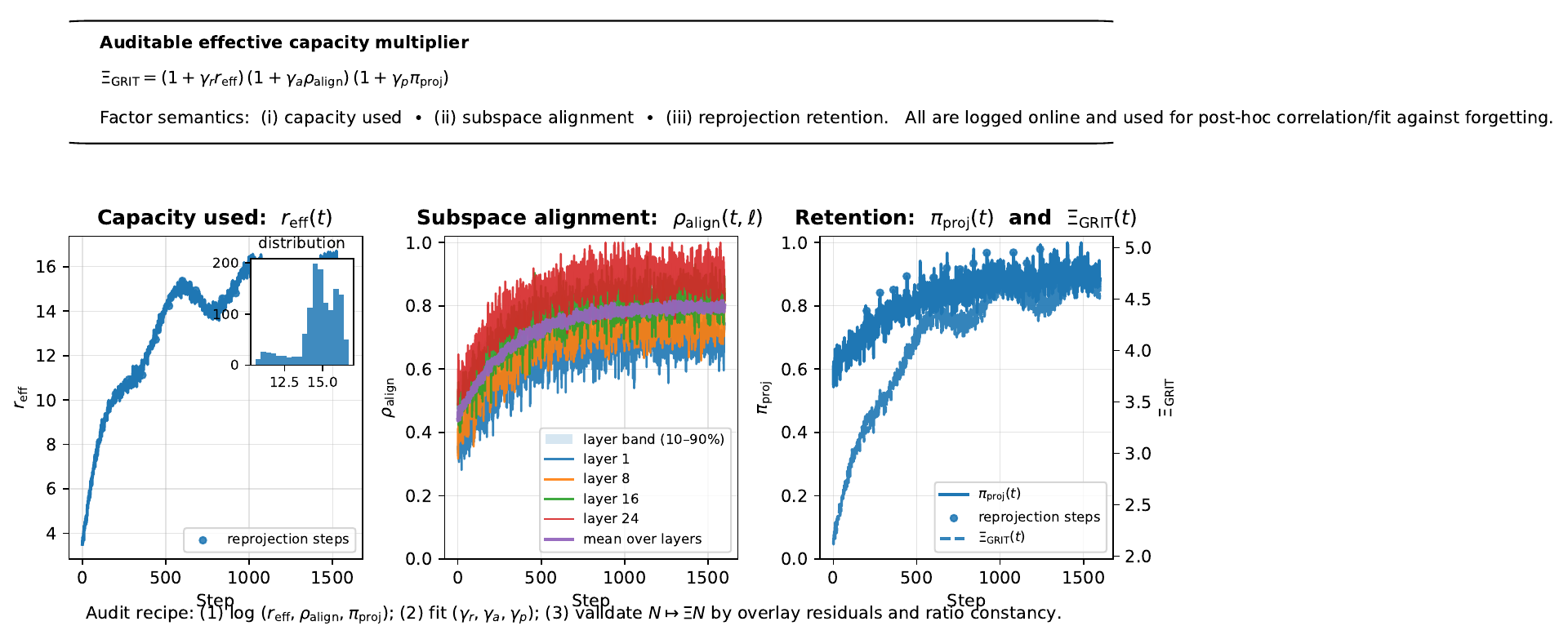}
\vspace{-0.6em}
\caption{\textbf{Auditable components of \(\Xi_{\textsc{GRIT}}\).} Illustrate how effective rank \(r_{\mathrm{eff}}\), alignment \(\rho_{\mathrm{align}}\), and reprojection-retained mass \(\pi_{\mathrm{proj}}\) compose multiplicatively to yield an effective capacity multiplier.}
\vspace{-0.8em}
\label{fig:xi_components_nonum}
\end{figure}

\paragraph{What to log to make the derivation falsifiable.}
To operationalize the law, log per-layer: spectra for \(r_{\mathrm{eff}}\), overlap/alignment scores for \(\rho_{\mathrm{align}}\), and reprojection-retained mass for \(\pi_{\mathrm{proj}}\), along with matched \(\Delta L_{\mathrm{pt}}\) under controlled sweeps of rank and reprojection frequency.

\end{document}